\documentclass{article} %
\usepackage{arxiv,times}

\usepackage{amsmath,amsfonts,bm}

\def\eqref#1{equation~\ref{#1}}

\def\1{\bm{1}}

\DeclareMathAlphabet{\mathsfit}{\encodingdefault}{\sfdefault}{m}{sl}
\SetMathAlphabet{\mathsfit}{bold}{\encodingdefault}{\sfdefault}{bx}{n}

\usepackage{hyperref}
\usepackage{url}
\usepackage{bbm}

\usepackage{booktabs}
\usepackage{graphicx}

\usepackage{xcolor}   %

\usepackage{kotex}
\usepackage{multirow}
\usepackage{adjustbox}
\usepackage{algorithm}
\usepackage{algpseudocode}
\usepackage{amssymb}
\usepackage{titletoc}
\usepackage{enumitem}

\usepackage{xcolor}
\usepackage{listings}
\usepackage[most]{tcolorbox}
\usepackage[table]{xcolor}

\usepackage{fontawesome5}

\definecolor{codebg}{RGB}{246,247,255}
\definecolor{codeblue}{RGB}{55,95,220}
\definecolor{codepurple}{RGB}{125,70,180}
\definecolor{codegray}{RGB}{95,100,110}

\lstdefinestyle{pytorch}{
    language=Python,
    basicstyle=\ttfamily\footnotesize,
    keywordstyle=\color{codeblue}\bfseries,
    commentstyle=\color{codegray},
    showstringspaces=false,
    breaklines=true,
    keepspaces=true,
    columns=fullflexible,
    frame=none,
    numbers=none,
    emph={
        fista_sparse_code,
        refit_coefficients,
        stop_gradient,
        nearest_coactive_pairs,
        reconstruction_loss,
        coactive_similarity_loss,
        normalize_dictionary_atoms
    },
    emphstyle=\color{codepurple}\bfseries
}

\newtcblisting{pytorchcode}{
    enhanced,
    listing only,
    listing engine=listings,
    listing options={style=pytorch},
    colback=codebg,
    colframe=codebg,
    boxrule=0pt,
    arc=2mm,
    left=2mm,
    right=2mm,
    top=1mm,
    bottom=1mm
}
\definecolor{darkblue}{RGB}{0,0,139}
\definecolor{darkgreen}{RGB}{0,100,0}

\hypersetup{
    colorlinks=true,      %
    linkcolor=darkblue,       %
    citecolor=darkgreen,      %
    urlcolor=darkblue,     %
    filecolor=cyan,       %
}

\usepackage{xspace}

\newcommand{\vrnt}[1]{#1}

\newcommand{\Refit}{\vrnt{+\,Refitting}\xspace}
\newcommand{\SimReg}{\vrnt{+\,Similarity Reg.}\xspace}
\newcommand{\Full}{\vrnt{Full ConSDL}\xspace}

\title{Beyond SAEs: Consistent Concept\\Discovery with Sparse Coding}

\author{
Jinho Choi\textsuperscript{1,2}\thanks{This work was done during a research visit at Helmholtz Munich.} \quad
Hyesu Lim\textsuperscript{2} \quad
Jaegul Choo\textsuperscript{1} \quad
Steffen Schneider\textsuperscript{2}\thanks{Correspondence: \texttt{steffen.schneider@helmholtz-munich.de}} \\
\textsuperscript{1}KAIST \qquad
\textsuperscript{2}Helmholtz Munich
}

\iclrfinalcopy
\begin{document}

\maketitle

\begin{abstract}
Sparse autoencoders (SAEs) explain foundation model representations by decomposing activations into sparse combinations of learned dictionary atoms.
However, the learned dictionaries can vary across training runs, limiting the reliability and reproducibility of SAE-based analyses.
We propose Consistent Sparse Dictionary Learning (ConSDL), which replaces the standard linear encoder with per-sample sparse coding.
This is motivated by our synthetic analysis showing that coefficients from a single linear encoder pass are suboptimal compared with those obtained by per-sample optimization, suggesting that sparse coding is a reliable mechanism for dictionary learning.
Across 16 synthetic settings, ConSDL achieves near-perfect dictionary and coefficient recovery. These gains translate to state-of-the-art cross-seed consistency across four vision and language settings, where ConSDL improves dictionary and coefficient consistency by 17.6\% and 35.9\%, respectively, over the average of six SAE baselines. 

\end{abstract}

\begin{center}
\small
\href{https://github.com/dynamical-inference/consdl}{\faGithub\ Code}
\quad \textbar \quad
\href{https://huggingface.co/dynamical-inference/consdl}{\faDownload\ Weights}
\quad \textbar \quad
\href{https://dynamical-inference.github.io/consdl}{\faGlobe\ Project Page}
\end{center}

\section{Introduction}

Sparse autoencoders (SAEs) \citep{cunningham2023sparse} have become a widely used approach for analyzing the internal representations of foundation models. SAEs decompose each model activation into a sparse linear combination of vectors from a learned overcomplete dictionary. Empirically, these dictionary vectors often align with human-interpretable concepts, such as colors, textures, and objects in vision models \citep{thasarathan2025universal} or characters, words, and syntactic patterns in language models~\citep{elhage2022toy, bricken2023monosemanticity}. Importantly, these concepts can be learned without a predefined vocabulary, enabling scalable explanations of model behavior in vision \citep{lim2025sparse, pach2026sparse} and language \citep{templeton2026scaling}, audits of hidden biases in models and datasets \citep{ahsan2025can, choi2026conceptscope}, and identification of candidate factors in scientific and medical data \citep{dasdelen2025cytosae, nerrise2026geosae}.

Yet, a critical limitation remains before SAEs can serve as practical tools for analysis. SAEs trained on the same data with the same configuration can learn substantially different dictionaries across random seeds \citep{paulo2025sparse}. This limits reproducibility and reliability. More fundamentally, it raises the question of whether the features learned by SAEs genuinely reflect the model's internal representations or are artifacts of the training procedure. Recent work has further heightened this concern by showing that SAEs can fail to recover known ground-truth features in controlled synthetic settings \citep{korznikov2026sanity,pacela2026stop}.

In this paper, we propose Consistent Sparse Dictionary Learning (ConSDL) to make sparse decompositions of model activations consistent across independent training runs, enabling reliable and reproducible analysis. To this end, ConSDL replaces the linear encoder of SAEs with per-sample sparse coding.
The algorithm design is motivated by prior work showing that coefficients predicted by a single encoder pass can be suboptimal compared with those obtained by per-sample optimization~\citep{o2024compute}. We hypothesize that updating the dictionary with these suboptimal coefficients contributes to inconsistency across training runs. ConSDL therefore uses per-sample sparse coding throughout training to obtain coefficients before each dictionary update. To make this procedure stable in practice, we additionally use coefficient refitting to correct shrinkage from the sparsity penalty and a co-active similarity regularizer to reduce dictionary redundancy.

Across diverse synthetic and real-world settings, ConSDL consistently outperforms existing SAEs. In controlled synthetic experiments, ConSDL achieves near-perfect recovery of ground-truth dictionaries and coefficients, even with larger dictionaries and highly imbalanced atom-selection probabilities, where existing methods degrade. These results support our hypothesis that improved coefficient estimation leads to more consistent dictionary learning. ConSDL substantially improves dictionary and coefficient consistency over existing methods on vision and language foundation models. 

Our contributions are threefold. First, we introduce ConSDL, which incorporates per-sample sparse coding to produce consistent sparse decompositions across training runs. Second, we systematically study SAE recovery under synthetic conditions and identify when existing methods fail and what factors contribute to these failures. Third, we show that ConSDL improves cross-seed dictionary and coefficient consistency on foundation-model activations across multiple vision and language backbones, dictionary widths, and sparsity levels.

\section{Related work}
\textbf{Consistency of SAEs.} Recent work has drawn attention to the inconsistency of SAEs, showing that independent training runs can learn substantially different dictionaries~\citep{paulo2025sparse}. \citet{fel2025archetypal} improve dictionary consistency by constraining dictionary atoms to lie within the convex hull of the activation data. \citet{nelson2026toward} instead regularizes the dictionary geometry to reduce ambiguity among sparse codes, improving dictionary and coefficient consistency. 
However, these studies differ in key experimental choices, including dictionary size, sparsity, and training procedure, making direct comparison difficult. \citet{song2025position} further show that SAE consistency can vary substantially across such configurations. It therefore remains unclear whether the reported gains hold under matched settings. We address this by comparing existing methods across a broad range of settings using the same evaluation protocol.

\textbf{Sparse coding and dictionary learning.}
Given a fixed dictionary, sparse coding infers a sparse coefficient to reconstruct each input as a linear combination of dictionary codes \citep{olshausen1997sparse}. A common formulation uses an $\ell_1$ penalty and can be solved with ISTA \citep{daubechies2004iterative} or its accelerated variant FISTA \citep{beck2009fast}, while LISTA \citep{gregor2010learning} unrolls ISTA into a fixed-depth network with learned parameters for faster inference. Alternatively, matching pursuit greedily selects atoms that best explain the current residual \citep{mallat1993matching}. Dictionary learning jointly learns the dictionary by alternating between sparse coefficient estimation and dictionary updates \citep{aharon2006k,mairal2010online}. Our proposed ConSDL adopts this alternating formulation, using FISTA for per-sample sparse inference before each dictionary update. Under suitable assumptions, such iterative approaches are theoretically guaranteed to recover the underlying dictionary and coefficients \citep{agarwal2016learning}.

\textbf{From amortized inference to per-sample sparse coding.}
Standard SAEs use a shared learned encoder to predict sparse coefficients rather than optimizing them separately for each input. This amortized inference~\citep{marino2018iterative} avoids costly per-sample optimization and enables efficient large-scale training, but at a cost to coefficient quality. \citet{o2024compute} show that optimizing coefficients directly under a fixed dictionary substantially improves recovery. \citet{sunprice} connect amortized inference to feature absorption, splitting, and dead atoms. \citet{joung2026fista} retain an amortized encoder but iteratively refine its predicted coefficients during dictionary learning. 
\citet{pacela2026stop} instead remove the learned encoder and use per-sample sparse coding throughout training, but report degraded recovery for larger dictionaries. 
Our ConSDL builds on this non-amortized approach while improving its stability and scalability to foundation-model activations. 

\textbf{Recovery guarantees for sparse coding and dictionary learning}
Sparse recovery theory characterizes when the underlying coefficients can be recovered from observations generated by a known dictionary~\citep{donoho2003optimally,candes2006stable}. Exact recovery is possible when coefficients are sufficiently sparse and the dictionary atoms are sufficiently distinct, so that different sparse codes do not produce the same observation~\citep{donoho2003optimally}. Restricted isometry provides a stronger formal condition by requiring the dictionary to approximately preserve the geometry of sparse vectors~\citep{candes2006stable}. When the dictionary is also unknown, the problem additionally requires identifying the dictionary from multiple sparse observations. Under suitable assumptions, including sufficient observations and an initialization close to the ground-truth dictionary, alternating minimization can recover both an overcomplete dictionary and its coefficients~\citep{agarwal2016learning}. 
Additional related work is discussed in Appendix~\ref{sec:appendix:extened_related_work}.

\section{Method}
\begin{figure}[t]
\centering
\includegraphics[width=1.0\textwidth]{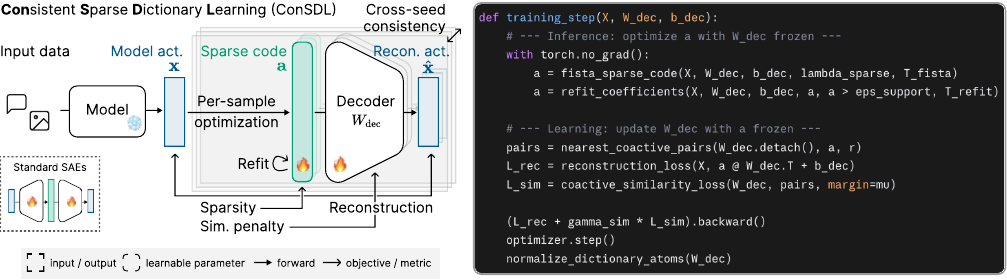}
\vspace{-1.5em}
\caption{
\textbf{Method.}
\textbf{Left}: Consistent Sparse Dictionary Learning (ConSDL) alternates between per-sample sparse coefficient inference with refitting and decoder updates under a reconstruction objective with co-active similarity regularization, holding the other component fixed at each stage. Standard SAEs (inset) instead use a learned encoder trained jointly with the decoder. Stacked panels indicate independent training seeds, across which ConSDL aims to recover consistent codes and dictionaries. \textbf{Right}: Python-style pseudocode.
}
\label{fig:method}
\end{figure}

We propose Consistent Sparse Dictionary Learning (ConSDL) to improve the consistency of dictionaries and coefficients across training runs (Fig. \ref{fig:method}). ConSDL replaces the learned encoder with per-input coefficient optimization and updates the dictionary from these estimates. It then refits the magnitudes of the selected coefficients without the sparsity penalty to correct shrinkage from sparse coding, and encourages distinct co-active atoms by penalizing the similarity of the most similar co-active pair.

\subsection{Per-sample Sparse Coding}
\label{sec:sparse_coding_refitting}

Let $\mathbf{x}\in\mathbb{R}^{d}$ denote a model activation with dimension $d$, which we aim to decompose into a linear combination of $m$ dictionary atoms weighted by non-negative, sparse coefficient vector. Let $W_{\mathrm{dec}}\in\mathbb{R}^{d\times m}$ denote the learned dictionary, corresponding to the decoder weight in an SAE. Its $j$-th
column $\mathbf{w}_j\in\mathbb{R}^{d}$ represents a unit-norm dictionary atom. Let $\mathbf{b}_{\mathrm{dec}}\in\mathbb{R}^{d}$ denote the decoder bias, and $\mathbf{a}\in\mathbb{R}_{\geq 0}^{m}$ denote the nonnegative coefficient vector, with $a_j\in\mathbb{R}$ specifying the contribution of atom $j$ to the reconstruction.

Standard SAEs use a linear encoder followed by a non-linearity to predict the sparse coefficients $\mathbf{a}$ from each input in a single forward pass.
These coefficients directly influence the gradients used to update each dictionary atom during training, yet coefficients produced by a single encoder pass can be suboptimal compared with those optimized for the current dictionary \citep{o2024compute}. We therefore replace the encoder in a conventional SAE with per-sample sparse coding and infer the coefficients for each input as the solution to the optimization problem: 
\begin{equation}
\mathbf{a}^{\mathrm{sc}}(\mathbf{x})
=
\operatorname*{arg\,min}_{\mathbf{a}\geq 0}
\frac{1}{2}
\left\|
\mathbf{x}
-
W_{\mathrm{dec}}\mathbf{a}
-
\mathbf{b}_{\mathrm{dec}}
\right\|_2^2
+
\lambda\|\mathbf{a}\|_1,
\label{eq:sparse_coding}
\end{equation}
where $\lambda>0$ controls the sparsity penalty. 
We estimate the solution using the Fast Iterative Shrinkage-Thresholding Algorithm (FISTA; \citealp{beck2009fast}). Each FISTA iteration updates the coefficients by taking a gradient step on the reconstruction term, followed by nonnegative soft thresholding for the \(\ell_1\) penalty.

\subsection{Coefficient Refitting}
While the $\ell_1$ penalty promotes sparsity by penalizing coefficient magnitude, it also introduces \textit{shrinkage bias}: even after the active atoms are correctly identified, the optimization objective favors smaller nonzero coefficients to reduce the penalty~\citep{zou2006adaptive}. Using these shrunken coefficients directly for decoder updates can underestimate the contributions of the selected atoms \citep{tolooshams2021stable}. We therefore retain the nonzero coefficients identified by FISTA and refit them without the $\ell_1$ penalty, following post-selection refitting procedures in sparse estimation~\citep{figueiredo2007gradient,chzhen2019lasso}.
Collecting the indices of non-zero coefficients in the support $S^{\mathrm{sc}}(\mathbf{x}) = \{j:a_j^{\mathrm{sc}}(\mathbf{x})>0\}$, we solve
\begin{equation}
\mathbf{a}^{\mathrm{refit}}(\mathbf{x})
=
\operatorname*{arg\,min}_{
\substack{
a_j \forall j\in S^{\mathrm{sc}}(\mathbf{x})
}
}
\frac{1}{2}
\left\|
\mathbf{x}
-
W_{\mathrm{dec}}\mathbf{a}
-
\mathbf{b}_{\mathrm{dec}}
\right\|_2^2,
\label{eq:coefficient_refitting}
\end{equation}
using projected gradient descent initialized from $\mathbf{a}^{\mathrm{sc}}(\mathbf{x})$, constraining the coefficients to remain nonnegative and supported on $S^{\mathrm{sc}}(\mathbf{x})$. Thus, refitting adjusts only the magnitudes of the selected coefficients without activating new atoms. Additional details on the FISTA optimization and refitting procedure are provided in Appendix~\ref{sec:app:method_details}.

\subsection{Co-active Similarity Regularization}
\label{sec:coactive_similarity}
Per-sample sparse coding improves coefficient estimation, but applying it throughout dictionary learning can still lead multiple atoms to converge to similar directions, resulting in redundant dictionary elements. To stabilize dictionary learning in practice, we introduce a co-active similarity regularizer that penalizes high similarity only between atoms active for the same sample, rather than between all atom pairs in the dictionary.

For each input $\mathbf{x}_i$, let
$A_i=\{\mathbf{w}_j : a_{i,j}^{\mathrm{refit}}>0\}$
denote the set of active dictionary atoms, where $j$ refers to the atom index. For each
$\mathbf{w}_j\in A_i$, we define its most similar co-active atom as
\begin{equation}
\widetilde{\mathbf{w}}_{i,j}
=
\operatorname*{arg\,max}_{\mathbf{w}\in A_i\setminus\{\mathbf{w}_j\}}
\mathbf{w}_j^\top\mathbf{w},
\qquad
s_{i,j}
=
\mathbf{w}_j^\top\widetilde{\mathbf{w}}_{i,j}.
\label{eq:coactive_neighbor}
\end{equation}
where $s_{i,j}$ is their cosine similarity, since the atoms are unit normalized. 
To focus the regularization on highly similar pairs, we follow margin-based
contrastive objectives~\citep{hadsell2006dimensionality} and formulate the
co-active similarity penalty as
\begin{equation}
\mathcal{L}_{\mathrm{sim}}
=
\frac{1}{|\mathcal{V}_{\mu}|}
\sum_{(i,j)\in\mathcal{V}_{\mu}}
\left(s_{i,j}-\mu\right)^2,
\label{eq:coactive_similarity_loss}
\end{equation}
where $\mu\in[0,1)$ is a similarity margin and
$\mathcal{V}_{\mu}
=
\{(i,j) : \mathbf{w}_j\in A_i,\ s_{i,j}>\mu\}$
is the set of matched co-active pairs whose similarity exceeds the margin.
Implementation details of the regularizer are provided in
Appendix~\ref{sec:app:method_details}.

\subsection{Training Objective and Alternating Optimization}
\label{sec:training_objective}

Given the refitted coefficients, we define the reconstruction loss as
\begin{equation}
\mathcal{L}_{\mathrm{rec}}
=
\frac{1}{2}
\left\|
\mathbf{x}
-
W_{\mathrm{dec}}\mathbf{a}^{\mathrm{refit}}(\mathbf{x})
-
\mathbf{b}_{\mathrm{dec}}
\right\|_2^2.
\end{equation}
The overall training objective is
$
\mathcal{L}
=
\mathcal{L}_{\mathrm{rec}}
+
\gamma_{\mathrm{sim}}\mathcal{L}_{\mathrm{sim}},$ where $\gamma_{\mathrm{sim}}>0$ controls the strength of co-active similarity regularization. For each update, we first infer and refit the coefficients under the current dictionary and identify the co-active atom pairs. These quantities are then held fixed while optimizing $W_{\mathrm{dec}}$ and $\mathbf{b}_{\mathrm{dec}}$. Thus, gradients do not propagate through coefficient estimation or pair selection, and training alternates between per-sample sparse coding and dictionary updates. Full optimization details are provided in Appendix~\ref{sec:app:method_details}, with hyperparameter settings in Appendix~\ref{sec:app:exp}.

\begin{figure}[t]
\centering
\includegraphics[width=1.0\textwidth]{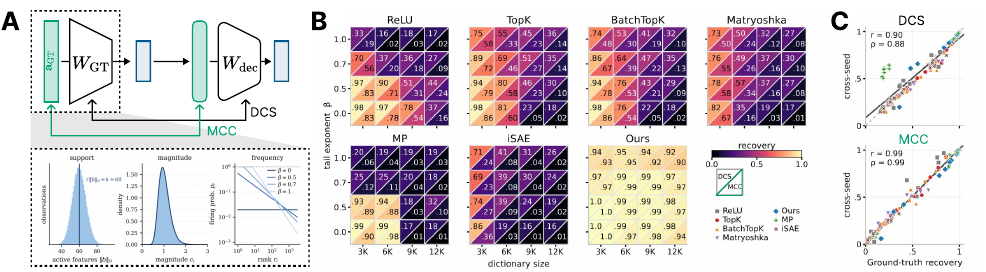}
\vspace{-2em}
\caption{\textbf{Synthetic recovery and cross-seed consistency.} \textbf{(A)} Activations are generated from a ground-truth dictionary $W_{\mathrm{GT}}$ with sparse codes $\mathrm{a}_{\mathrm{GT}}$. Insets show the sampling distributions for the active-set size, coefficient magnitudes, and atom selection frequencies. \textbf{(B)} Dictionary (\texttt{DCS}; upper-left) and coefficient  (\texttt{MCC}; lower-right) recovery across dictionary sizes and tail exponents $\beta$. \textbf{(C)} Cross-seed consistency correlates with ground-truth recovery for both dictionaries (top) and coefficients (bottom). Each point is one method at a given dictionary size and $\beta$. $r$ and $\rho$ denote the Pearson and Spearman correlation coefficients.}
\label{fig:synthetic_heatmap}
\end{figure}

\section{Synthetic Analysis of Dictionary and Coefficient Recovery}
\label{sec:preliminary}
To study the conditions under which existing SAE methods recover the underlying dictionary and coefficients, and whether ConSDL improves recovery where they fail, we use controlled synthetic data with a known ground-truth decomposition.

\subsection{Experimental Setup}
\label{sec:synthetic_setting}

\textbf{Synthetic data generation.} We generate synthetic data as
$\mathbf{x}=\sum_{i=1}^{m}a_i\mathbf{w}_i,$ where $\mathbf{x}\in\mathbb{R}^{d}$ is an observed activation, $\mathbf{w}_i\in\mathbb{R}^{d}$ denotes a ground-truth dictionary atom, and $a_i\in\mathbb{R}_{\geq0}$ is its nonnegative coefficient. In the baseline setting, we use activation dimension $d=768$, dictionary width $m=3072$, and an expected k=60 active atoms per sample (Fig.~\ref{fig:synthetic_heatmap}A).
Following prior synthetic SAE studies \citep{korznikov2026sanity, o2024compute}, we sample dictionary atoms independently from an isotropic Gaussian distribution and normalize them to unit norm. For each sample, atoms are selected independently with equal probability, yielding random sparse supports over an approximately incoherent dictionary, as commonly assumed in sparse coding analyses~\citep{arora2015simple}. Each nonzero coefficient is drawn from a log-normal distribution with expected value $1$. 

We extend the baseline setting along two axes motivated by properties of foundation-model representations: dictionary size and atom-selection imbalance. First, we vary the ground-truth dictionary width over $4\times$, $8\times$, $12\times$, and $16\times$ the activation dimension, corresponding to $m=3072$--$12288$. This range reflects evidence that larger dictionaries continue to uncover additional concepts in foundation model activations~\citep{templeton2026scaling}. Second, we vary atom-selection imbalance using Zipf exponents $\beta\in\{0,0.5,0.7,1.0\}$, where $\beta=0$ corresponds to uniform selection and larger values produce increasingly long-tailed frequencies, reflecting the highly imbalanced concept frequencies in web-scale pretraining data~\citep{udandarao2024no}. This results in 16 synthetic configurations.

\textbf{Baselines.} We compare a standard ReLU SAE~\citep{cunningham2023sparse} with TopK~\citep{gao2025scaling}, BatchTopK~\citep{bussmann2024batchtopk}, Matryoshka SAE~\citep{bussmann2025learning}, MP-SAE~\citep{costa2026flat}, and iSAE~\citep{hindupur2026projecting}. For each synthetic configuration, we generate $10^6$ observations and train each method for five epochs, reducing the likelihood that recovery failures arise from insufficient data or training. We repeat each experiment with three random seeds.

\textbf{Metrics.}
We evaluate dictionary recovery using decoder cosine similarity (\texttt{DCS})~\citep{paulo2025sparse} after Hungarian matching to the ground-truth dictionary. Coefficient recovery is measured by mean coefficient correlation (\texttt{MCC})~\citep{pacela2026stop} under the same matching. For cross-seed consistency, we apply the same metrics between independently trained runs and average over all seed pairs. Higher values indicate better recovery or consistency. Full metric definitions and experimental details are provided in Appendix~\ref{sec:app:exp}.

\subsection{Findings}

\begin{table*}[t]
\centering
\caption{\textbf{Per-sample re-optimization improves coefficient recovery under a fixed dictionary.}
Under uniform atom-selection frequencies ($\beta=0$), we compare amortized coefficients with per-sample FISTA re-optimization using fixed trained decoders.
Darker green indicates larger improvements over amortized inference.}
\label{tab:coeff_recovery_optimized_uniform}
\resizebox{\textwidth}{!}{%
\begin{tabular}{ll c cc cc cc}
\toprule
\multirow{2}{*}{\textbf{Architecture}} &
\multirow{2}{*}{\textbf{\# Atoms}} &
\multirow{2}{*}{\textbf{DCS}} &
\multicolumn{2}{c}{\textbf{$\ell_0$}} &
\multicolumn{2}{c}{\textbf{MCC}} &
\multicolumn{2}{c}{\textbf{Input $R^2$}} \\
\cmidrule(lr){4-5}
\cmidrule(lr){6-7}
\cmidrule(lr){8-9}
&
&
&
Amortized &
Re-optimized &
Amortized &
Re-optimized &
Amortized &
Re-optimized \\
\midrule
\multirow{4}{*}{ReLU}
& 3K
& 0.983 & 67.4 & 63.8 & 0.833 & \cellcolor{green!46}0.975 & 0.659 & \cellcolor{green!65}0.885 \\
& 6K
& 0.970 & 76.5 & 75.6 & 0.780 & \cellcolor{green!59}0.962 & 0.633 & \cellcolor{green!65}0.872 \\
& 9K
& 0.777 & 62.5 & 92.8 & 0.555 & \cellcolor{green!38}0.672 & 0.434 & \cellcolor{green!49}0.585 \\
& 12K
& 0.367 & 86.9 & 70.6 & 0.157 & 0.168 & 0.268 & 0.249 \\
\midrule
\multirow{4}{*}{TopK}
& 3K
& 0.982 & 60.0 & 60.0 & 0.862 & \cellcolor{green!28}0.947 & 0.812 & \cellcolor{green!35}0.920 \\
& 6K
& 0.810 & 60.0 & 60.0 & 0.611 & \cellcolor{green!19}0.669 & 0.646 & \cellcolor{green!17}0.699 \\
& 9K
& 0.227 & 60.0 & 60.0 & 0.048 & 0.048 & 0.403 & 0.427 \\
& 12K
& 0.181 & 60.0 & 60.0 & 0.023 & 0.023 & 0.400 & 0.423 \\
\bottomrule
\end{tabular}%
}
\end{table*}

\textbf{Recovery strongly depends on the data setting.} Despite differences across SAE variants, dictionary and coefficient recovery are generally high for smaller dictionaries and less long-tailed atom selection distributions (lower-left region of each heatmap in Fig.~\ref{fig:synthetic_heatmap}B). As dictionary size grows and atom selection becomes more long-tailed, both decline sharply (upper-right region). This sensitivity may help explain why prior synthetic studies have reported both successful~\citep{song2025position} and failed~\citep{pacela2026stop} SAE recovery. Since it is unclear which setting best reflects real model activations, these results motivate methods that remain reliable across a wider range of conditions.

\textbf{Amortization gap in coefficient inference.} Even in settings with high dictionary recovery, the encoder does not fully recover the corresponding coefficients. One possible explanation is the amortization gap identified in prior work~\citep{o2024compute}. We examine this by re-optimizing the coefficients for each sample while keeping the learned dictionary fixed. This improves both coefficient recovery and input reconstruction (Table \ref{tab:coeff_recovery_optimized_uniform} and \ref{tab:coeff_recovery_optimized}), showing that the encoder leaves recoverable error in coefficient inference. However, when dictionary recovery is poor, coefficient re-optimization yields limited gains, suggesting that a more fundamental solution requires reliable dictionary learning.

\textbf{ConSDL remains robust under challenging synthetic conditions.}
Across all 16 synthetic settings, ConSDL achieves near-perfect recovery of both the ground-truth coefficients and dictionary, including conditions where existing SAE methods degrade (Fig.~\ref{fig:synthetic_heatmap}B). This shows that incorporating per-sample sparse coding into dictionary learning not only addresses the coefficient-inference gap, but also improves dictionary recovery when incorporated into dictionary learning. By recovering nearly the same ground-truth decomposition across runs, ConSDL also achieves high cross-seed consistency. Full results are provided in Appendix~\ref{sec:app:additional_results} (Tables \ref{tab:synthetic_3k}--\ref{tab:synthetic_12k}).

\textbf{Cross-seed consistency is associated with ground-truth recovery.}
Across the synthetic sweeps, higher cross-seed consistency corresponds to stronger ground-truth recovery (Fig. \ref{fig:synthetic_heatmap}C). Decoder consistency is strongly correlated with dictionary recovery (Pearson $r=0.90$, Spearman $\rho=0.88$), while coefficient consistency shows an even stronger correlation with coefficient recovery ($r=0.99$, $\rho=0.98$). Although cross-seed consistency does not by itself establish ground-truth recovery, these strong correlations support its use as a complementary measure for real model activations, where the ground truth is unavailable.

\begin{figure}[t]
\centering
\includegraphics[width=1.0\textwidth]{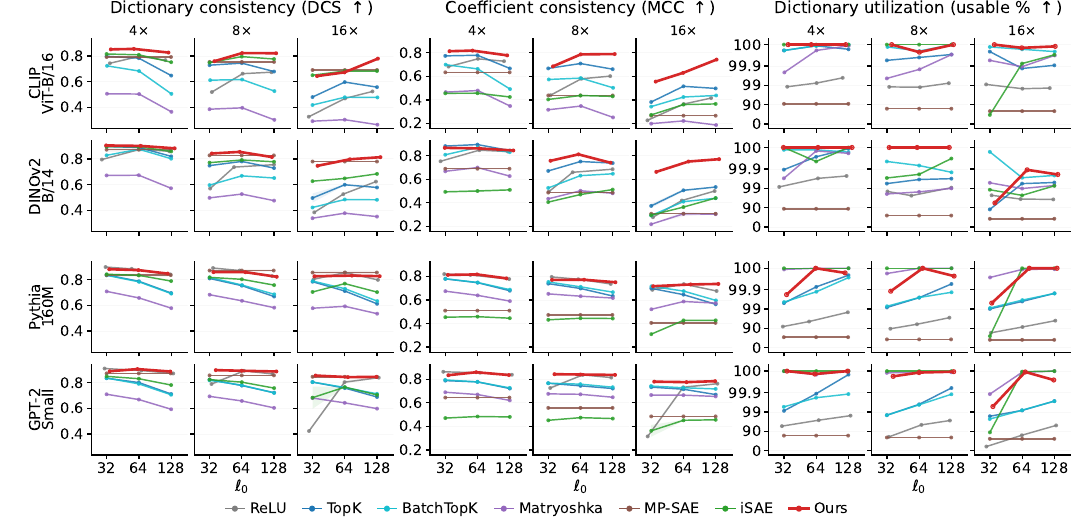}
\vspace{-2em}
\caption{\textbf{Consistency and dictionary utilization of foundation-model activations.}
We compare seven methods across two vision and two language backbones, three expansion factors, and three sparsity levels $\ell_0$. Cross-seed dictionary and coefficient consistency are measured by \texttt{DCS} and \texttt{MCC}, and dictionary utilization by the fraction of atoms that are neither dead nor duplicated. ConSDL achieves consistently higher cross-seed consistency, with the largest gains at $16\times$, while maintaining high dictionary utilization. 
}
\label{fig:real_consistency_lines}
\end{figure}

\section{Evaluation on Foundation Model Activations}
\label{sec:exp}
We next evaluate whether the gains observed in synthetic settings extend to foundation-model activations in vision and language domains. Since ground-truth decompositions are unavailable, we use cross-seed consistency as the primary evaluation criterion.

\subsection{Experimental Setup and Evaluation}

\textbf{Foundation model activations.}
For vision, we train SAEs on penultimate-layer activations from CLIP ViT-B/16~\citep{radford2021learning} and DINOv2~\citep{oquab2023dinov2}. We use ImageNet~\citep{russakovsky2015imagenet} and include both class-token and spatial-token activations following prior work~\citep{lim2025sparse}, training each SAE for one epoch over approximately 1.2M images. For language, we train SAEs on layer-9 activations from GPT-2 Small~\citep{radford2019language} and Pythia-160M~\citep{biderman2023pythia}. We train each SAE on 500M FineWeb tokens~\citep{penedo2024fineweb} with a context length of 1024. The activation dimension is 768 for all four models. We evaluate the trained SAEs on the ImageNet validation set for vision and on 20M FineWeb tokens for language. Additional model, data, and training details are provided in Appendix~\ref{sec:app:real_model}.

\textbf{Baselines and metrics.}
We use the same SAE baselines as in the synthetic evaluation. We evaluate dictionary widths of $4\times$, $8\times$, and $16\times$ the activation dimension, each at three sparsity levels, $\ell_0 \approx 32$, $64$, and $128$.  Because ground-truth decompositions are unavailable for foundation models, we evaluate cross-seed consistency of the dictionary and coefficients by training each configuration with three random seeds and computing \texttt{DCS} and \texttt{MCC} between independently trained runs, as in the synthetic evaluation. We further report dictionary utilization metrics to assess whether improved consistency comes at the cost of dictionary capacity. We measure the usable atom fraction, defined as the fraction of atoms that are neither dead nor duplicated. Dead atoms are never activated, while duplicated atoms point in highly similar decoder directions, reducing concept diversity.

\subsection{Main Results}

\textbf{ConSDL is more consistent than baselines across vision and language backbones.} 
Across 36 configurations spanning four backbones, three dictionary sizes, and three sparsity levels, ConSDL improves both dictionary and coefficient consistency over existing SAE baselines (Fig.~\ref{fig:real_consistency_lines}). Averaged across backbones, ConSDL achieves a \texttt{DCS} of 0.84 and an \texttt{MCC} of 0.78, corresponding to relative gains of 17.6\% and 35.9\% over the baseline average (Table \ref{tab:summary_consistency}). Compared with the strongest baseline for each metric, ConSDL improves dictionary and coefficient consistency by 1.9\% and 13.3\% on average. No single baseline is competitive on both metrics: MP-SAE achieves comparable \texttt{DCS} but substantially lower \texttt{MCC}, while ReLU and TopK achieve stronger coefficient consistency among the baselines but lower \texttt{DCS}. ConSDL maintains high consistency in both the learned dictionaries and coefficients across vision and language backbones.\footnote{We also compare \texttt{DCS} and \texttt{MCC} under activation-footprint matching instead of decoder cosine similarity, and obtain the same ranking across methods. See Table~\ref{tab:fp_axis_matching_basis}.} Detailed results for each backbone and configuration, along with hyperparameter sensitivity analyses are provided in Appendix~\ref{sec:app:additional_results} (Tables~\ref{tab:appendix_clip_full}--\ref{tab:appendix_pythia_full} and Fig.~\ref{fig:hyperparam_sensitivity}). 

\textbf{Consistency gains do not come at the cost of dictionary quality.} Duplicate atoms appear across several baselines, particularly ReLU and MP-SAE, while TopK-based methods and iSAE contain many dead atoms in some settings. As shown in the dictionary utilization panel of Fig.~\ref{fig:real_consistency_lines}, ConSDL maintains high usable atom fractions across backbones, dictionary sizes, and sparsity levels. Averaged across backbones, ConSDL achieves a usable atom fraction of 1.00, a 5.6\% improvement over the baseline average and matching the strongest baseline (Table~\ref{tab:summary_consistency}).
Reconstruction quality remains at a similar level to the baselines across configurations. We also report input $R^2$ in Tables~\ref{tab:appendix_clip_full}--~\ref{tab:appendix_pythia_full}.

\begin{figure}[t]
\centering
\includegraphics[width=1.0\textwidth]{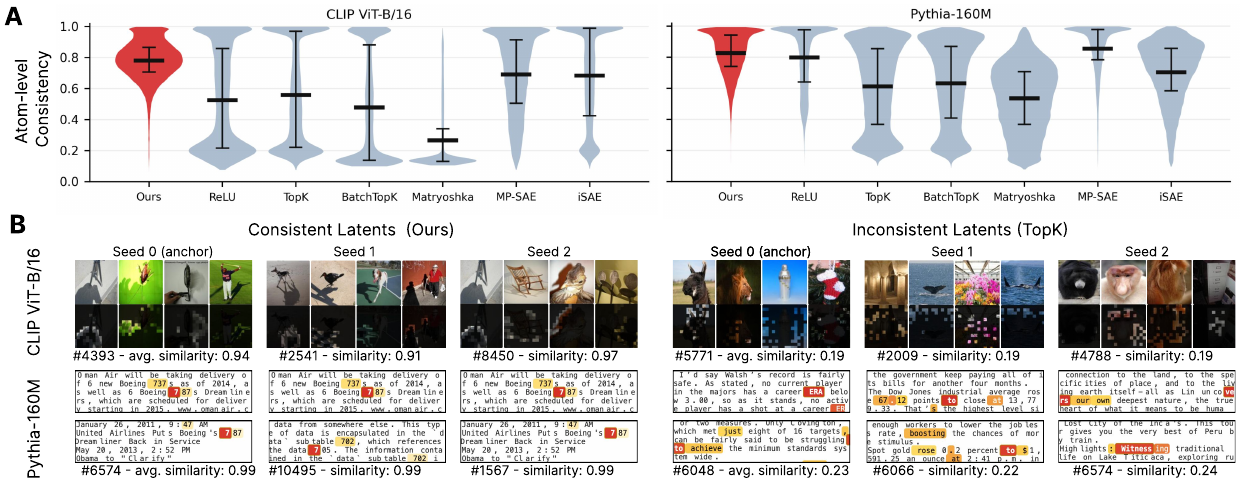}
\vspace{-2em}
\caption{\textbf{Atom-level cross-seed consistency distributions and activation patterns.}
\textbf{(A)} Distributions for CLIP ViT-B/16 and Pythia-160M. Each anchor atom's consistency is the mean decoder cosine similarity to its matched atoms in the other two seeds.
\textbf{(B)} Selected consistent ConSDL and inconsistent TopK atoms across three seeds, showing mean similarity for anchors and pairwise similarity for matches.
}
\label{fig:viloin_plots_with_examples}
\end{figure}

\textbf{Most ConSDL atoms exhibit high cross-seed consistency.}
Aggregate dictionary consistency scores do not reveal how consistency is distributed across individual atoms. We therefore examine atom-level consistency, defined as the mean decoder cosine similarity between each anchor atom and its matches in the other two seeds. Most ConSDL atoms exhibit high consistency on CLIP ViT-B/16 and Pythia-160M, whereas many baseline atoms have low consistency, particularly in vision (Fig.~\ref{fig:viloin_plots_with_examples}A). Similar trends are observed for DINOv2-B/14 and GPT-2 Small (Fig.~\ref{fig:consistency_dist_dino_gpt}). Representative examples show that consistent ConSDL atoms retain similar semantic activation patterns across seeds, whereas selected inconsistent TopK atoms respond to unrelated images or tokens (Fig.~\ref{fig:viloin_plots_with_examples}B). We also examine the number of atoms with consistency above 0.7 across dictionary sizes and sparsity levels (Fig.~\ref{fig:num_consistent_latents}). As dictionary size increases, ConSDL learns more such atoms, while most baselines show slower growth or saturation, particularly in vision. Higher sparsity levels yield smaller gains.

\subsection{Sparse Probing Evaluation}

\textbf{Evaluating concept localization with sparse probing.}
We next use sparse probing \citep{gurnee2023finding} to test whether class relevant information can be captured by a small number of dictionary atoms. Concentrating such information in a few atoms can make the learned representation easier to inspect and interpret. We evaluate vision models on six image classification datasets and language models on 113 binary tasks from SAE-Probes \citep{kantamneni2025sparse}, using the top one or five most discriminative latents. Additional experimental details are provided in Appendix~\ref{sec:app:sparse_probing}.

\textbf{ConSDL preserves sparse probing performance.}
Across vision and language models, ConSDL performs close to the strongest SAE baselines, achieving the best results in several settings and otherwise differing only marginally (Table \ref{tab:probing_x16}).
This suggests that the improved cross-seed consistency of ConSDL do not come at the cost of information localized to a small number of dictionary atoms. Full results are in Tables \ref{tab:vision_probing_main} and \ref{tab:lm_probing_main}. 

\begin{table}[t]
\centering
\caption{\textbf{Sparse probing performance at $16\times$ dictionary expansion ($\ell_0 \approx 128$).}
Probes use the $k$ most discriminative latents ($k \in \{1,5\}$).
Vision reports class-wise AUPRC averaged over six datasets.
Language reports AUROC averaged over 113 binary tasks in SAE-Probes.
Values are mean $\pm$ standard deviation over three seeds. Boldface and underlining indicate the best and second-best mean scores in each column, respectively.
}
\label{tab:probing_x16}

\setlength{\tabcolsep}{4pt}
\small
\resizebox{\textwidth}{!}{%
\begin{tabular}{l cc cc cc cc}
\toprule
& \multicolumn{4}{c}{\textbf{Vision (AUPRC)}} & \multicolumn{4}{c}{\textbf{Language (AUROC)}} \\
\cmidrule(lr){2-5}
\cmidrule(lr){6-9}
& \multicolumn{2}{c}{CLIP ViT-B/16} & \multicolumn{2}{c}{DINOv2-B/14} & \multicolumn{2}{c}{Pythia-160M} & \multicolumn{2}{c}{GPT-2 Small} \\
\cmidrule(lr){2-3}
\cmidrule(lr){4-5}
\cmidrule(lr){6-7}
\cmidrule(lr){8-9}
\textbf{Method} & @1 & @5 & @1 & @5 & @1 & @5 & @1 & @5 \\
\midrule
ReLU & 0.48\,{\scriptsize $\pm$0.01} & \underline{0.55}\,{\scriptsize $\pm$0.00} & \textbf{0.45}\,{\scriptsize $\pm$0.00} & \underline{0.53}\,{\scriptsize $\pm$0.00} & 0.70\,{\scriptsize $\pm$0.00} & 0.78\,{\scriptsize $\pm$0.00} & \underline{0.74}\,{\scriptsize $\pm$0.00} & \underline{0.79}\,{\scriptsize $\pm$0.00} \\
TopK & 0.46\,{\scriptsize $\pm$0.01} & 0.53\,{\scriptsize $\pm$0.00} & 0.43\,{\scriptsize $\pm$0.00} & \underline{0.53}\,{\scriptsize $\pm$0.00} & \underline{0.72}\,{\scriptsize $\pm$0.00} & 0.80\,{\scriptsize $\pm$0.00} & 0.75\,{\scriptsize $\pm$0.00} & 0.80\,{\scriptsize $\pm$0.00} \\
BatchTopK & 0.45\,{\scriptsize $\pm$0.00} & 0.52\,{\scriptsize $\pm$0.00} & 0.43\,{\scriptsize $\pm$0.00} & \underline{0.53}\,{\scriptsize $\pm$0.00} & 0.73\,{\scriptsize $\pm$0.00} & \textbf{0.80}\,{\scriptsize $\pm$0.00} & 0.75\,{\scriptsize $\pm$0.00} & 0.80\,{\scriptsize $\pm$0.00} \\
Matryoshka & 0.42\,{\scriptsize $\pm$0.01} & 0.46\,{\scriptsize $\pm$0.00} & 0.42\,{\scriptsize $\pm$0.00} & 0.51\,{\scriptsize $\pm$0.00} & \underline{0.72}\,{\scriptsize $\pm$0.00} & \underline{0.79}\,{\scriptsize $\pm$0.00} & \underline{0.74}\,{\scriptsize $\pm$0.00} & \underline{0.79}\,{\scriptsize $\pm$0.00} \\
MP-SAE & \underline{0.49}\,{\scriptsize $\pm$0.01} & \textbf{0.57}\,{\scriptsize $\pm$0.00} & \underline{0.44}\,{\scriptsize $\pm$0.00} & \textbf{0.54}\,{\scriptsize $\pm$0.00} & 0.67\,{\scriptsize $\pm$0.00} & 0.74\,{\scriptsize $\pm$0.00} & 0.67\,{\scriptsize $\pm$0.00} & 0.75\,{\scriptsize $\pm$0.00} \\
iSAE & 0.24\,{\scriptsize $\pm$0.01} & 0.50\,{\scriptsize $\pm$0.00} & 0.24\,{\scriptsize $\pm$0.02} & 0.47\,{\scriptsize $\pm$0.00} & \textbf{0.73}\,{\scriptsize $\pm$0.00} & \underline{0.79}\,{\scriptsize $\pm$0.00} & \underline{0.74}\,{\scriptsize $\pm$0.00} & 0.80\,{\scriptsize $\pm$0.00} \\
ConSDL & \textbf{0.52}\,{\scriptsize $\pm$0.01} & \underline{0.55}\,{\scriptsize $\pm$0.00} & \underline{0.44}\,{\scriptsize $\pm$0.00} & 0.50\,{\scriptsize $\pm$0.01} & 0.73\,{\scriptsize $\pm$0.00} & 0.78\,{\scriptsize $\pm$0.00} & \underline{0.74}\,{\scriptsize $\pm$0.00} & \underline{0.79}\,{\scriptsize $\pm$0.00} \\
\bottomrule
\end{tabular}%
}

\end{table}

\begin{table*}[t]
\centering
\caption{
\textbf{Ablation study at $16\times$ expansion and matched sparsity ($\ell_0 \approx 128$).}
Starting from per-sample sparse coding, we add coefficient refitting or similarity regularization individually, while \textbf{Full} combines both. Since these components affect sparsity, we compare configurations matched by $\ell_0$. Values are mean $\pm$ standard deviation over three seeds.
}

\label{tab:ablation_x16}

\setlength{\tabcolsep}{4pt}
\resizebox{\textwidth}{!}{%
\begin{tabular}{ll cc c cc ccc}
\toprule
\multirow{2}{*}{\textbf{Backbone}} &
\multirow{2}{*}{\textbf{Variant}} &
\multicolumn{2}{c}{\textbf{Sparsity}} &
\multicolumn{1}{c}{\textbf{Reconstruction}} &
\multicolumn{2}{c}{\textbf{Cross-seed Consistency}} &
\multicolumn{3}{c}{\textbf{Dictionary Utilization}} \\
\cmidrule(lr){3-4}
\cmidrule(lr){5-5}
\cmidrule(lr){6-7}
\cmidrule(lr){8-10}
&
&
$\lambda$ &
$\ell_0$ &
$R^2$ &
DCS &
MCC &
Usable atom frac. &
\# Dup. &
\# Dead \\
\midrule
\multirow{4}{*}{CLIP ViT-B/16}
& Sparse coding
& 1.2 & 116\,{\scriptsize $\pm$5.36}
& 0.87\,{\scriptsize $\pm$0.00}
& 0.84\,{\scriptsize $\pm$0.00} & \textbf{0.72}\,{\scriptsize $\pm$0.00}
& 0.25\,{\scriptsize $\pm$0.00} & 9,270\,{\scriptsize $\pm$35.3} & 0\,{\scriptsize $\pm$0.00} \\
& + Refitting
& 0.8 & 116\,{\scriptsize $\pm$1.88}
& \textbf{0.92}\,{\scriptsize $\pm$0.00}
& \textbf{0.85}\,{\scriptsize $\pm$0.00} & 0.70\,{\scriptsize $\pm$0.00}
& 0.55\,{\scriptsize $\pm$0.00} & 5,564\,{\scriptsize $\pm$74.9} & 0\,{\scriptsize $\pm$0.00} \\
& + Similarity reg.
& 1.0 & 110\,{\scriptsize $\pm$2.56}
& 0.89\,{\scriptsize $\pm$0.00}
& 0.70\,{\scriptsize $\pm$0.00} & 0.49\,{\scriptsize $\pm$0.00}
& 1.00\,{\scriptsize $\pm$0.00} & 56\,{\scriptsize $\pm$1.76} & 0\,{\scriptsize $\pm$0.00} \\
& \textbf{Full}
& 0.75 & 130\,{\scriptsize $\pm$2.13}
& 0.92\,{\scriptsize $\pm$0.00}
& 0.78\,{\scriptsize $\pm$0.00} & 0.66\,{\scriptsize $\pm$0.00}
& \textbf{1.00}\,{\scriptsize $\pm$0.00} & \textbf{1}\,{\scriptsize $\pm$0.90} & 0\,{\scriptsize $\pm$0.00} \\
\midrule
\multirow{4}{*}{Pythia-160M}
& Sparse coding
& 3.0 & 113\,{\scriptsize $\pm$2.45}
& 0.69\,{\scriptsize $\pm$0.00}
& 0.29\,{\scriptsize $\pm$0.00} & 0.17\,{\scriptsize $\pm$0.00}
& 0.03\,{\scriptsize $\pm$0.01} & 2,144\,{\scriptsize $\pm$52.4} & 9,767\,{\scriptsize $\pm$103} \\
& + Refitting
& 1.5 & 108\,{\scriptsize $\pm$0.32}
& 0.92\,{\scriptsize $\pm$0.02}
& \textbf{0.84}\,{\scriptsize $\pm$0.00} & \textbf{0.73}\,{\scriptsize $\pm$0.00}
& 0.74\,{\scriptsize $\pm$0.00} & 3,191\,{\scriptsize $\pm$26.7} & 0\,{\scriptsize $\pm$0.00} \\
& + Similarity reg.
& 2.9 & 133\,{\scriptsize $\pm$2.48}
& 0.72\,{\scriptsize $\pm$0.00}
& 0.30\,{\scriptsize $\pm$0.00} & 0.16\,{\scriptsize $\pm$0.00}
& 0.28\,{\scriptsize $\pm$0.02} & 239\,{\scriptsize $\pm$26.0} & 8,652\,{\scriptsize $\pm$345} \\
& \textbf{Full}
& 1.0 & 110\,{\scriptsize $\pm$0.47}
& \textbf{0.93}\,{\scriptsize $\pm$0.00}
& 0.81\,{\scriptsize $\pm$0.00} & 0.70\,{\scriptsize $\pm$0.00}
& \textbf{1.00}\,{\scriptsize $\pm$0.00} & \textbf{0}\,{\scriptsize $\pm$0.00} & 0\,{\scriptsize $\pm$0.00} \\
\bottomrule
\end{tabular}%
}
\end{table*}

\subsection{Ablation Study}

\textbf{Setup.} We ablate coefficient refitting and co-active similarity regularization to identify how each component contributes to the cross-seed consistency of ConSDL. We evaluate $16\times$ dictionaries at $\ell_0 \approx 128$ on CLIP ViT-B/16 and Pythia-160M (Table~\ref{tab:ablation_x16}). Because removing either component changes the resulting sparsity, we tune the sparsity penalty separately for each variant to obtain comparable $\ell_0$.

\textbf{Coefficient refitting stabilizes dictionary learning.} Comparing sparse coding (SC) with \Refit, SC requires a larger \(\lambda\) to reach a similar sparsity level, especially on Pythia (\(3.0\) vs. \(1.5\)). This stronger sparsity penalty is accompanied by 9,767 dead atoms and substantially lower reconstruction quality (\(R^2=0.69\)). This severe failure of SC on Pythia is consistent with prior reports that sparse-coding-based dictionary learning can degrade at larger scales~\citep{pacela2026stop}. On CLIP, SC does not produce dead atoms but still requires a stronger sparsity penalty. Refitting removes the dead atoms on Pythia and raises \(R^2\) to \(0.92\) on both backbones, from \(0.69\) on Pythia and \(0.87\) on CLIP.

\textbf{Similarity regularization reduces redundancy.} However, \Refit still leaves thousands of duplicated atoms on both backbones. Comparing \Refit with \Full, similarity regularization reduces duplicated atoms from 5,564 to 1 on CLIP and from 3,191 to 0 on Pythia. Although \Refit achieves higher DCS, its DCS decreases after deduplication across cosine similarity thresholds, whereas \Full maintains high DCS with almost no duplicates (Fig.~\ref{fig:effect_of_dup}). The same effect appears without refitting. Applying similarity regularization without refitting (\SimReg) also substantially reduces duplication, but does not resolve the instability observed on Pythia. These results indicate that both components are necessary for stable dictionary learning.

\begin{figure}[t]
\centering
\includegraphics[width=1.0\textwidth]{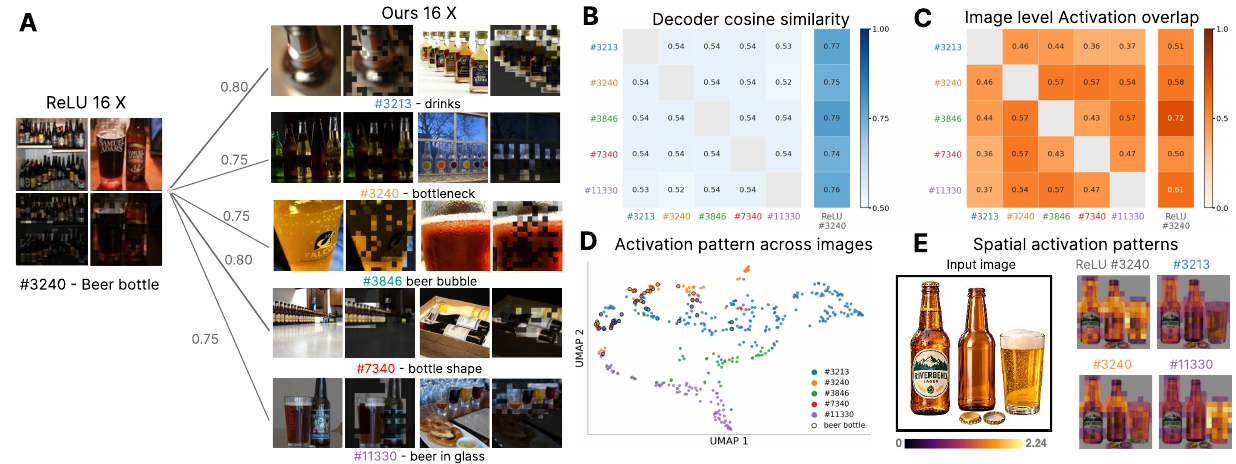}
\vspace{-2em}
\caption{\textbf{ConSDL captures distinct visual features associated with a broad concept.}
We compare ConSDL and ReLU SAE on CLIP ViT-B/16 ($16\times$, $\ell_0 \approx 128$).
\textbf{(A)} Five ConSDL atoms have decoder cosine similarities above 0.7 to a ReLU ``beer bottle'' atom.
\textbf{(B--C)} These atoms exhibit moderate pairwise decoder similarity ($\sim 0.5$) and partially overlapping image-level activations (Jaccard 0.36--0.57).
\textbf{(D)} UMAP of image-level activation vectors reveals distinct clusters.
\textbf{(E)} ConSDL atoms exhibit overlapping but distinct spatial activation patterns, whereas the ReLU atom responds broadly to beer-related objects.}
\label{fig:ours_vs_baseline}
\end{figure}

\section{Analyzing Latent Consistency and Concept Discovery}

We further investigate the learned atoms by examining the granularity of their semantic concepts and how these concepts change with dictionary size. We then assess whether cross-seed consistency extends beyond individual atoms to the overall dictionary structure.

\subsection{Finer-Grained Concept Discovery}

\textbf{ConSDL captures finer-grained concepts than baseline SAEs.}
On CLIP ViT-B/16 at $16\times$ expansion ($\ell_0 \approx 128$), ConSDL learns approximately 9K consistent atoms (cross-seed cosine similarity $>0.7$), compared with 4K for ReLU SAE (Fig.~\ref{fig:num_consistent_latents}). To understand what these additional atoms represent, we examine ConSDL atoms with decoder directions similar to a single ReLU atom. In the ``beer bottle'' example, a ReLU atom broadly responds to beer-related objects, whereas matched ConSDL atoms capture more specific features, such as bottle necks, bubbles, and beer in glasses (Fig.~\ref{fig:ours_vs_baseline}). These atoms can activate on the same images while responding to different visual features or regions. A similar pattern appears in Pythia-160M, where a ReLU atom associated with ``baseball'' matches several ConSDL atoms responding to more specific concepts, including team names, pitching, and field positions, with limited token-level activation overlap (Fig.~\ref{fig:ours_vs_baseline_language}). Across all four backbones, atoms from several baselines more often have multiple ConSDL counterparts than the reverse, extending this pattern beyond individual examples (Figs.~\ref{fig:ours_vs_baseline_heatmap} and \ref{fig:ours_vs_baseline_dist}).

\begin{figure}[t]
\centering
\includegraphics[width=1.0\textwidth]{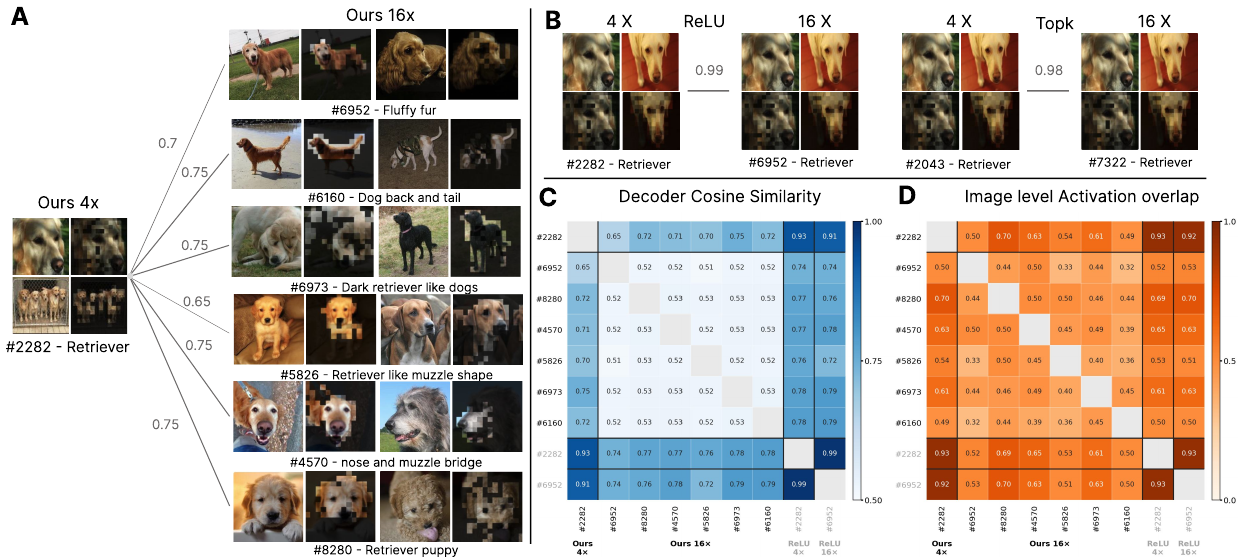}
\vspace{-2em}
\caption{\textbf{Finer-grained concepts in larger ConSDL dictionaries.}
We compare CLIP ViT-B/16 dictionaries at $4\times$ ($\ell_0 \approx 32$) and $16\times$ ($\ell_0 \approx 128$) expansion.
\textbf{(A)} A ConSDL atom associated with ``Golden Retriever'' at $4\times$ corresponds to six $16\times$ atoms (decoder cosine similarity 0.65--0.75), capturing features such as fur, tails, muzzle shapes, and puppies.
\textbf{(B)} ReLU and TopK retain nearly identical retriever-related atoms across dictionary sizes (cosine similarity 0.99 and 0.98).
\textbf{(C--D)} The ConSDL atoms exhibit moderate pairwise decoder similarity ($\sim 0.5$) and partial image-level activation overlap (Jaccard 0.32--0.50).}
\label{fig:x4_vs_x16}
\end{figure}

\textbf{Larger ConSDL dictionaries reveal more detailed concepts.}
We next examine how learned concepts change across dictionary sizes. In the ``Golden Retriever'' example, ConSDL learns several related atoms at $16\times$ that capture different visual attributes, whereas ReLU and TopK largely preserve the original broad concept (Fig.~\ref{fig:x4_vs_x16}). Although the ConSDL atoms share a common association with retrievers, their decoder directions and activation patterns exhibit meaningful differences. A similar pattern appears in Pythia-160M, where a $4\times$ ConSDL atom associated with plants matches several $16\times$ atoms capturing plant species, cultivation, and flowers, while ReLU retains a highly similar plant-related atom across dictionary sizes (Fig.~\ref{fig:x4_vs_x16_language}). Across all four backbones, consistent ConSDL atoms at $4\times$ are more likely to have multiple counterparts at $16\times$ than those from ReLU or TopK (Fig.~\ref{fig:x4_x16_dist}).

\subsection{Cross-Seed Consistency of Dictionary Structure}

\textbf{Measuring cross-seed agreement in cluster sizes and centroids.} We next examine whether atom-level consistency extends to the overall structure of the dictionary. We focus on CLIP ViT-B/16 and cluster atoms based on decoder direction similarity. We select one training run as a reference and evaluate whether other runs recover similar cluster sizes and centroids. For cluster sizes, we assign atoms from each run to the nearest reference centroid and compare the resulting cluster sizes using Spearman correlation. For cluster centroids, we independently cluster each run, match its centroids one-to-one to the reference centroids, and measure their cosine similarity. We compare this agreement with that obtained by rerunning \(k\)-means on the reference dictionary to account for clustering variability.

\begin{figure}[t]
\centering
\includegraphics[width=1.0\textwidth]{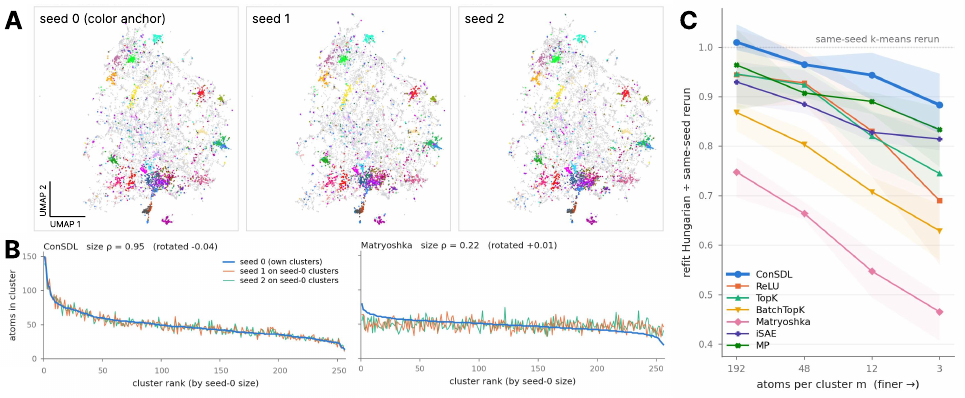}
\vspace{-2em}
\caption{
\textbf{Dictionary cluster consistency across training runs.}
\textbf{(A)} Joint UMAP of ConSDL decoder atoms at $16\times$, $\ell_0 \approx 128$, colored by top 30 clusters from a reference run.
\textbf{(B)} Cluster-size profiles for ConSDL and Matryoshka. $\rho$ denotes the Spearman correlation of cluster sizes across runs.
\textbf{(C)} Normalized cluster-centroid agreement across granularities. Lines show the mean over nine CLIP settings, and shading shows the interquartile range.
}
\label{fig:clustering}
\end{figure}

\textbf{ConSDL preserves fine-grained dictionary structure across seeds.} As shown in Fig.~\ref{fig:clustering}, ConSDL reproduces both cluster sizes and cluster centroids more consistently across training runs. At $16\times$ expansion and $\ell_0 \approx 128$, its cluster-size rank correlation reaches $0.95$, compared with $0.22$ for Matryoshka (Fig.~\ref{fig:clustering}B). ConSDL also achieves the highest normalized centroid agreement at every granularity, with scores of $1$, $0.96$, $0.94$, and $0.88$ at 192, 48, 12, and 3 atoms per cluster, respectively (Fig.~\ref{fig:clustering}C). A score of 1 indicates that cross-run agreement matches that obtained by rerunning $k$-means on the same dictionary. The gap between methods widens at finer granularities, indicating that ConSDL more reliably preserves fine-grained dictionary structure across training runs.

\section{Discussion and Limitations}
Although ConSDL improves cross-seed consistency over existing SAEs on foundation-model activations, it remains lower than the recovery achieved in synthetic settings. This gap may reflect the greater complexity of real activations, including a larger number of underlying concepts, stronger frequency imbalance, and deviations from the assumption that activations are sparse linear combinations of concept directions. Recent work has also questioned whether all concepts can be represented by individual linear directions, suggesting that some may instead exhibit nonlinear or manifold-like structure~\citep{bhalla2026sparse}. However, some meaningful structure may still be representable by linear directions~\citep{park2023linear}. ConSDL therefore focuses on improving cross-seed consistency within this sparse linear assumption, rather than addressing whether all concepts can be represented linearly.

ConSDL replaces a single encoder pass with iterative per-sample sparse coding, increasing training cost relative to standard SAE baselines. We analyze training and inference costs in Appendix~\ref{sec:app:compute}. This additional computation reflects a trade-off between efficiency and improved consistency. Future work could reduce this overhead through more efficient sparse solvers. %
Furthermore, our experiments focus on relatively small foundation models because evaluating multiple methods across dictionary widths, sparsity, and random seeds is computationally demanding. We therefore do not establish whether the gains in cross-seed consistency extend to models with billions of parameters or substantially larger dictionaries. Evaluating ConSDL at larger scales remains an important direction for future work.

\section{Conclusion}
In this work, we introduce ConSDL, which revisits classical sparse coding to improve the cross-seed consistency over existing SAEs. By replacing the linear encoder with per-sample sparse coding and incorporating coefficient refitting and regularization against redundancy, ConSDL learns more consistent dictionaries and coefficients. It achieves substantially better ground-truth recovery across diverse synthetic settings. These benefits also transfer to foundation-model activations, where ConSDL improves cross-seed dictionary and coefficient consistency across multiple vision and language backbones, dictionary widths, and sparsity levels, while improving dictionary utilization. Future work will investigate scaling ConSDL to larger foundation models and extending it beyond linear feature representations.

\subsection*{AI use statement}
In this work, we used generative AI tools to assist with research ideation and experimental design, implementation of methods and experimental pipelines, retrieval and discovery of related work, drafting portions of the manuscript, and editing for English grammar and clarity. We did not use generative AI tools for other tasks requiring disclosure, and the remaining required-disclosure tasks were not applicable to this work. All AI-assisted research decisions, code, retrieved references, and manuscript text were reviewed, validated, and revised by the authors as appropriate. We take responsibility for the final content of this work, including text, claims, and artifacts produced with the aid of generative AI.

\subsection*{Ethics statement}
This work investigates the consistency of sparse representations learned from pretrained foundation models. It does not involve human subjects, personal data collection, or deployment in safety-critical settings. We are not aware of any ethical concerns specific to the proposed method beyond those associated with foundation-model interpretability research.

\subsection*{Reproducibility statement}
We provide detailed descriptions of the proposed method, experimental setup, evaluation protocols, and training hyperparameters in the main text and appendix. All datasets and pretrained foundation model checkpoints used in our experiments are publicly available and cited in the corresponding sections. We release our source code and experiment configurations at \href{https://github.com/dynamical-inference/consdl}{GitHub}, along with trained model weights at \href{https://huggingface.co/dynamical-inference/consdl}{Hugging Face}.

\bibliography{iclr2027_conference}
\bibliographystyle{iclr2027_conference}

\newpage
\appendix

\section*{Appendix}
\addcontentsline{toc}{section}{Appendix}

\startcontents[appendix]
\printcontents[appendix]{l}{1}{}

\newpage

\section{Extended Related Work}
\label{sec:appendix:extened_related_work}

\textbf{Improving SAE architectures and dictionaries.}
SAEs gained growing attention in mechanistic interpretability as their application to large language model activations demonstrated the potential to uncover large numbers of interpretable concepts~\citep{bricken2023monosemanticity}. Subsequent work initially focused on improving reconstruction while maintaining sparse activations. JumpReLU SAEs~\citep{rajamanoharan2024jumping} use learned activation thresholds to optimize sparsity more directly, while TopK SAEs~\citep{gao2025scaling} and BatchTopK SAEs~\citep{bussmann2024batchtopk} enforce sparsity through explicit feature selection. More recent work has shifted attention beyond reconstruction toward the quality and stability of the learned dictionary. SAEBench evaluates feature-level pathologies such as splitting and absorption~\citep{karvonen2025saebench}, while Matryoshka SAEs explicitly target these failures through nested dictionaries~\citep{bussmann2025learning}. MP-SAE~\citep{costa2026flat} instead predicts coefficients iteratively to better capture correlated concepts that may be missed by a single encoder pass.

\textbf{Ground-Truth Recovery in Synthetic Settings.}
The inconsistency of SAE dictionaries raises a more fundamental question of identifiability~\citep{nelson2026statistical}, namely whether SAEs recover the true underlying decomposition. This cannot be determined from real model activations because the ground truth is unavailable. Prior work has therefore used synthetic data with known dictionaries and sparse coefficients to measure recovery directly, and report substantial recovery failures.~\citep{korznikov2026sanity,o2024compute,pacela2026stop}.  However, as with prior consistency studies, the reported results are based on specific data distributions and training settings. It therefore remains unclear when SAE recovery succeeds and when it breaks down. We address this through a broader synthetic study spanning a range of difficulty levels.

\section{Implementation Details of ConSDL}
\label{sec:app:method_details}

We provide additional implementation details for the sparse coding, coefficient refitting, and co-active repulsion procedures described in Sections~\ref{sec:sparse_coding_refitting} and~\ref{sec:coactive_similarity}.

\textbf{FISTA optimization.}
Let $\mathbf{r}=\mathbf{x}-\mathbf{b}_{\mathrm{dec}}$. We use the
standard FISTA iterations \citep{beck2009fast}, initialized with
$\mathbf{a}_0=\mathbf{a}_{-1}=\mathbf{0}$ and $t_0=1$. The momentum
update is
$$
t_{k+1}
=
\frac{1+\sqrt{1+4t_k^2}}{2},
\qquad
\mathbf{y}_k
=
\mathbf{a}_k
+
\frac{t_k-1}{t_{k+1}}
(\mathbf{a}_k-\mathbf{a}_{k-1}).
$$
The proximal-gradient update is then
$$
\mathbf{u}_k
=
\mathbf{y}_k
-
\eta W_{\mathrm{dec}}^\top
\left(
W_{\mathrm{dec}}\mathbf{y}_k-\mathbf{r}
\right),
\qquad
\mathbf{a}_{k+1}
=
[\mathbf{u}_k-\eta\lambda]_+,
$$
where $\mathbf{r}=\mathbf{x}-\mathbf{b}_{\mathrm{dec}}$,
$\lambda$ controls the sparsity of the inferred coefficients, and
$[\cdot]_+$ denotes element-wise projection onto nonnegative values.

The step size is set to
$$
\eta
=
\frac{1}{\sigma_{\max}(W_{\mathrm{dec}})^2},
$$
where $\sigma_{\max}(W_{\mathrm{dec}})$ is estimated from the current
decoder using 32 power iterations. Since the decoder changes during
training, the step size is recomputed under the current dictionary.

\textbf{Coefficient refitting.}
Let $\mathbf{a}^{\mathrm{sc}}(\mathbf{x})$ denote the coefficient vector
returned by the FISTA solve above. Using its selected support
$S^{\mathrm{sc}}(\mathbf{x})$ defined in
Section~\ref{sec:sparse_coding_refitting}, we initialize
$\mathbf{a}^{(0)}=\mathbf{a}^{\mathrm{sc}}(\mathbf{x})$ and perform
projected gradient descent on the reconstruction objective without the
$\ell_1$ penalty. Each iteration computes
$$
\widetilde{\mathbf{a}}^{(q+1)}
=
\mathbf{a}^{(q)}
-
\eta W_{\mathrm{dec}}^\top
\left(
W_{\mathrm{dec}}\mathbf{a}^{(q)}-\mathbf{r}
\right),
$$
followed by
$$
a_j^{(q+1)}
=
\begin{cases}
\max(\widetilde{a}_j^{(q+1)},0),
& j\in S^{\mathrm{sc}}(\mathbf{x}),\\
0,
& j\notin S^{\mathrm{sc}}(\mathbf{x}).
\end{cases}
$$
We use the same step size $\eta$ as in the FISTA solve and perform 5--30 refitting iterations depending on the experimental setting. In implementation, a numerical threshold of $10^{-8}$ is used to determine the support returned by FISTA.

\begin{figure}[t]
\begin{pytorchcode}
def training_step(X, W_dec, b_dec):
    # Sparse coding
    A_sc = fista_sparse_code(
        X, W_dec, b_dec, lambda_sparse, T_fista
    )

    # Coefficient refitting
    support = A_sc > eps_support
    A_refit = refit_coefficients(
        X, W_dec, b_dec, A_sc, support, T_refit
    )
    A_refit = stop_gradient(A_refit)

    # Co-active similarity
    neighbors = nearest_coactive_pairs(
        W_dec.detach(), A_refit, shortlist_size=r
    )

    X_hat = A_refit @ W_dec.T + b_dec
    L_rec = reconstruction_loss(X, X_hat)
    L_sim = coactive_similarity_loss(
        W_dec, neighbors, margin=mu
    )

    # Dictionary update
    loss = L_rec + gamma_sim * L_sim
    loss.backward()
    optimizer.step()
    normalize_dictionary_atoms(W_dec)
\end{pytorchcode}
\caption{
Python pseudocode for one training step.
}
\label{fig:training_pseudocode}
\end{figure}

\textbf{Co-active similarity loss.}
Section~\ref{sec:coactive_similarity} presents the regularizer as a
squared margin penalty for clarity. In implementation, we normalize the
similarity above the margin and apply a logarithmic transformation. For each
matched co-active pair with $s_{i,j}>\mu$, we use
$$
\phi_{\mu}(s_{i,j})
=
-\log
\left[
1-
\frac{1}{1+\delta}
\left(
\frac{s_{i,j}-\mu}{1-\mu}
\right)^2
\right],
$$
where $\delta>0$ is a small constant that keeps the loss finite as
$s_{i,j}\rightarrow1$. We then compute
$$
\mathcal{L}_{\mathrm{sim}}
=
\frac{1}{|\mathcal{V}_{\mu}|}
\sum_{(i,j)\in\mathcal{V}_{\mu}}
\phi_{\mu}(s_{i,j}),
$$
where
$\mathcal{V}_{\mu}
=
\{(i,j):\mathbf{w}_j\in A_i,\ s_{i,j}>\mu\}$.
The normalization maps similarities between the margin $\mu$ and the maximum
similarity of $1$ to a common range, while the logarithmic transformation
increases the penalty as two atoms approach identical directions. When
$\mathcal{V}_{\mu}$ is empty, the similarity loss is set to zero.

\textbf{Co-active neighbor search.} The nearest-neighbor search in Section~\ref{sec:coactive_similarity} is restricted to the $r$ most similar decoder atoms for computational efficiency. At every training step, we compute the candidate neighbors from the current normalized decoder using cosine similarity and retain the top $r=32$ neighbors of each atom, excluding the atom itself. For each input and active atom, we keep only candidates that are also active for that input and select the candidate with the highest cosine similarity. If none of the candidate neighbors is co-active, that atom does not contribute to the repulsion loss. 

\textbf{Full algorithm.} Fig.~\ref{fig:training_pseudocode} provides Python code summarizing the complete training step and how the components described above are combined. The code also includes renormalization after each dictionary update to maintain the unit-norm constraint on the dictionary atoms throughout training.

\section{Computational Cost Analysis}

\begin{table*}[t]
\centering
\caption{
Per-token inference and training cost across SAE architectures and dictionary expansion factors. We report latency in $\mu$s with MFLOPs in parentheses, using matched sparsity budgets of $\ell_0=32$, $64$, and $128$ for $4\times$, $8\times$, and $16\times$, respectively.
}
\label{tab:compute_cost}

\setlength{\tabcolsep}{6pt}

\resizebox{\textwidth}{!}{%
\begin{tabular}{lrrrrrr}
\toprule
\multirow{2}{*}{\textbf{Architecture}} &
\multicolumn{2}{c}{$4\times$} &
\multicolumn{2}{c}{$8\times$} &
\multicolumn{2}{c}{$16\times$} \\
\cmidrule(lr){2-3}
\cmidrule(lr){4-5}
\cmidrule(lr){6-7}
&
\textbf{Inference} & \textbf{Training} &
\textbf{Inference} & \textbf{Training} &
\textbf{Inference} & \textbf{Training} \\
\midrule
\multicolumn{7}{l}{\textbf{Pythia-160M}} \\
\addlinespace[2pt]
ReLU & 0.04 (4.7) & 0.32 (28.3) & 0.08 (9.4) & 0.53 (56.6) & 0.14 (18.9) & 0.97 (113.2) \\
TopK & 0.12 (4.7) & 0.36 (28.3) & 0.21 (9.4) & 0.59 (56.6) & 0.37 (18.9) & 1.05 (113.2) \\
BatchTopK & 0.07 (4.7) & 0.48 (28.3) & 0.13 (9.4) & 0.75 (56.6) & 0.24 (18.9) & 1.40 (113.2) \\
Matryoshka & 0.07 (4.7) & 0.81 (33.0) & 0.13 (9.4) & 1.20 (66.1) & 0.24 (18.9) & 2.02 (132.1) \\
MP & 2.38 (302.0) & 2.23 (287.8) & 8.38 (1,208.0) & 3.78 (575.7) & 31.06 (4,831.8) & 7.14 (1,151.3) \\
iSAE & 0.13 (4.7) & 0.37 (28.3) & 0.22 (9.4) & 0.59 (56.6) & 0.39 (18.9) & 1.03 (113.2) \\
\textbf{Ours} & 2.28 (325.6) & 3.00 (338.6) & 4.26 (651.2) & 5.30 (684.2) & 8.22 (1,302.3) & 10.45 (1,396.7) \\
\addlinespace[4pt]
\midrule
\multicolumn{7}{l}{\textbf{CLIP ViT-B/16}} \\
\addlinespace[2pt]
ReLU & 0.03 (4.7) & 0.21 (28.3) & 0.07 (9.4) & 0.37 (56.6) & 0.14 (18.9) & 0.76 (113.2) \\
TopK & 0.10 (4.7) & 0.24 (28.3) & 0.18 (9.4) & 0.42 (56.6) & 0.35 (18.9) & 0.84 (113.2) \\
BatchTopK & 0.06 (4.7) & 0.31 (28.3) & 0.12 (9.4) & 0.58 (56.6) & 0.23 (18.9) & 1.18 (113.2) \\
Matryoshka & 0.06 (4.7) & 0.59 (33.0) & 0.12 (9.4) & 0.94 (66.1) & 0.23 (18.9) & 1.68 (132.1) \\
MP & 1.88 (302.0) & 1.61 (287.8) & 7.22 (1,208.0) & 2.89 (575.7) & 29.01 (4,831.8) & 6.15 (1,151.3) \\
iSAE & 0.11 (4.7) & 0.24 (28.3) & 0.19 (9.4) & 0.41 (56.6) & 0.36 (18.9) & 0.81 (113.2) \\
\textbf{Ours} & 2.03 (325.6) & 2.25 (335.6) & 4.00 (651.2) & 4.35 (672.3) & 8.04 (1,302.3) & 8.77 (1,349.3) \\
\bottomrule
\end{tabular}%
}
\end{table*}

\begin{table}[t]
\centering
\caption{
Breakdown of ConSDL training-step cost across dictionary expansion factors.
Values report latency in milliseconds per minibatch, with the fraction of total training time in parentheses.
}
\label{tab:compute_cost_ours}

\setlength{\tabcolsep}{4pt}
\resizebox{\columnwidth}{!}{
\begin{tabular}{llrrrrr}
\toprule
\textbf{Backbone} &
\textbf{Expansion} &
\shortstack{\textbf{Sparse}\\\textbf{coding}} &
\textbf{Refitting} &
\shortstack{\textbf{Similarity}\\\textbf{reg.}} &
\shortstack{\textbf{Other}\\\textbf{update}} &
\textbf{Total} \\
\midrule
\multirow{3}{*}{Pythia-160M}
& $4\times$  & 7.7 (62\%)   & 1.7 (14\%)  & 2.3 (19\%)  & 0.6 (5\%) & 12.3 \\
& $8\times$  & 14.3 (66\%)  & 3.2 (15\%)  & 3.5 (16\%)  & 0.7 (3\%) & 21.7 \\
& $16\times$ & 27.6 (64\%)  & 6.1 (14\%)  & 8.0 (19\%)  & 1.2 (3\%) & 42.8 \\
\midrule
\multirow{3}{*}{CLIP ViT-B/16}
& $4\times$  & 42.0 (74\%)  & 9.3 (16\%)  & 3.8 (7\%)   & 1.6 (3\%) & 56.7 \\
& $8\times$  & 82.6 (75\%)  & 18.2 (17\%) & 6.6 (6\%)   & 2.3 (2\%) & 109.7 \\
& $16\times$ & 166.2 (75\%) & 36.6 (17\%) & 13.9 (6\%)  & 4.4 (2\%) & 221.1 \\
\bottomrule
\end{tabular}
}
\end{table}

\label{sec:app:compute}

\textbf{Measurement setup.}
To compare the computational cost of ConSDL with existing SAE architectures, we measure per-token inference cost, complete training-step cost, and a component-wise breakdown of the training step. Measurements are performed on an NVIDIA H100 80GB GPU using PyTorch 2.2.2 with CUDA 12.1, reporting the median of 20 timed iterations after 5 warm-up iterations. We use matched sparsity budgets of $\ell_0=32$, $64$, and $128$ for $4\times$, $8\times$, and $16\times$ expansion, respectively. Each step processes 4,096 activation tokens for Pythia-160M and 25,216 activation tokens for CLIP ViT-B/16, corresponding to 128 images with 197 tokens each. All costs are normalized per token, and the frozen backbone forward pass is excluded because it is identical across SAE methods.

\textbf{Inference and training cost.}
Table~\ref{tab:compute_cost} compares the per-token test-time inference and training costs of ConSDL with existing SAE architectures across dictionary expansion factors. Because ConSDL performs sparse coding separately for each sample, it requires more computation than conventional SAEs that obtain sparse coefficients with a single encoder pass. MP-SAE also performs iterative computation during both training and inference. In our measurements, we use 10 matching-pursuit iterations during MP-SAE training, while at test time we continue the iterative selection until the target sparsity is reached. Since each MP-SAE iteration selects a latent based on the current residual, its test-time cost increases with sparsity and exceeds that of ConSDL at higher sparsity levels. ConSDL uses a fixed budget of $T=30$ FISTA iterations and $T_{\mathrm{refit}}=5$ refitting iterations across all expansion factors. Although ConSDL has higher training cost than MP-SAE and single-pass SAE variants, this additional computation represents a trade-off for substantially higher cross-seed consistency and better dictionary utilization.

\textbf{Compute cost breakdown.}
Table~\ref{tab:compute_cost_ours} breaks down the training cost of ConSDL into sparse coding, coefficient refitting, similarity regularization, and the remaining update. Sparse coding accounts for most of the computation, averaging approximately 69\% of the training step across settings. In contrast, coefficient refitting contributes only about 16\% on average, despite its substantial effect on consistency observed in the ablation study. Similarity regularization accounts for a smaller fraction of the overall cost, while the remaining decoding and optimization steps contribute only a few percent. The relative cost of each component also remains similar as the dictionary expansion increases, indicating that the additional computation of ConSDL is dominated by the fixed iterative sparse-coding procedure rather than by refitting or similarity regularization.

\section{Experimental Details}
\label{sec:app:exp}
This section provides experimental details, including baselines (Section~\ref{sec:app:baseline}) and evaluation metrics (Section~\ref{sec:app:evaluation_metrics}) shared across all experiments, followed by details of the synthetic settings (Section~\ref{sec:app:synthetic_settings}) and foundation model experiments (Section~\ref{sec:app:real_model}).
 
\subsection{Baselines}
\label{sec:app:baseline}

\textbf{Standard and sparsity-controlled SAEs.}
\textit{ReLU SAE}~\citep{cunningham2023sparse} produces nonnegative coefficients using a ReLU encoder and an $\ell_1$ sparsity penalty. \textit{TopK SAE}~\citep{gao2025scaling} removes the $\ell_1$ penalty and retains the $k$ largest encoder coefficients for each input. \textit{BatchTopK SAE}~\citep{bussmann2024batchtopk} performs this selection jointly across a minibatch, constraining the average number of active coefficients while allowing sparsity to vary across individual inputs. \textit{Matryoshka SAE}~\citep{bussmann2025learning} trains a single ordered dictionary by applying reconstruction objectives to multiple nested prefixes, so that shorter prefixes are also optimized to reconstruct the input rather than relying on the full dictionary~\citep{bussmann2025learning}. TopK, BatchTopK, and Matryoshka use the AuxK loss~\citep{gao2025scaling}, which provides learning signal to inactive latents by using them to explain the residual reconstruction error. We use an auxiliary-loss coefficient of $1/32$ and set $k_{\mathrm{aux}}=d_{\mathrm{in}}/2$.

\textbf{Iterative sparse encoding.}
\textit{MP-SAE}~\citep{costa2026flat} replaces single-pass coefficient inference with an iterative matching-pursuit procedure that repeatedly selects dictionary atoms based on the current reconstruction residual. We include MP-SAE as an alternative approach to iterative sparse coefficient inference.

\textbf{Consistency-focused SAE.}
\textit{iSAE}~\citep{nelson2026toward} modifies TopK using two components designed to improve dictionary and coefficient identifiability. First, it adopts AbsTopK~\citep{zhu2026abstopk}, which selects coefficients by magnitude while preserving their signs. This allows a single dictionary atom to contribute with either sign rather than requiring separate atoms for opposite signed contributions. Second, iSAE regularizes the decoder toward an approximate restricted-isometry property~\citep{candes2005decoding,nelson2026toward}, discouraging distinct sparse coefficient vectors from producing nearly identical reconstructions. We use an aRIP regularization weight of $0.1$

\textbf{Related baselines excluded from quantitative comparison.}
We also considered two closely related methods but exclude them from the quantitative comparisons because we could not obtain stable operating points under our experimental settings. \textit{Archetypal SAE} \citep{fel2025archetypal} is designed to improve cross-run dictionary consistency by first summarizing the activation distribution with K-means centroids and then learning dictionary atoms from these anchors.
\textit{DL-FISTA}~\citep{pacela2026stop} removes the amortized encoder and uses alternating optimization. The original study reports a severe loss of dictionary recovery as the latent dimension increases. We attempted to reproduce both methods across our evaluation settings but were unable to obtain sufficiently stable results for reliable comparison, and therefore omit them from the reported quantitative results.

\subsection{Evaluation metrics}
\label{sec:app:evaluation_metrics}

We evaluate ConSDL and SAE baselines in terms of input reconstruction, dictionary utilization, and cross-seed consistency.

\subsubsection{Reconstruction}
A useful sparse representation should retain enough information to reconstruct the original activation. We measure reconstruction quality using input $R^2$,
\[
R^2
=
1-
\frac{
    \sum_{n=1}^{N}
    \left\|
        \mathbf{x}_n-\hat{\mathbf{x}}_n
    \right\|_2^2
}{
    \sum_{n=1}^{N}
    \left\|
        \mathbf{x}_n-\bar{\mathbf{x}}
    \right\|_2^2
},
\]
where $\mathbf{x}_n$ and $\hat{\mathbf{x}}_n$ denote the original and
reconstructed activations, respectively, and $\bar{\mathbf{x}}$ denotes
the mean activation.

\subsubsection{Dictionary utilization}

\textbf{Dead atoms.} These represent dictionary capacity that is effectively unused. For atom $j$, we define its activation frequency as
\[
p_j
=
\frac{1}{N}
\sum_{n=1}^{N}
\mathbb{I}
\left[
    a_{nj} \neq 0
\right].
\]
We classify an atom as dead when $p_j < 10^{-6}$ and report the number of dead atoms.

\textbf{Duplicated atoms.}
These represent redundant dictionary capacity, with multiple atoms pointing in nearly identical decoder directions. Prior work measures duplication using pairwise decoder cosine similarity at a single threshold~\citep{mudide2025efficient}, making the result sensitive to the particular threshold and without explicitly identifying groups of mutually similar atoms. We instead exclude dead atoms and apply complete-linkage clustering to the remaining normalized decoder atoms over cosine-similarity thresholds $\tau \in \{0.6, 0.7, 0.8, 0.9\}$. Complete linkage ensures that every pair of atoms within a duplicate cluster satisfies the similarity criterion. At each threshold, we classify all atoms in clusters of size greater than one as duplicated and report the duplicated-atom count averaged across thresholds.

\textbf{Usable atom fraction.}
We summarize dictionary utilization as the fraction of atoms that are neither dead nor duplicated. Let $N_{\mathrm{dead}}$ denote the number of dead atoms and $\bar{N}_{\mathrm{dup}}$ the duplicated-atom count averaged across the thresholds above. We define
\[
\mathrm{Usable\ Atom\ Fraction}
=
1-
\frac{
N_{\mathrm{dead}}+\bar{N}_{\mathrm{dup}}
}{
m
}.
\]
Higher values indicate that more of the dictionary capacity is both active and non-redundant.

\subsubsection{cross-seed consistency}

\textbf{Decoder consistency.}
We measure cross seed decoder consistency using one-to-one Hungarian matching between normalized decoder atoms based on cosine similarity, following prior work~\citep{paulo2025sparse}. We refer to the resulting score as decoder cosine similarity (\texttt{DCS}). For each seed pair, \texttt{DCS} is the mean cosine similarity of the matched atoms,
\[
\operatorname{DCS}_{r,s}
=
\frac{1}{m}
\sum_{j=1}^{m}
\operatorname{cos}
\left(
\mathbf{d}_{j}^{(r)},
\mathbf{d}_{\pi(j)}^{(s)}
\right),
\]
where $\pi$ denotes the optimal Hungarian assignment. We report \texttt{DCS} averaged over all seed pairs.

\textbf{Coefficient consistency.}
We use the same decoder-based assignment to evaluate whether matched atoms exhibit similar coefficient patterns across inputs. For each matched atom pair, we compute the Pearson correlation coefficient between their coefficients across evaluation samples. We report the mean coefficient correlation (\texttt{MCC}), averaged over all matched atom pairs and then over all seed pairs. For computational efficiency, coefficient correlations are computed on a fixed subset of sampled tokens rather than the full evaluation set.

\subsubsection{Matching basis}

\begin{table}[t]
\centering
\caption{Three cross-seed consistency metrics on CLIP ViT-B/16
($16\times$, $\ell_0 \approx 128$, 3 seeds), each evaluated under two
matchings: the absolute decoder cosine Hungarian used throughout the
paper ($\pi_{\mathrm{dec}}$), and an independent Hungarian computed from
activation-magnitude footprints over the 50{,}000-image ImageNet
validation split ($\pi_{\mathrm{fp}}$). Higher is better and best per column
in bold. The ranking is unchanged across the two matchings.}
\label{tab:fp_axis_matching_basis}

\setlength{\tabcolsep}{6pt}
\begin{tabular}{l cc cc cc}
\toprule
\multirow{2}{*}{\textbf{Architecture}} &
\multicolumn{2}{c}{\shortstack{\textbf{Decoder cosine}\\\textbf{(DCS)}}} &
\multicolumn{2}{c}{\shortstack{\textbf{Footprint cosine}\\\textbf{(50k images)}}} &
\multicolumn{2}{c}{\shortstack{\textbf{Code Pearson}\\\textbf{(MCC)}}} \\
\cmidrule(lr){2-3} \cmidrule(lr){4-5} \cmidrule(lr){6-7}
& $\pi_{\mathrm{dec}}$ & $\pi_{\mathrm{fp}}$
& $\pi_{\mathrm{dec}}$ & $\pi_{\mathrm{fp}}$
& $\pi_{\mathrm{dec}}$ & $\pi_{\mathrm{fp}}$ \\
\midrule
ReLU & 0.525 & 0.496 & 0.595 & 0.615 & 0.417 & 0.416 \\
TopK & 0.558 & 0.525 & 0.696 & 0.715 & 0.499 & 0.498 \\
BatchTopK & 0.480 & 0.428 & 0.524 & 0.575 & 0.437 & 0.458 \\
Matryoshka & 0.271 & 0.189 & 0.243 & 0.290 & 0.184 & 0.199 \\
MP-SAE & 0.694 & 0.644 & 0.653 & 0.672 & 0.276 & 0.288 \\
iSAE & 0.683 & 0.658 & 0.624 & 0.634 & 0.588 & 0.588 \\
Ours & \textbf{0.780} & \textbf{0.769} & \textbf{0.838} & \textbf{0.844} & \textbf{0.744} & \textbf{0.745} \\
\bottomrule
\end{tabular}
\end{table}

\textbf{The cross-seed ranking is consistent under both geometric and
activation-based matching.} Throughout the paper we align dictionaries
across seeds with a Hungarian assignment on decoder cosine similarity ($\pi_{\mathrm{dec}}$). To verify that our conclusions are not
an artifact of this geometric choice, we construct a second, independent
assignment from activation footprints: for each latent we record its mean
activation magnitude on every image of the ImageNet validation split
(50K images, 985K tokens), yielding a footprint in $\mathbb{R}^{50\mathrm{K}}$,
and match seeds by footprint cosine similarity ($\pi_{\mathrm{fp}}$).
We then score three consistency metrics (\texttt{DCS}, \texttt{MCC}, and footprint
cosine similarity) on CLIP ViT-B/16 ($16\times $, $\ell_0 \approx 128$,
3 seeds) under both assignments. Table~\ref{tab:fp_axis_matching_basis}
shows that the two assignments induce the identical ranking for every
metric, and Ours ranks first in all six columns. Since
$\pi_{\mathrm{dec}}$ requires no forward passes over the dataset, we adopt
it as the default matching basis in the main text.

\subsection{Details of Synthetic experimental}
\label{sec:app:synthetic_settings}

This section provides additional details for Section~\ref{sec:preliminary}.

\textbf{Support and coefficient distributions.}
We write each ground-truth coefficient as $a_i=b_i c_i$, where $b_i\in{0,1}$ determines whether feature $i$ is active and $c_i>0$ its activation magnitude. In the default setting, supports are sampled independently as $b_i\sim\mathrm{Bernoulli}(p_i)$ with $p_i=k/m$. Consequently, the number of active features varies across observations, with expected sparsity $k=60$. Conditional on activation, coefficient magnitudes are sampled independently from $\mathrm{LogNormal}(\mu=0,\sigma=0.25)$.

\textbf{Long-tailed firing probabilities.}
To vary atom-frequency imbalance independently of the ground-truth dictionary, we randomly permute the atom indices, assign each atom a rank $r_i\in{1,\ldots,m}$, and set $p_i=\min(1,Cr_i^{-\beta})$, where $C$ normalizes the expected number of active atoms to $k=60$ before clipping. We use the four Zipf exponents specified in the main text. Supports are otherwise sampled independently, so this sweep changes marginal atom selection frequencies without introducing explicit co-occurrence correlations.

\textbf{Training details.}
We train each SAE for five epochs over a fixed pool of $10^6$ observations per configuration. Each configuration is trained with three random seeds, $\{0,1,2\}$. Unless otherwise stated, all models use a batch size of 128, a learning rate of $10^{-4}$, and 500 warmup steps followed by a constant learning-rate schedule. For TopK, BatchTopK, Matryoshka, and iSAE, we sweep $k \in \{60,90,120\}$ and report the best-performing setting for each synthetic configuration. MP-SAE uses 60 matching-pursuit iterations during both training and inference. For ReLU SAE and ConSDL, the sparsity-controlling coefficient is selected separately for each synthetic configuration to obtain comparable sparsity levels. Exact sparsity parameters and experiment-specific hyperparameters are reported in Table~\ref{tab:synthetic_hparams}.

\subsection{Details of Foundation Model Experiments}
\label{sec:app:real_model}

We provide additional details on model backbones and SAE training data (Section~\ref{sec:app:model_data_details}), training settings (Section~\ref{sec:app:training_details}), and sparse probing protocols (Section~\ref{sec:app:sparse_probing}).

\subsubsection{Model and Data Details}
\label{sec:app:model_data_details}

\textbf{Vision models.} For CLIP, we use \texttt{openai/clip-vit-base-patch16}
(\url{https://huggingface.co/openai/clip-vit-base-patch16}).
Its ViT-B/16 vision encoder has approximately 86M parameters, 12
Transformer blocks, a hidden dimension of 768, 12 attention heads, and
a patch size of $16\times16$ for $224\times224$ inputs. For DINOv2, we use
\texttt{facebook/dinov2-base}
(\url{https://huggingface.co/facebook/dinov2-base}), which is a ViT-B/14
model with approximately 86M parameters, 12 Transformer blocks, a hidden
dimension of 768, 12 attention heads, and a patch size of $14\times14$.

\textbf{Vision dataset.} For the vision data, we use
\texttt{evanarlian/imagenet\_1k\_resized\_256}
(\url{https://huggingface.co/datasets/evanarlian/imagenet_1k_resized_256}).
This release preserves the original ImageNet-1K splits while resizing
the shorter side of each image to 256 pixels. Its training split contains
1,281,167 images and validation split contains 50,000 images

\textbf{Language models.} We use the pretrained Pythia-160M checkpoint
\texttt{EleutherAI/pythia-160m}
(\url{https://huggingface.co/EleutherAI/pythia-160m}).
Pythia-160M contains 162M parameters and 12 Transformer layers, with a
hidden dimension of 768 and 12 attention heads. For GPT-2 Small, we use
\texttt{openai-community/gpt2}
(\url{https://huggingface.co/openai-community/gpt2}).
GPT-2 Small contains 124M parameters and 12 Transformer layers, with a
hidden dimension of 768, 12 attention heads, and a maximum context
length of 1024.

\textbf{Language dataset.} We use FineWeb
(\texttt{HuggingFaceFW/fineweb};
\url{https://huggingface.co/datasets/HuggingFaceFW/fineweb}),
a CommonCrawl-based corpus containing more than 18.5T tokens. For each
language backbone, we tokenize and cache a fixed subset of 500M tokens
using the corresponding model tokenizer. The same token set is reused
across random seeds, with only the sequence order shuffled between runs.

\subsubsection{Optimization and hyperparameters}
\label{sec:app:training_details}

\textbf{Training details.}
Unless otherwise stated, vision and language experiments use the same optimization settings. We use a learning rate of $3\times10^{-4}$ with 500 linear warmup steps, followed by a constant learning-rate schedule. We train all models with three random seeds, $\{0,1,2\}$, where the seed controls both parameter initialization and data ordering. Vision models use batches of 128 images, whereas language models use 4,096 tokens per optimization step.

\textbf{Architecture-specific settings.} 
For TopK, BatchTopK, Matryoshka, and iSAE, $k$ is selected according to the target sparsity. MP-SAE uses 10 or 32 matching-pursuit iterations during training. At inference time, matching pursuit is continued until the target $L_0$ is reached. For ReLU SAE and ConSDL, sparsity is controlled indirectly through the $\ell_1$ coefficient and sparse-coding penalty, respectively. We therefore consider operating points within $\pm15\%$ of the target $L_0$ and select the setting with the highest \texttt{DCS} $\times$ usable atom fraction within this range. Exact sparsity parameters and experiment-specific hyperparameters are reported in Table~\ref{tab:hparams}.

\section{Full Experimental Results}
\label{sec:app:additional_results}

This section provides full results for the synthetic experiments from Section~\ref{sec:preliminary} in Section~\ref{sec:app:synthetic_results}, and for the foundation model experiments from Section~\ref{sec:exp} in Section~\ref{sec:app:real_result}.

\subsection{Synthetic Results}
\label{sec:app:synthetic_results}

\begin{table*}[t]
\centering
\caption{
\textbf{Per-sample re-optimization improves coefficient recovery under a fixed dictionary.}
For each trained decoder, we compare amortized coefficients with per-sample FISTA re-optimization. For dictionaries with \texttt{DCS} $\ge 0.5$, re-optimized cells are shaded by the improvement over amortized inference, with darker \textcolor{green}{green} indicating larger gains up to $+0.2$. The gains are larger when dictionary recovery is stronger.
}

\label{tab:coeff_recovery_optimized}
\resizebox{\textwidth}{!}{%
\begin{tabular}{lll c cc cc cc}
\toprule
\multirow{2}{*}{\textbf{Architecture}} &
\multirow{2}{*}{$\beta$} &
\multirow{2}{*}{\textbf{\# Atoms}} &
\multirow{2}{*}{\textbf{DCS}} &
\multicolumn{2}{c}{\textbf{$\ell_0$}} &
\multicolumn{2}{c}{\textbf{MCC}} &
\multicolumn{2}{c}{\textbf{Input $R^2$}} \\
\cmidrule(lr){5-6}
\cmidrule(lr){7-8}
\cmidrule(lr){9-10}
&
&
&
&
Amortized &
Re-optimized &
Amortized &
Re-optimized &
Amortized &
Re-optimized \\
\midrule
\multirow{16}{*}{ReLU}
& 0.0 & 3K
& 0.983 & 67.4 & 63.8 & 0.833 & \cellcolor{green!46}0.975 & 0.659 & \cellcolor{green!65}0.885 \\
& 0.0 & 6K
& 0.970 & 76.5 & 75.6 & 0.780 & \cellcolor{green!59}0.962 & 0.633 & \cellcolor{green!65}0.872 \\
& 0.0 & 9K
& 0.777 & 62.5 & 92.8 & 0.555 & \cellcolor{green!38}0.672 & 0.434 & \cellcolor{green!49}0.585 \\
& 0.0 & 12K
& 0.367 & 86.9 & 70.6 & 0.157 & 0.168 & 0.268 & 0.249 \\
\cmidrule(lr){2-10}
& 0.5 & 3K
& 0.967 & 113 & 69.6 & 0.832 & \cellcolor{green!40}0.954 & 0.755 & \cellcolor{green!37}0.869 \\
& 0.5 & 6K
& 0.899 & 116 & 66.0 & 0.713 & \cellcolor{green!38}0.830 & 0.659 & \cellcolor{green!3}0.668 \\
& 0.5 & 9K
& 0.509 & 76.8 & 55.9 & 0.314 & \cellcolor{green!6}0.332 & 0.336 & 0.290 \\
& 0.5 & 12K
& 0.437 & 103 & 66.6 & 0.239 & 0.256 & 0.350 & 0.278 \\
\cmidrule(lr){2-10}
& 0.7 & 3K
& 0.702 & 120 & 69.2 & 0.564 & \cellcolor{green!13}0.605 & 0.512 & 0.476 \\
& 0.7 & 6K
& 0.369 & 73.0 & 107 & 0.204 & 0.223 & 0.176 & 0.305 \\
& 0.7 & 9K
& 0.359 & 110 & 61.5 & 0.183 & 0.197 & 0.260 & 0.180 \\
& 0.7 & 12K
& 0.270 & 101 & 68.5 & 0.102 & 0.112 & 0.175 & 0.154 \\
\cmidrule(lr){2-10}
& 1.0 & 3K
& 0.334 & 102 & 72.2 & 0.182 & 0.207 & -0.067 & -0.061 \\
& 1.0 & 6K
& 0.163 & 56.5 & 99.5 & 0.024 & 0.037 & -0.381 & -0.147 \\
& 1.0 & 9K
& 0.181 & 125 & 53.9 & 0.034 & 0.045 & -0.297 & -0.246 \\
& 1.0 & 12K
& 0.167 & 108 & 62.9 & 0.020 & 0.030 & -0.281 & -0.223 \\
\midrule
\multirow{16}{*}{TopK}
& 0.0 & 3K
& 0.982 & 60.0 & 60.0 & 0.862 & \cellcolor{green!28}0.947 & 0.812 & \cellcolor{green!35}0.920 \\
& 0.0 & 6K
& 0.810 & 60.0 & 60.0 & 0.611 & \cellcolor{green!19}0.669 & 0.646 & \cellcolor{green!17}0.699 \\
& 0.0 & 9K
& 0.227 & 60.0 & 60.0 & 0.048 & 0.048 & 0.403 & 0.427 \\
& 0.0 & 12K
& 0.181 & 60.0 & 60.0 & 0.023 & 0.023 & 0.400 & 0.423 \\
\cmidrule(lr){2-10}
& 0.5 & 3K
& 0.943 & 90.0 & 90.0 & 0.791 & \cellcolor{green!39}0.910 & 0.786 & \cellcolor{green!40}0.908 \\
& 0.5 & 6K
& 0.803 & 90.0 & 90.0 & 0.582 & \cellcolor{green!24}0.655 & 0.671 & \cellcolor{green!24}0.743 \\
& 0.5 & 9K
& 0.463 & 60.0 & 60.0 & 0.236 & 0.246 & 0.460 & 0.486 \\
& 0.5 & 12K
& 0.297 & 60.0 & 60.0 & 0.100 & 0.101 & 0.414 & 0.433 \\
\cmidrule(lr){2-10}
& 0.7 & 3K
& 0.886 & 90.0 & 90.0 & 0.716 & \cellcolor{green!33}0.816 & 0.686 & \cellcolor{green!41}0.813 \\
& 0.7 & 6K
& 0.695 & 90.0 & 90.0 & 0.462 & \cellcolor{green!16}0.512 & 0.553 & \cellcolor{green!29}0.643 \\
& 0.7 & 9K
& 0.512 & 90.0 & 90.0 & 0.268 & \cellcolor{green!6}0.285 & 0.484 & \cellcolor{green!22}0.551 \\
& 0.7 & 12K
& 0.354 & 60.0 & 60.0 & 0.146 & 0.149 & 0.352 & 0.402 \\
\cmidrule(lr){2-10}
& 1.0 & 3K
& 0.755 & 60.0 & 60.0 & 0.578 & \cellcolor{green!23}0.649 & 0.388 & \cellcolor{green!58}0.568 \\
& 1.0 & 6K
& 0.551 & 60.0 & 60.0 & 0.328 & \cellcolor{green!9}0.356 & 0.267 & \cellcolor{green!34}0.370 \\
& 1.0 & 9K
& 0.445 & 60.0 & 60.0 & 0.230 & 0.238 & 0.113 & 0.271 \\
& 1.0 & 12K
& 0.356 & 60.0 & 60.0 & 0.141 & 0.144 & 0.143 & 0.196 \\
\bottomrule
\end{tabular}%
}
\end{table*}

\textbf{Frozen dictionary coefficient re-optimization.}
To separate coefficient-inference error from dictionary-recovery error, we re-optimize coefficients while keeping each trained decoder fixed (Table~\ref{tab:coeff_recovery_optimized}). For each sample, we replace the amortized encoder output with coefficients obtained by running 50 FISTA iterations against the same learned dictionary, without updating the decoder. We then recompute coefficient recovery (\texttt{MCC}) and input reconstruction $R^2$. When the learned dictionary remains accurate, per-sample re-optimization substantially improves \texttt{MCC} and generally improves reconstruction, showing that the amortized encoder can leave recoverable coefficient information unused. As dictionary recovery degrades with larger dictionaries and more long-tailed atom selection frequencies, these gains become much smaller. Better coefficient inference alone therefore cannot compensate for an inaccurate dictionary, motivating improvements to the dictionary learning process itself.

\textbf{Full results of synthetic experiments.}
As a supplement to Fig.~\ref{fig:synthetic_heatmap}, we report the full synthetic results across all baselines, including input reconstruction, dictionary utilization (\# Dead and \# Dup.), ground-truth recovery, and cross-seed consistency, in Tables~\ref{tab:synthetic_3k}, \ref{tab:synthetic_6k}, \ref{tab:synthetic_9k}, and~\ref{tab:synthetic_12k}. Existing baselines generally show degraded ground-truth recovery and cross-seed consistency as the settings become more challenging, often accompanied by poorer dictionary utilization. In contrast, our method remains substantially more robust across the sweep, including the largest dictionaries and most strongly long-tailed activation distributions.

\subsection{Results on Foundation Model Activations}
\label{sec:app:real_result}

\begin{figure}[t]
\centering
\includegraphics[width=1.0\textwidth]{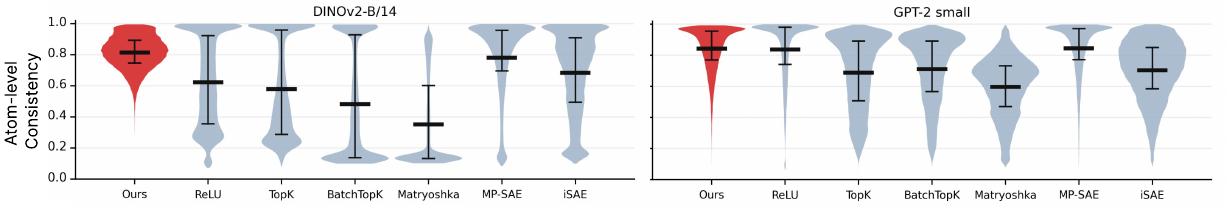}
\vspace{-2em}
\caption{\textbf{Atom-level consistency distributions for DINOv2-B/14 and GPT-2 Small.} Most ConSDL atoms exhibit high consistency in both backbones, whereas several baselines show broader distributions with a larger proportion of low-consistency atoms, particularly in DINOv2-B/14.}
\label{fig:consistency_dist_dino_gpt}
\end{figure}

\begin{figure}[t]
\centering
\includegraphics[width=1.0\textwidth]{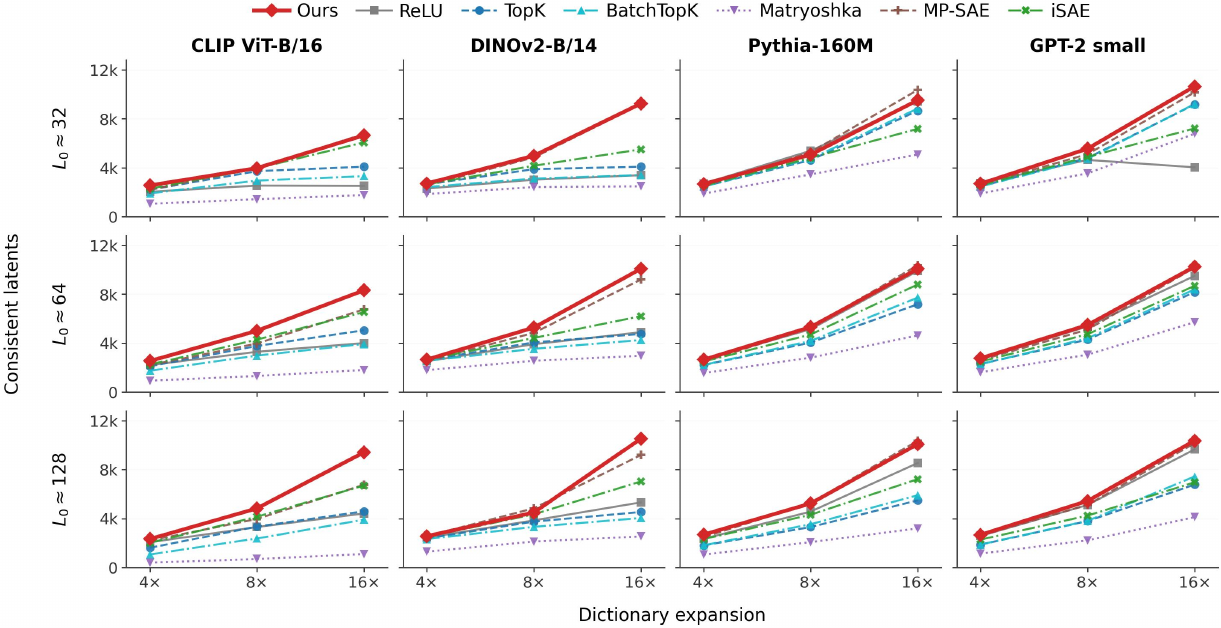}
\vspace{-2em}
\caption{\textbf{Number of consistent atoms across dictionary sizes and sparsity levels.}
We report the number of atoms with cross-seed consistency above 0.7 across four backbones, three dictionary expansions ($4\times$, $8\times$, and $16\times$), and three sparsity levels ($\ell_0 \approx 32, 64, 128$).
ConSDL consistently learns more consistent atoms as dictionary size increases, whereas most baselines exhibit slower growth or saturation, particularly in vision models. Higher sparsity levels generally yield smaller increases.}
\label{fig:num_consistent_latents}
\end{figure}

\textbf{Atom-level consistency distributions.}
We present consistency distributions for CLIP ViT-B/16 and Pythia-160M in Fig.~\ref{fig:viloin_plots_with_examples}A, with additional results for DINOv2-B/14 and GPT-2 Small in Fig.~\ref{fig:consistency_dist_dino_gpt}. Across all four backbones, ConSDL exhibits a greater concentration of high-consistency atoms than most baselines, particularly in vision models. This is also reflected in the number of atoms with consistency above 0.7 across dictionary sizes and sparsity levels (Fig.~\ref{fig:num_consistent_latents}).

\begin{figure}[t]
\centering
\includegraphics[width=1.0\textwidth]{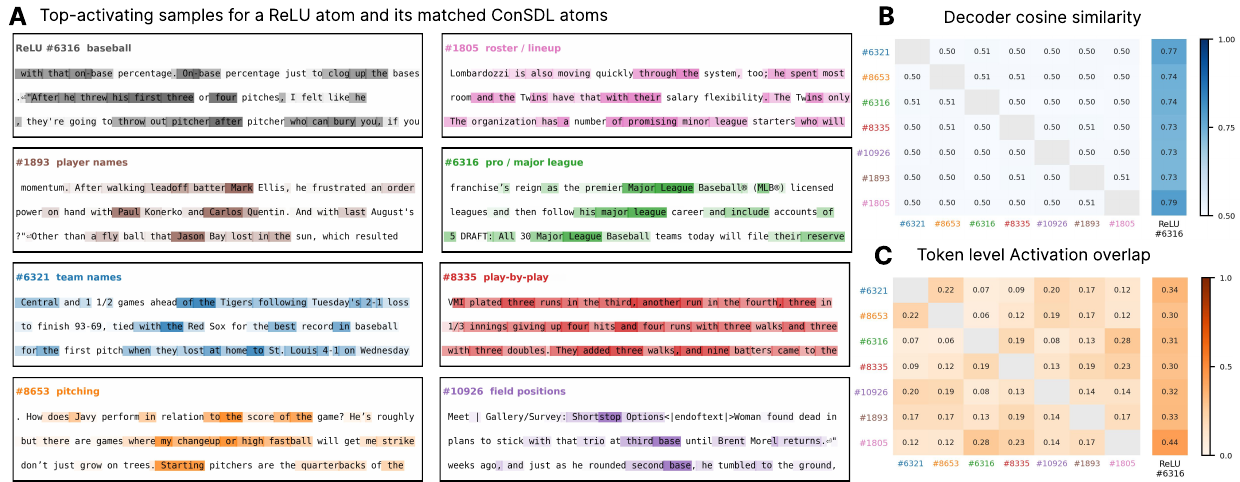}
\vspace{-2em}
\caption{\textbf{Finer-grained language concepts identified by ConSDL on Pythia-160M.}
\textbf{(A)} Top-activating text samples and token activation patterns for a ReLU ``baseball'' atom and seven matched ConSDL atoms capturing more specific concepts, including team names, pitching, and field positions. Concept labels were assigned by Claude Opus 5.5 based on top-activating samples.
\textbf{(B)} The matched ConSDL atoms have high decoder cosine similarity to the ReLU atom (0.73--0.79) but substantially lower pairwise similarity.
\textbf{(C)} The ConSDL atoms exhibit limited pairwise token-level activation overlap (0.06--0.28).}
\label{fig:ours_vs_baseline_language}
\end{figure}

\begin{figure}[t]
\centering
\includegraphics[width=1.0\textwidth]{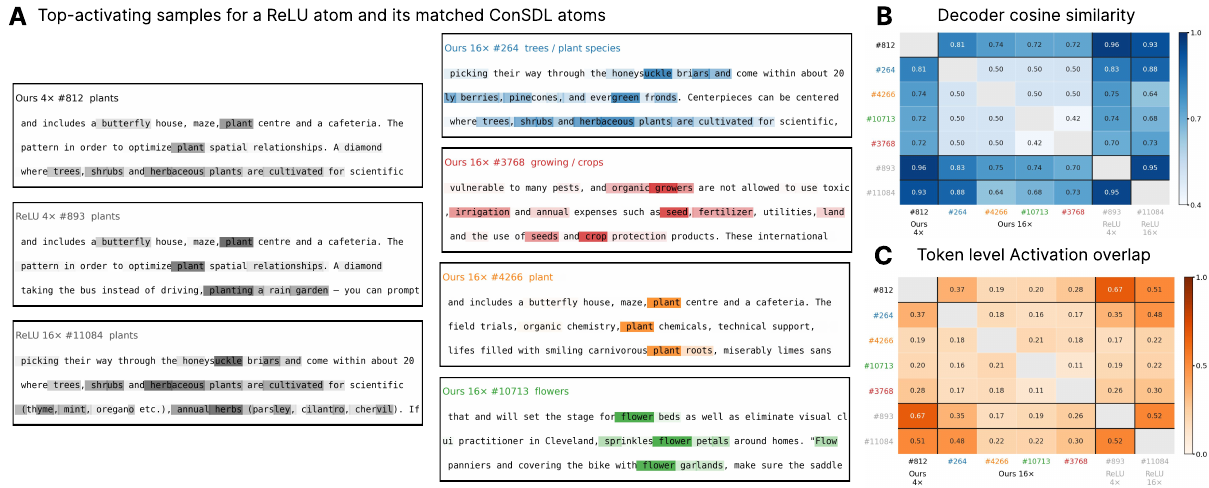}
\vspace{-2em}
\caption{\textbf{Finer-grained language concepts in larger ConSDL dictionaries.}
We compare Pythia-160M dictionaries at $4\times$ ($\ell_0 \approx 32$) and $16\times$ ($\ell_0 \approx 128$) expansion.
\textbf{(A)} A ConSDL atom associated with ``plants'' at $4\times$ matches four $16\times$ atoms capturing more specific concepts, including plant species, crops, and flowers. Concept labels were assigned by Claude Opus 5.5 based on top-activating samples.
\textbf{(B)} ReLU retains highly similar plant-related atoms across dictionary sizes (cosine similarity 0.95), whereas the matched ConSDL atoms exhibit low pairwise decoder similarity.
\textbf{(C)} The ConSDL $16\times$ atoms exhibit limited token-level activation overlap (0.11--0.21).}
\label{fig:x4_vs_x16_language}
\end{figure}

\textbf{Examples of finer-grained concepts in Pythia-160M.}
We observe similar patterns in Pythia-160M, extending our findings from CLIP ViT-B/16 (Figs.~\ref{fig:ours_vs_baseline} and \ref{fig:x4_vs_x16}). In the comparison between ConSDL and ReLU SAE (Fig.~\ref{fig:ours_vs_baseline_language}), a ReLU atom broadly responds to baseball-related text, whereas matched ConSDL atoms capture more specific information, including player and team names and gameplay descriptions. In the $4\times$ versus $16\times$ comparison (Fig.~\ref{fig:x4_vs_x16_language}), a ConSDL atom at $4\times$ matches several $16\times$ atoms associated with plant species, cultivation, and flowers, whereas ReLU retains highly similar plant-related atoms across dictionary sizes. For qualitative interpretation, we examine highly activating tokens and their surrounding contexts from the FineWeb evaluation corpus. We use Claude Opus 5.5 to assign descriptive concept labels based on these examples, relying on the surrounding context when peak activations occur on function words.

\begin{figure}[t]
\centering
\includegraphics[width=1.0\textwidth]{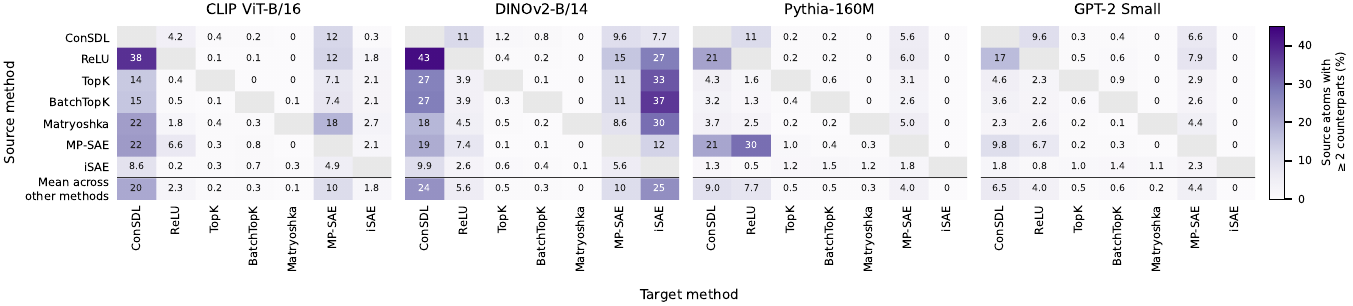}
\vspace{-2em}
\caption{\textbf{One-to-many atom correspondence across methods.}
Each cell shows the percentage of source atoms with at least two counterparts in the target dictionary. We restrict the analysis to atoms with cross-seed consistency $\geq 0.7$ and identify counterparts using decoder cosine similarity $\geq 0.7$. The bottom row reports the mean across the six other source methods.}
\label{fig:ours_vs_baseline_heatmap}
\end{figure}

\begin{figure}[t]
\centering
\includegraphics[width=1.0\textwidth]{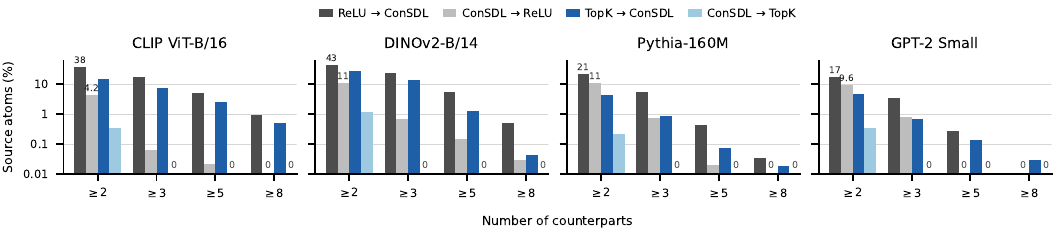}
\vspace{-2em}
\caption{\textbf{Distribution of cross-method atom correspondence.}
Percentage of source atoms with at least 2, 3, 5, or 8 counterparts in the target dictionary, comparing ConSDL with ReLU and TopK across four backbones. Only atoms with cross-seed consistency $\geq 0.7$ are considered, and counterparts are identified using decoder cosine similarity $\geq 0.7$. The vertical axis uses a logarithmic scale.}
\label{fig:ours_vs_baseline_dist}
\end{figure}

\textbf{One-to-many correspondence across methods.} To examine whether the finer-grained concepts observed in examples (Fig. \ref{fig:ours_vs_baseline} and Fig. \ref{fig:ours_vs_baseline_language}) reflect a broader pattern across vision and language models, we analyze one-to-many atom correspondence across methods (Fig.~\ref{fig:ours_vs_baseline_heatmap}). Across all four backbones, a greater fraction of consistent ReLU atoms have multiple ConSDL counterparts than vice versa. Averaged across source methods, ConSDL achieves the highest or joint-highest correspondence fraction across all four backbones. This asymmetry persists when requiring three, five, or eight counterparts, and similar trends are observed against TopK (Fig.~\ref{fig:ours_vs_baseline_dist}). Together with its low dictionary redundancy, these results suggest that ConSDL often represents features captured by a single baseline atom using multiple, potentially more specific atoms.

\begin{figure}[t]
\centering
\includegraphics[width=1.0\textwidth]{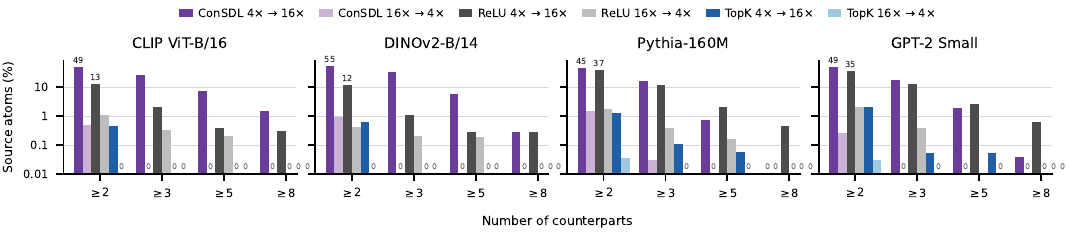}
\vspace{-2em}
\caption{\textbf{One-to-many atom correspondence across dictionary sizes.}
Percentage of source atoms with at least 2, 3, 5, or 8 counterparts when matching $4\times$ ($\ell_0 \approx 32$) and $16\times$ ($\ell_0 \approx 128$) dictionaries in both directions. The analysis includes only atoms with cross-seed consistency $\geq 0.7$, using decoder cosine similarity $\geq 0.7$ to identify counterparts. The vertical axis uses a logarithmic scale.}
\label{fig:x4_x16_dist}
\end{figure}

\textbf{One-to-many correspondence across dictionary sizes.}
While the preceding analysis compares different methods, we now examine atom correspondence between $4\times$ and $16\times$ dictionaries within each method (Fig.~\ref{fig:x4_x16_dist}), extending the qualitative observations in Fig.~\ref{fig:x4_vs_x16} and Fig.~\ref{fig:x4_vs_x16_language}. Across the four backbones, 45--55\% of consistent ConSDL atoms at $4\times$ have at least two consistent counterparts at $16\times$, compared with 12--37\% for ReLU and substantially lower fractions for TopK. This difference is particularly pronounced in vision, where ConSDL also exhibits higher correspondence fractions when requiring three or five counterparts. Given the low dictionary redundancy of ConSDL, these results support our qualitative observations that larger dictionaries capture multiple distinct but related features associated with broader concepts.

\begin{figure}[t]
\centering
\includegraphics[width=1.0\textwidth]{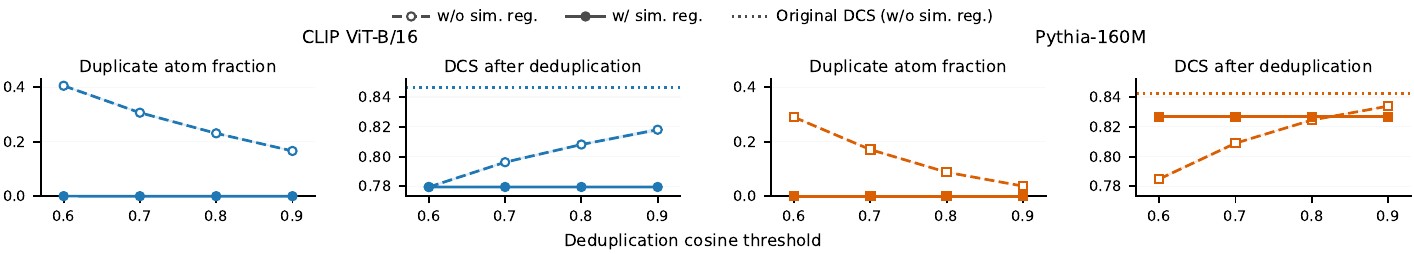}
\vspace{-2em}
\caption{\textbf{Effect of dictionary redundancy on cross-seed consistency.}
Duplicate atom fraction and DCS after deduplication across cosine similarity thresholds for CLIP ViT-B/16 and Pythia-160M. Dotted horizontal lines indicate DCS before deduplication without similarity regularization.}
\label{fig:effect_of_dup}
\end{figure}

\textbf{Dictionary consistency and redundancy.} In the ablation study (Table \ref{tab:ablation_x16}, removing similarity regularization yields higher DCS but also substantially more duplicate atoms. To examine how these duplicates affect measured consistency, we recompute DCS after deduplication across cosine similarity thresholds from 0.6 to 0.9 (Fig.~\ref{fig:effect_of_dup}). Without regularization, deduplication at a threshold of 0.6 removes 41\% and 29\% of atoms in CLIP ViT-B/16 and Pythia-160M, reducing DCS from 0.85 to 0.78 and from 0.84 to 0.79, respectively. As the threshold increases, fewer atoms are removed and DCS approaches its original value. In contrast, ConSDL has no identified duplicate atoms across the tested thresholds while maintaining high DCS. These findings suggest that duplicate atoms can inflate DCS, highlighting the importance of evaluating cross-seed consistency alongside dictionary redundancy. We therefore also report the usable atom fraction, which accounts for both dead and duplicate atoms, throughout our evaluation.

\begin{figure}[t]
\centering
\includegraphics[width=1.0\textwidth]{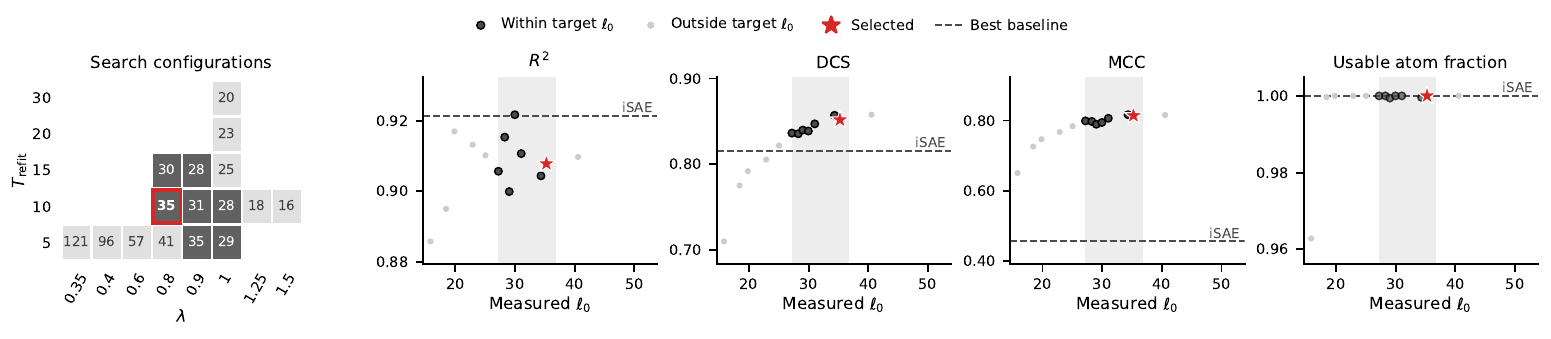}
\vspace{-2em}
\caption{\textbf{Hyperparameter sensitivity of ConSDL.}
Results on CLIP ViT-B/16 ($4\times$, $\ell_0 \approx 32$) across sparsity penalties $\lambda$ and refitting iterations $T_{\mathrm{refit}}$. The left panel shows measured sparsity, while the remaining panels compare performance. Dark markers and gray shading indicate the target sparsity range, red highlights mark the selected configuration, and dashed lines show the best baseline.}
\label{fig:hyperparam_sensitivity}
\end{figure}

\textbf{Hyperparameter sensitivity.} To assess whether the consistency gains depend on a particular hyperparameter choice, we vary the sparsity penalty $\lambda$ and the number of refitting iterations $T_{\mathrm{refit}}$, while keeping the FISTA iterations and similarity regularization strength fixed (Fig.~\ref{fig:hyperparam_sensitivity}). Multiple configurations within the target sparsity range outperform the strongest baseline in both DCS and MCC while maintaining nearly full dictionary utilization. Although reconstruction quality varies across configurations and does not consistently exceed the best baseline, the improvements in dictionary and coefficient consistency are not limited to the selected configuration.

\begin{table*}[t]
\centering
\caption{
\textbf{Cross-seed consistency and dictionary utilization across
foundation-model backbones.}
For each backbone and method, we report the mean and standard
deviation across nine combinations of dictionary expansion
($4\times$, $8\times$, $16\times$) and sparsity
($\ell_0 \approx 32$, $64$, $128$).
The variation therefore reflects differences across these settings
rather than training seeds.
The last two columns report the relative improvement of ConSDL over
the average of the six SAE baselines and the strongest baseline for
each backbone and metric.
The final three rows report averages over the four backbones with
equal weighting.
}
\label{tab:summary_consistency}

\setlength{\tabcolsep}{3.2pt}

\resizebox{\textwidth}{!}{%
\begin{tabular}{ll cccccc c cc}
\toprule
\textbf{Backbone} &
\textbf{Metric} &
\textbf{ReLU} & \textbf{TopK} & \textbf{BatchTopK} & \textbf{Matryoshka} & \textbf{MP-SAE} & \textbf{iSAE} & \textbf{ConSDL} &
\textbf{vs.\ Avg.} &
\textbf{vs.\ Best} \\
\midrule
\multirow{3}{*}{CLIP ViT-B/16}
& DCS & 0.61\,{\scriptsize $\pm$0.15} & 0.67\,{\scriptsize $\pm$0.10} & 0.56\,{\scriptsize $\pm$0.10} & 0.37\,{\scriptsize $\pm$0.08} & 0.75\,{\scriptsize $\pm$0.04} & 0.75\,{\scriptsize $\pm$0.06} & \textbf{0.78\,{\scriptsize $\pm$0.07}} & +26.7\% & +4.8\% \\
& MCC & 0.53\,{\scriptsize $\pm$0.17} & 0.63\,{\scriptsize $\pm$0.13} & 0.52\,{\scriptsize $\pm$0.11} & 0.31\,{\scriptsize $\pm$0.10} & 0.45\,{\scriptsize $\pm$0.15} & 0.40\,{\scriptsize $\pm$0.06} & \textbf{0.73\,{\scriptsize $\pm$0.09}} & +54.7\% & +16.8\% \\
& Usable atom frac. & 0.99\,{\scriptsize $\pm$0.00} & 1.00\,{\scriptsize $\pm$0.00} & 1.00\,{\scriptsize $\pm$0.00} & 1.00\,{\scriptsize $\pm$0.00} & 0.85\,{\scriptsize $\pm$0.05} & 0.96\,{\scriptsize $\pm$0.10} & \textbf{1.00\,{\scriptsize $\pm$0.00}} & +3.5\% & +0.0\% \\
\midrule
\multirow{3}{*}{DINOv2-B/14}
& DCS & 0.68\,{\scriptsize $\pm$0.16} & 0.73\,{\scriptsize $\pm$0.13} & 0.65\,{\scriptsize $\pm$0.15} & 0.50\,{\scriptsize $\pm$0.12} & 0.83\,{\scriptsize $\pm$0.04} & 0.77\,{\scriptsize $\pm$0.10} & \textbf{0.84\,{\scriptsize $\pm$0.05}} & +21.4\% & +1.6\% \\
& MCC & 0.61\,{\scriptsize $\pm$0.19} & 0.69\,{\scriptsize $\pm$0.17} & 0.61\,{\scriptsize $\pm$0.19} & 0.47\,{\scriptsize $\pm$0.16} & 0.49\,{\scriptsize $\pm$0.16} & 0.44\,{\scriptsize $\pm$0.07} & \textbf{0.78\,{\scriptsize $\pm$0.06}} & +42.3\% & +14.1\% \\
& Usable atom frac. & 0.98\,{\scriptsize $\pm$0.01} & 0.98\,{\scriptsize $\pm$0.04} & \textbf{1.00\,{\scriptsize $\pm$0.00}} & 0.99\,{\scriptsize $\pm$0.01} & 0.75\,{\scriptsize $\pm$0.11} & 0.99\,{\scriptsize $\pm$0.01} & 0.99\,{\scriptsize $\pm$0.02} & +4.5\% & -0.6\% \\
\midrule
\multirow{3}{*}{Pythia-160M}
& DCS & 0.85\,{\scriptsize $\pm$0.04} & 0.74\,{\scriptsize $\pm$0.07} & 0.75\,{\scriptsize $\pm$0.06} & 0.62\,{\scriptsize $\pm$0.05} & \textbf{0.85\,{\scriptsize $\pm$0.01}} & 0.78\,{\scriptsize $\pm$0.05} & 0.85\,{\scriptsize $\pm$0.02} & +10.8\% & -0.8\% \\
& MCC & 0.76\,{\scriptsize $\pm$0.05} & 0.69\,{\scriptsize $\pm$0.06} & 0.70\,{\scriptsize $\pm$0.05} & 0.61\,{\scriptsize $\pm$0.04} & 0.46\,{\scriptsize $\pm$0.04} & 0.43\,{\scriptsize $\pm$0.04} & \textbf{0.77\,{\scriptsize $\pm$0.03}} & +26.0\% & +1.1\% \\
& Usable atom frac. & 0.93\,{\scriptsize $\pm$0.04} & 1.00\,{\scriptsize $\pm$0.00} & 1.00\,{\scriptsize $\pm$0.00} & \textbf{1.00\,{\scriptsize $\pm$0.00}} & 0.65\,{\scriptsize $\pm$0.05} & 0.97\,{\scriptsize $\pm$0.08} & 1.00\,{\scriptsize $\pm$0.00} & +8.1\% & -0.1\% \\
\midrule
\multirow{3}{*}{GPT-2 Small}
& DCS & 0.81\,{\scriptsize $\pm$0.14} & 0.77\,{\scriptsize $\pm$0.05} & 0.77\,{\scriptsize $\pm$0.05} & 0.65\,{\scriptsize $\pm$0.04} & 0.86\,{\scriptsize $\pm$0.01} & 0.78\,{\scriptsize $\pm$0.05} & \textbf{0.88\,{\scriptsize $\pm$0.02}} & +13.7\% & +2.3\% \\
& MCC & 0.75\,{\scriptsize $\pm$0.16} & 0.74\,{\scriptsize $\pm$0.03} & 0.75\,{\scriptsize $\pm$0.02} & 0.66\,{\scriptsize $\pm$0.02} & 0.56\,{\scriptsize $\pm$0.07} & 0.46\,{\scriptsize $\pm$0.03} & \textbf{0.82\,{\scriptsize $\pm$0.03}} & +25.8\% & +9.6\% \\
& Usable atom frac. & 0.87\,{\scriptsize $\pm$0.18} & 0.99\,{\scriptsize $\pm$0.01} & 0.99\,{\scriptsize $\pm$0.01} & \textbf{1.00\,{\scriptsize $\pm$0.00}} & 0.79\,{\scriptsize $\pm$0.03} & 0.99\,{\scriptsize $\pm$0.04} & 1.00\,{\scriptsize $\pm$0.00} & +6.5\% & -0.1\% \\
\midrule
\multirow{3}{*}{\textbf{All backbones}}
& \textbf{DCS} & 0.74 & 0.73 & 0.68 & 0.53 & 0.82 & 0.77 & \textbf{0.84} & \textbf{+17.6\%} & \textbf{+1.9\%} \\
& \textbf{MCC} & 0.66 & 0.69 & 0.65 & 0.51 & 0.49 & 0.43 & \textbf{0.78} & \textbf{+35.9\%} & \textbf{+13.3\%} \\
& \textbf{Usable atom frac.} & 0.94 & 0.99 & 1.00 & 1.00 & 0.76 & 0.98 & \textbf{1.00} & \textbf{+5.6\%} & \textbf{+0.0\%} \\
\bottomrule
\end{tabular}%
}
\end{table*}

\textbf{Full results on foundation-model activations.}
We report complete results across three dictionary expansion factors ($4\times$, $8\times$, and $16\times$) and three matched sparsity levels ($\ell_0 \approx 32$, $64$, and $128$). For each configuration, we evaluate reconstruction quality ($R^2$), cross-seed dictionary consistency (\texttt{DCS}), coefficient consistency (\texttt{MCC}), and usable atom fraction. Detailed results are provided in Tables~\ref{tab:appendix_clip_full}, \ref{tab:appendix_dinov2_full}, \ref{tab:appendix_gpt2_full}, and \ref{tab:appendix_pythia_full} for CLIP ViT-B/16, DINOv2 ViT-B/14, GPT-2 Small, and Pythia-160M, respectively. Table~\ref{tab:summary_consistency} provides a summary.

\begin{table*}[t]
\centering
\caption{
Sparse probing of vision dictionaries.
AUPRC from the $k$ most discriminative dictionary atoms per class, $k \in \{1, 5\}$,
at each expansion and its matched sparsity
($4\times$ at $\ell_0 \approx 32$, $8\times$ at $\ell_0 \approx 64$, $16\times$ at $\ell_0 \approx 128$).
Avg.\ is the unweighted mean over the six evaluation datasets.
$\pm$ is the standard deviation across training seeds of the reported score. 
}
\label{tab:vision_probing_main}

\setlength{\tabcolsep}{2.5pt}

\resizebox{\textwidth}{!}{%
\begin{tabular}{lll cc cc cc cc cc cc cc}
\toprule
\multirow{2}{*}{\textbf{Backbone}} &
\multirow{2}{*}{\textbf{Cell}} &
\multirow{2}{*}{\textbf{Method}} &
\multicolumn{2}{c}{\textbf{ImageNet}\,\textit{\footnotesize Object}} &
\multicolumn{2}{c}{\textbf{SUN397}\,\textit{\footnotesize Scene}} &
\multicolumn{2}{c}{\textbf{Food101}\,\textit{\footnotesize Food}} &
\multicolumn{2}{c}{\textbf{DTD}\,\textit{\footnotesize Texture}} &
\multicolumn{2}{c}{\textbf{Caltech256}\,\textit{\footnotesize Object}} &
\multicolumn{2}{c}{\textbf{RAF-DB}\,\textit{\footnotesize Emotion}} &
\multicolumn{2}{c}{\textbf{Avg.}} \\
\cmidrule(lr){4-5}
\cmidrule(lr){6-7}
\cmidrule(lr){8-9}
\cmidrule(lr){10-11}
\cmidrule(lr){12-13}
\cmidrule(lr){14-15}
\cmidrule(lr){16-17}
&
&
&
@1 & @5 &
@1 & @5 &
@1 & @5 &
@1 & @5 &
@1 & @5 &
@1 & @5 &
@1 & @5 \\
\midrule
\multirow{21}{*}{CLIP ViT-B/16}
& \multirow{7}{*}{\shortstack{$4\times$\\{\footnotesize $\ell_0 \approx 32$}}} & ReLU
& \textbf{0.33}\,{\tiny $\pm$0.00} & 0.39\,{\tiny $\pm$0.00} & 0.35\,{\tiny $\pm$0.00} & 0.43\,{\tiny $\pm$0.00} & 0.42\,{\tiny $\pm$0.01} & 0.46\,{\tiny $\pm$0.00} & 0.51\,{\tiny $\pm$0.01} & 0.61\,{\tiny $\pm$0.01} & 0.70\,{\tiny $\pm$0.00} & 0.77\,{\tiny $\pm$0.00} & 0.45\,{\tiny $\pm$0.04} & 0.51\,{\tiny $\pm$0.01} & 0.46\,{\tiny $\pm$0.01} & 0.53\,{\tiny $\pm$0.00} \\
&  & TopK
& 0.28\,{\tiny $\pm$0.00} & 0.36\,{\tiny $\pm$0.00} & 0.34\,{\tiny $\pm$0.00} & \textbf{0.45}\,{\tiny $\pm$0.00} & 0.42\,{\tiny $\pm$0.00} & \textbf{0.50}\,{\tiny $\pm$0.00} & 0.49\,{\tiny $\pm$0.01} & \textbf{0.61}\,{\tiny $\pm$0.00} & 0.67\,{\tiny $\pm$0.00} & 0.76\,{\tiny $\pm$0.00} & \textbf{0.48}\,{\tiny $\pm$0.00} & \textbf{0.52}\,{\tiny $\pm$0.01} & 0.45\,{\tiny $\pm$0.00} & \textbf{0.54}\,{\tiny $\pm$0.00} \\
&  & BatchTopK
& 0.28\,{\tiny $\pm$0.00} & 0.36\,{\tiny $\pm$0.00} & 0.34\,{\tiny $\pm$0.00} & 0.45\,{\tiny $\pm$0.00} & 0.42\,{\tiny $\pm$0.01} & 0.50\,{\tiny $\pm$0.00} & 0.49\,{\tiny $\pm$0.01} & 0.61\,{\tiny $\pm$0.00} & 0.67\,{\tiny $\pm$0.00} & 0.76\,{\tiny $\pm$0.00} & 0.46\,{\tiny $\pm$0.04} & 0.51\,{\tiny $\pm$0.01} & 0.44\,{\tiny $\pm$0.01} & 0.53\,{\tiny $\pm$0.00} \\
&  & Matryoshka
& 0.26\,{\tiny $\pm$0.00} & 0.33\,{\tiny $\pm$0.00} & 0.31\,{\tiny $\pm$0.00} & 0.42\,{\tiny $\pm$0.00} & 0.39\,{\tiny $\pm$0.01} & 0.46\,{\tiny $\pm$0.01} & 0.48\,{\tiny $\pm$0.01} & 0.59\,{\tiny $\pm$0.00} & 0.61\,{\tiny $\pm$0.01} & 0.71\,{\tiny $\pm$0.01} & 0.42\,{\tiny $\pm$0.04} & 0.49\,{\tiny $\pm$0.00} & 0.41\,{\tiny $\pm$0.01} & 0.50\,{\tiny $\pm$0.00} \\
&  & MP-SAE
& 0.28\,{\tiny $\pm$0.00} & 0.36\,{\tiny $\pm$0.00} & 0.34\,{\tiny $\pm$0.00} & 0.45\,{\tiny $\pm$0.00} & 0.42\,{\tiny $\pm$0.00} & 0.50\,{\tiny $\pm$0.01} & 0.51\,{\tiny $\pm$0.01} & 0.60\,{\tiny $\pm$0.01} & 0.67\,{\tiny $\pm$0.00} & 0.76\,{\tiny $\pm$0.00} & 0.42\,{\tiny $\pm$0.03} & 0.50\,{\tiny $\pm$0.02} & 0.44\,{\tiny $\pm$0.00} & 0.53\,{\tiny $\pm$0.00} \\
&  & iSAE
& 0.15\,{\tiny $\pm$0.00} & 0.29\,{\tiny $\pm$0.00} & 0.17\,{\tiny $\pm$0.01} & 0.37\,{\tiny $\pm$0.00} & 0.25\,{\tiny $\pm$0.04} & 0.41\,{\tiny $\pm$0.01} & 0.25\,{\tiny $\pm$0.02} & 0.58\,{\tiny $\pm$0.00} & 0.36\,{\tiny $\pm$0.02} & 0.68\,{\tiny $\pm$0.00} & 0.24\,{\tiny $\pm$0.05} & 0.42\,{\tiny $\pm$0.01} & 0.24\,{\tiny $\pm$0.02} & 0.46\,{\tiny $\pm$0.00} \\
&  & ConSDL
& 0.32\,{\tiny $\pm$0.00} & 0.39\,{\tiny $\pm$0.00} & \textbf{0.36}\,{\tiny $\pm$0.00} & 0.44\,{\tiny $\pm$0.00} & \textbf{0.43}\,{\tiny $\pm$0.00} & 0.46\,{\tiny $\pm$0.00} & \textbf{0.52}\,{\tiny $\pm$0.01} & 0.61\,{\tiny $\pm$0.00} & \textbf{0.72}\,{\tiny $\pm$0.00} & \textbf{0.77}\,{\tiny $\pm$0.00} & 0.48\,{\tiny $\pm$0.01} & 0.51\,{\tiny $\pm$0.02} & \textbf{0.47}\,{\tiny $\pm$0.00} & 0.53\,{\tiny $\pm$0.00} \\
\cmidrule(lr){2-17}
& \multirow{7}{*}{\shortstack{$8\times$\\{\footnotesize $\ell_0 \approx 64$}}} & ReLU
& \textbf{0.36}\,{\tiny $\pm$0.00} & \textbf{0.40}\,{\tiny $\pm$0.00} & 0.37\,{\tiny $\pm$0.00} & 0.45\,{\tiny $\pm$0.00} & \textbf{0.46}\,{\tiny $\pm$0.01} & 0.50\,{\tiny $\pm$0.00} & 0.51\,{\tiny $\pm$0.00} & 0.61\,{\tiny $\pm$0.00} & 0.72\,{\tiny $\pm$0.00} & 0.79\,{\tiny $\pm$0.00} & 0.41\,{\tiny $\pm$0.04} & 0.53\,{\tiny $\pm$0.00} & 0.47\,{\tiny $\pm$0.01} & 0.55\,{\tiny $\pm$0.00} \\
&  & TopK
& 0.29\,{\tiny $\pm$0.00} & 0.36\,{\tiny $\pm$0.00} & 0.35\,{\tiny $\pm$0.00} & 0.45\,{\tiny $\pm$0.00} & 0.44\,{\tiny $\pm$0.00} & 0.52\,{\tiny $\pm$0.00} & 0.48\,{\tiny $\pm$0.01} & 0.61\,{\tiny $\pm$0.00} & 0.68\,{\tiny $\pm$0.00} & 0.76\,{\tiny $\pm$0.00} & 0.47\,{\tiny $\pm$0.02} & 0.51\,{\tiny $\pm$0.01} & 0.45\,{\tiny $\pm$0.00} & 0.54\,{\tiny $\pm$0.00} \\
&  & BatchTopK
& 0.29\,{\tiny $\pm$0.00} & 0.36\,{\tiny $\pm$0.00} & 0.35\,{\tiny $\pm$0.00} & 0.44\,{\tiny $\pm$0.00} & 0.43\,{\tiny $\pm$0.00} & 0.51\,{\tiny $\pm$0.00} & 0.50\,{\tiny $\pm$0.00} & 0.60\,{\tiny $\pm$0.01} & 0.68\,{\tiny $\pm$0.00} & 0.76\,{\tiny $\pm$0.00} & \textbf{0.48}\,{\tiny $\pm$0.01} & 0.50\,{\tiny $\pm$0.00} & 0.45\,{\tiny $\pm$0.00} & 0.53\,{\tiny $\pm$0.00} \\
&  & Matryoshka
& 0.26\,{\tiny $\pm$0.00} & 0.32\,{\tiny $\pm$0.00} & 0.33\,{\tiny $\pm$0.00} & 0.41\,{\tiny $\pm$0.00} & 0.42\,{\tiny $\pm$0.00} & 0.48\,{\tiny $\pm$0.00} & 0.49\,{\tiny $\pm$0.00} & 0.59\,{\tiny $\pm$0.00} & 0.62\,{\tiny $\pm$0.00} & 0.71\,{\tiny $\pm$0.00} & 0.43\,{\tiny $\pm$0.01} & 0.47\,{\tiny $\pm$0.01} & 0.42\,{\tiny $\pm$0.00} & 0.50\,{\tiny $\pm$0.00} \\
&  & MP-SAE
& 0.34\,{\tiny $\pm$0.00} & 0.39\,{\tiny $\pm$0.00} & 0.37\,{\tiny $\pm$0.00} & \textbf{0.46}\,{\tiny $\pm$0.00} & 0.44\,{\tiny $\pm$0.00} & \textbf{0.54}\,{\tiny $\pm$0.00} & 0.52\,{\tiny $\pm$0.00} & \textbf{0.61}\,{\tiny $\pm$0.00} & 0.72\,{\tiny $\pm$0.00} & \textbf{0.79}\,{\tiny $\pm$0.00} & 0.40\,{\tiny $\pm$0.01} & 0.51\,{\tiny $\pm$0.00} & 0.46\,{\tiny $\pm$0.00} & \textbf{0.55}\,{\tiny $\pm$0.00} \\
&  & iSAE
& 0.17\,{\tiny $\pm$0.00} & 0.29\,{\tiny $\pm$0.00} & 0.18\,{\tiny $\pm$0.01} & 0.38\,{\tiny $\pm$0.00} & 0.21\,{\tiny $\pm$0.03} & 0.45\,{\tiny $\pm$0.00} & 0.25\,{\tiny $\pm$0.02} & 0.58\,{\tiny $\pm$0.00} & 0.33\,{\tiny $\pm$0.02} & 0.69\,{\tiny $\pm$0.00} & 0.27\,{\tiny $\pm$0.05} & 0.47\,{\tiny $\pm$0.01} & 0.23\,{\tiny $\pm$0.01} & 0.48\,{\tiny $\pm$0.00} \\
&  & ConSDL
& 0.36\,{\tiny $\pm$0.00} & 0.40\,{\tiny $\pm$0.00} & \textbf{0.38}\,{\tiny $\pm$0.00} & 0.44\,{\tiny $\pm$0.00} & 0.46\,{\tiny $\pm$0.01} & 0.47\,{\tiny $\pm$0.00} & \textbf{0.54}\,{\tiny $\pm$0.00} & 0.61\,{\tiny $\pm$0.00} & \textbf{0.74}\,{\tiny $\pm$0.00} & 0.78\,{\tiny $\pm$0.00} & 0.48\,{\tiny $\pm$0.01} & \textbf{0.54}\,{\tiny $\pm$0.02} & \textbf{0.49}\,{\tiny $\pm$0.00} & 0.54\,{\tiny $\pm$0.00} \\
\cmidrule(lr){2-17}
& \multirow{7}{*}{\shortstack{$16\times$\\{\footnotesize $\ell_0 \approx 128$}}} & ReLU
& 0.37\,{\tiny $\pm$0.00} & 0.41\,{\tiny $\pm$0.00} & 0.38\,{\tiny $\pm$0.00} & 0.45\,{\tiny $\pm$0.00} & 0.49\,{\tiny $\pm$0.01} & 0.50\,{\tiny $\pm$0.00} & 0.50\,{\tiny $\pm$0.01} & 0.61\,{\tiny $\pm$0.00} & 0.72\,{\tiny $\pm$0.01} & 0.79\,{\tiny $\pm$0.00} & 0.41\,{\tiny $\pm$0.03} & 0.52\,{\tiny $\pm$0.01} & 0.48\,{\tiny $\pm$0.01} & 0.55\,{\tiny $\pm$0.00} \\
&  & TopK
& 0.31\,{\tiny $\pm$0.00} & 0.34\,{\tiny $\pm$0.00} & 0.36\,{\tiny $\pm$0.00} & 0.44\,{\tiny $\pm$0.00} & 0.46\,{\tiny $\pm$0.01} & 0.51\,{\tiny $\pm$0.00} & 0.48\,{\tiny $\pm$0.01} & 0.59\,{\tiny $\pm$0.00} & 0.69\,{\tiny $\pm$0.00} & 0.74\,{\tiny $\pm$0.00} & 0.45\,{\tiny $\pm$0.01} & 0.52\,{\tiny $\pm$0.01} & 0.46\,{\tiny $\pm$0.01} & 0.53\,{\tiny $\pm$0.00} \\
&  & BatchTopK
& 0.28\,{\tiny $\pm$0.00} & 0.34\,{\tiny $\pm$0.00} & 0.35\,{\tiny $\pm$0.00} & 0.44\,{\tiny $\pm$0.00} & 0.44\,{\tiny $\pm$0.00} & 0.50\,{\tiny $\pm$0.00} & 0.48\,{\tiny $\pm$0.00} & 0.59\,{\tiny $\pm$0.00} & 0.68\,{\tiny $\pm$0.00} & 0.74\,{\tiny $\pm$0.00} & 0.44\,{\tiny $\pm$0.03} & 0.53\,{\tiny $\pm$0.01} & 0.45\,{\tiny $\pm$0.00} & 0.52\,{\tiny $\pm$0.00} \\
&  & Matryoshka
& 0.26\,{\tiny $\pm$0.00} & 0.29\,{\tiny $\pm$0.00} & 0.33\,{\tiny $\pm$0.00} & 0.37\,{\tiny $\pm$0.00} & 0.40\,{\tiny $\pm$0.01} & 0.44\,{\tiny $\pm$0.01} & 0.48\,{\tiny $\pm$0.01} & 0.57\,{\tiny $\pm$0.01} & 0.62\,{\tiny $\pm$0.01} & 0.66\,{\tiny $\pm$0.01} & 0.43\,{\tiny $\pm$0.03} & 0.46\,{\tiny $\pm$0.00} & 0.42\,{\tiny $\pm$0.01} & 0.46\,{\tiny $\pm$0.00} \\
&  & MP-SAE
& \textbf{0.42}\,{\tiny $\pm$0.00} & \textbf{0.45}\,{\tiny $\pm$0.00} & 0.40\,{\tiny $\pm$0.01} & \textbf{0.49}\,{\tiny $\pm$0.00} & 0.47\,{\tiny $\pm$0.01} & \textbf{0.56}\,{\tiny $\pm$0.01} & 0.48\,{\tiny $\pm$0.01} & \textbf{0.62}\,{\tiny $\pm$0.01} & 0.77\,{\tiny $\pm$0.01} & \textbf{0.82}\,{\tiny $\pm$0.01} & 0.39\,{\tiny $\pm$0.05} & 0.48\,{\tiny $\pm$0.02} & 0.49\,{\tiny $\pm$0.01} & \textbf{0.57}\,{\tiny $\pm$0.00} \\
&  & iSAE
& 0.19\,{\tiny $\pm$0.00} & 0.30\,{\tiny $\pm$0.00} & 0.19\,{\tiny $\pm$0.01} & 0.39\,{\tiny $\pm$0.00} & 0.19\,{\tiny $\pm$0.02} & 0.46\,{\tiny $\pm$0.00} & 0.25\,{\tiny $\pm$0.02} & 0.58\,{\tiny $\pm$0.01} & 0.36\,{\tiny $\pm$0.01} & 0.70\,{\tiny $\pm$0.00} & 0.27\,{\tiny $\pm$0.06} & \textbf{0.55}\,{\tiny $\pm$0.02} & 0.24\,{\tiny $\pm$0.01} & 0.50\,{\tiny $\pm$0.00} \\
&  & ConSDL
& 0.38\,{\tiny $\pm$0.00} & 0.40\,{\tiny $\pm$0.00} & \textbf{0.41}\,{\tiny $\pm$0.00} & 0.44\,{\tiny $\pm$0.00} & 0.49\,{\tiny $\pm$0.01} & 0.48\,{\tiny $\pm$0.01} & \textbf{0.54}\,{\tiny $\pm$0.01} & 0.61\,{\tiny $\pm$0.01} & 0.77\,{\tiny $\pm$0.00} & 0.79\,{\tiny $\pm$0.00} & \textbf{0.51}\,{\tiny $\pm$0.03} & 0.55\,{\tiny $\pm$0.01} & \textbf{0.52}\,{\tiny $\pm$0.01} & 0.55\,{\tiny $\pm$0.00} \\
\midrule
\multirow{21}{*}{DINOv2-B/14}
& \multirow{7}{*}{\shortstack{$4\times$\\{\footnotesize $\ell_0 \approx 32$}}} & ReLU
& \textbf{0.31}\,{\tiny $\pm$0.00} & 0.39\,{\tiny $\pm$0.00} & 0.31\,{\tiny $\pm$0.00} & 0.42\,{\tiny $\pm$0.00} & 0.36\,{\tiny $\pm$0.01} & 0.46\,{\tiny $\pm$0.01} & 0.51\,{\tiny $\pm$0.01} & 0.67\,{\tiny $\pm$0.01} & \textbf{0.68}\,{\tiny $\pm$0.00} & 0.78\,{\tiny $\pm$0.00} & 0.37\,{\tiny $\pm$0.01} & 0.41\,{\tiny $\pm$0.00} & 0.42\,{\tiny $\pm$0.00} & 0.52\,{\tiny $\pm$0.00} \\
&  & TopK
& 0.29\,{\tiny $\pm$0.00} & \textbf{0.39}\,{\tiny $\pm$0.00} & 0.31\,{\tiny $\pm$0.00} & \textbf{0.45}\,{\tiny $\pm$0.00} & 0.36\,{\tiny $\pm$0.00} & \textbf{0.50}\,{\tiny $\pm$0.00} & \textbf{0.53}\,{\tiny $\pm$0.00} & \textbf{0.69}\,{\tiny $\pm$0.00} & 0.67\,{\tiny $\pm$0.00} & 0.79\,{\tiny $\pm$0.00} & \textbf{0.40}\,{\tiny $\pm$0.00} & 0.42\,{\tiny $\pm$0.01} & \textbf{0.43}\,{\tiny $\pm$0.00} & \textbf{0.54}\,{\tiny $\pm$0.00} \\
&  & BatchTopK
& 0.29\,{\tiny $\pm$0.00} & 0.39\,{\tiny $\pm$0.00} & \textbf{0.31}\,{\tiny $\pm$0.00} & 0.45\,{\tiny $\pm$0.00} & \textbf{0.36}\,{\tiny $\pm$0.00} & 0.50\,{\tiny $\pm$0.00} & 0.51\,{\tiny $\pm$0.01} & 0.68\,{\tiny $\pm$0.00} & 0.68\,{\tiny $\pm$0.00} & 0.79\,{\tiny $\pm$0.00} & 0.37\,{\tiny $\pm$0.00} & 0.44\,{\tiny $\pm$0.01} & 0.42\,{\tiny $\pm$0.00} & 0.54\,{\tiny $\pm$0.00} \\
&  & Matryoshka
& 0.27\,{\tiny $\pm$0.00} & 0.37\,{\tiny $\pm$0.00} & 0.29\,{\tiny $\pm$0.00} & 0.42\,{\tiny $\pm$0.00} & 0.35\,{\tiny $\pm$0.00} & 0.47\,{\tiny $\pm$0.00} & 0.53\,{\tiny $\pm$0.01} & 0.66\,{\tiny $\pm$0.00} & 0.63\,{\tiny $\pm$0.01} & 0.76\,{\tiny $\pm$0.00} & 0.37\,{\tiny $\pm$0.02} & \textbf{0.44}\,{\tiny $\pm$0.01} & 0.41\,{\tiny $\pm$0.00} & 0.52\,{\tiny $\pm$0.00} \\
&  & MP-SAE
& 0.28\,{\tiny $\pm$0.00} & 0.38\,{\tiny $\pm$0.00} & 0.29\,{\tiny $\pm$0.00} & 0.43\,{\tiny $\pm$0.00} & 0.35\,{\tiny $\pm$0.01} & 0.49\,{\tiny $\pm$0.00} & 0.52\,{\tiny $\pm$0.01} & 0.67\,{\tiny $\pm$0.00} & 0.65\,{\tiny $\pm$0.00} & 0.77\,{\tiny $\pm$0.00} & 0.38\,{\tiny $\pm$0.02} & 0.41\,{\tiny $\pm$0.01} & 0.41\,{\tiny $\pm$0.00} & 0.53\,{\tiny $\pm$0.00} \\
&  & iSAE
& 0.13\,{\tiny $\pm$0.00} & 0.31\,{\tiny $\pm$0.00} & 0.16\,{\tiny $\pm$0.02} & 0.35\,{\tiny $\pm$0.00} & 0.20\,{\tiny $\pm$0.02} & 0.38\,{\tiny $\pm$0.00} & 0.29\,{\tiny $\pm$0.03} & 0.64\,{\tiny $\pm$0.00} & 0.34\,{\tiny $\pm$0.02} & 0.70\,{\tiny $\pm$0.01} & 0.24\,{\tiny $\pm$0.07} & 0.40\,{\tiny $\pm$0.01} & 0.23\,{\tiny $\pm$0.01} & 0.46\,{\tiny $\pm$0.00} \\
&  & ConSDL
& 0.29\,{\tiny $\pm$0.00} & 0.38\,{\tiny $\pm$0.00} & 0.30\,{\tiny $\pm$0.00} & 0.42\,{\tiny $\pm$0.00} & 0.35\,{\tiny $\pm$0.01} & 0.48\,{\tiny $\pm$0.00} & 0.48\,{\tiny $\pm$0.01} & 0.67\,{\tiny $\pm$0.01} & 0.66\,{\tiny $\pm$0.01} & 0.78\,{\tiny $\pm$0.00} & 0.37\,{\tiny $\pm$0.01} & 0.41\,{\tiny $\pm$0.01} & 0.41\,{\tiny $\pm$0.00} & 0.52\,{\tiny $\pm$0.00} \\
\cmidrule(lr){2-17}
& \multirow{7}{*}{\shortstack{$8\times$\\{\footnotesize $\ell_0 \approx 64$}}} & ReLU
& \textbf{0.33}\,{\tiny $\pm$0.00} & \textbf{0.41}\,{\tiny $\pm$0.00} & \textbf{0.33}\,{\tiny $\pm$0.00} & 0.44\,{\tiny $\pm$0.00} & \textbf{0.38}\,{\tiny $\pm$0.01} & 0.48\,{\tiny $\pm$0.00} & 0.53\,{\tiny $\pm$0.00} & 0.68\,{\tiny $\pm$0.00} & \textbf{0.71}\,{\tiny $\pm$0.00} & \textbf{0.80}\,{\tiny $\pm$0.00} & 0.37\,{\tiny $\pm$0.01} & 0.40\,{\tiny $\pm$0.01} & \textbf{0.44}\,{\tiny $\pm$0.00} & 0.54\,{\tiny $\pm$0.00} \\
&  & TopK
& 0.30\,{\tiny $\pm$0.00} & 0.40\,{\tiny $\pm$0.00} & 0.32\,{\tiny $\pm$0.00} & \textbf{0.46}\,{\tiny $\pm$0.00} & 0.37\,{\tiny $\pm$0.00} & 0.49\,{\tiny $\pm$0.00} & 0.53\,{\tiny $\pm$0.01} & 0.69\,{\tiny $\pm$0.00} & 0.70\,{\tiny $\pm$0.00} & 0.79\,{\tiny $\pm$0.00} & \textbf{0.40}\,{\tiny $\pm$0.01} & \textbf{0.42}\,{\tiny $\pm$0.01} & 0.44\,{\tiny $\pm$0.00} & \textbf{0.54}\,{\tiny $\pm$0.00} \\
&  & BatchTopK
& 0.30\,{\tiny $\pm$0.00} & 0.40\,{\tiny $\pm$0.00} & 0.32\,{\tiny $\pm$0.00} & 0.46\,{\tiny $\pm$0.00} & 0.37\,{\tiny $\pm$0.00} & 0.50\,{\tiny $\pm$0.00} & 0.52\,{\tiny $\pm$0.01} & 0.69\,{\tiny $\pm$0.00} & 0.69\,{\tiny $\pm$0.00} & 0.79\,{\tiny $\pm$0.00} & 0.39\,{\tiny $\pm$0.00} & 0.42\,{\tiny $\pm$0.01} & 0.43\,{\tiny $\pm$0.00} & 0.54\,{\tiny $\pm$0.00} \\
&  & Matryoshka
& 0.28\,{\tiny $\pm$0.00} & 0.37\,{\tiny $\pm$0.00} & 0.30\,{\tiny $\pm$0.00} & 0.43\,{\tiny $\pm$0.00} & 0.36\,{\tiny $\pm$0.00} & 0.48\,{\tiny $\pm$0.00} & 0.53\,{\tiny $\pm$0.00} & 0.67\,{\tiny $\pm$0.01} & 0.66\,{\tiny $\pm$0.01} & 0.77\,{\tiny $\pm$0.00} & 0.36\,{\tiny $\pm$0.01} & 0.41\,{\tiny $\pm$0.01} & 0.42\,{\tiny $\pm$0.00} & 0.52\,{\tiny $\pm$0.00} \\
&  & MP-SAE
& 0.30\,{\tiny $\pm$0.00} & 0.39\,{\tiny $\pm$0.00} & 0.32\,{\tiny $\pm$0.00} & 0.45\,{\tiny $\pm$0.00} & 0.37\,{\tiny $\pm$0.00} & \textbf{0.51}\,{\tiny $\pm$0.00} & 0.52\,{\tiny $\pm$0.01} & 0.68\,{\tiny $\pm$0.00} & 0.69\,{\tiny $\pm$0.00} & 0.79\,{\tiny $\pm$0.00} & 0.37\,{\tiny $\pm$0.00} & 0.39\,{\tiny $\pm$0.01} & 0.43\,{\tiny $\pm$0.00} & 0.54\,{\tiny $\pm$0.00} \\
&  & iSAE
& 0.14\,{\tiny $\pm$0.00} & 0.32\,{\tiny $\pm$0.00} & 0.15\,{\tiny $\pm$0.01} & 0.37\,{\tiny $\pm$0.00} & 0.20\,{\tiny $\pm$0.00} & 0.39\,{\tiny $\pm$0.00} & 0.28\,{\tiny $\pm$0.05} & 0.63\,{\tiny $\pm$0.00} & 0.31\,{\tiny $\pm$0.02} & 0.72\,{\tiny $\pm$0.00} & 0.20\,{\tiny $\pm$0.10} & 0.38\,{\tiny $\pm$0.00} & 0.21\,{\tiny $\pm$0.02} & 0.47\,{\tiny $\pm$0.00} \\
&  & ConSDL
& 0.31\,{\tiny $\pm$0.00} & 0.40\,{\tiny $\pm$0.00} & 0.32\,{\tiny $\pm$0.00} & 0.43\,{\tiny $\pm$0.00} & 0.37\,{\tiny $\pm$0.00} & 0.46\,{\tiny $\pm$0.00} & 0.50\,{\tiny $\pm$0.02} & 0.68\,{\tiny $\pm$0.01} & 0.68\,{\tiny $\pm$0.01} & 0.79\,{\tiny $\pm$0.00} & 0.36\,{\tiny $\pm$0.01} & 0.40\,{\tiny $\pm$0.01} & 0.42\,{\tiny $\pm$0.00} & 0.53\,{\tiny $\pm$0.00} \\
\cmidrule(lr){2-17}
& \multirow{7}{*}{\shortstack{$16\times$\\{\footnotesize $\ell_0 \approx 128$}}} & ReLU
& \textbf{0.35}\,{\tiny $\pm$0.00} & 0.41\,{\tiny $\pm$0.00} & \textbf{0.34}\,{\tiny $\pm$0.00} & 0.44\,{\tiny $\pm$0.00} & 0.38\,{\tiny $\pm$0.01} & 0.47\,{\tiny $\pm$0.00} & 0.52\,{\tiny $\pm$0.01} & 0.67\,{\tiny $\pm$0.01} & \textbf{0.72}\,{\tiny $\pm$0.01} & 0.79\,{\tiny $\pm$0.00} & 0.37\,{\tiny $\pm$0.01} & 0.40\,{\tiny $\pm$0.00} & \textbf{0.45}\,{\tiny $\pm$0.00} & 0.53\,{\tiny $\pm$0.00} \\
&  & TopK
& 0.31\,{\tiny $\pm$0.00} & 0.39\,{\tiny $\pm$0.00} & 0.32\,{\tiny $\pm$0.00} & 0.44\,{\tiny $\pm$0.00} & 0.37\,{\tiny $\pm$0.00} & 0.48\,{\tiny $\pm$0.00} & 0.52\,{\tiny $\pm$0.00} & 0.69\,{\tiny $\pm$0.00} & 0.69\,{\tiny $\pm$0.00} & 0.78\,{\tiny $\pm$0.00} & 0.38\,{\tiny $\pm$0.02} & 0.40\,{\tiny $\pm$0.00} & 0.43\,{\tiny $\pm$0.00} & 0.53\,{\tiny $\pm$0.00} \\
&  & BatchTopK
& 0.30\,{\tiny $\pm$0.00} & 0.39\,{\tiny $\pm$0.00} & 0.32\,{\tiny $\pm$0.00} & 0.44\,{\tiny $\pm$0.00} & 0.36\,{\tiny $\pm$0.00} & 0.48\,{\tiny $\pm$0.00} & 0.52\,{\tiny $\pm$0.00} & \textbf{0.69}\,{\tiny $\pm$0.01} & 0.69\,{\tiny $\pm$0.00} & 0.78\,{\tiny $\pm$0.00} & 0.38\,{\tiny $\pm$0.01} & 0.40\,{\tiny $\pm$0.00} & 0.43\,{\tiny $\pm$0.00} & 0.53\,{\tiny $\pm$0.00} \\
&  & Matryoshka
& 0.28\,{\tiny $\pm$0.00} & 0.37\,{\tiny $\pm$0.00} & 0.31\,{\tiny $\pm$0.00} & 0.42\,{\tiny $\pm$0.00} & 0.37\,{\tiny $\pm$0.01} & 0.47\,{\tiny $\pm$0.00} & 0.51\,{\tiny $\pm$0.01} & 0.66\,{\tiny $\pm$0.01} & 0.66\,{\tiny $\pm$0.01} & 0.76\,{\tiny $\pm$0.00} & 0.37\,{\tiny $\pm$0.00} & 0.40\,{\tiny $\pm$0.00} & 0.42\,{\tiny $\pm$0.00} & 0.51\,{\tiny $\pm$0.00} \\
&  & MP-SAE
& 0.34\,{\tiny $\pm$0.00} & \textbf{0.42}\,{\tiny $\pm$0.00} & 0.33\,{\tiny $\pm$0.00} & \textbf{0.45}\,{\tiny $\pm$0.00} & 0.37\,{\tiny $\pm$0.00} & \textbf{0.50}\,{\tiny $\pm$0.01} & 0.52\,{\tiny $\pm$0.00} & 0.68\,{\tiny $\pm$0.01} & 0.72\,{\tiny $\pm$0.00} & \textbf{0.81}\,{\tiny $\pm$0.00} & 0.34\,{\tiny $\pm$0.01} & 0.38\,{\tiny $\pm$0.00} & 0.44\,{\tiny $\pm$0.00} & \textbf{0.54}\,{\tiny $\pm$0.00} \\
&  & iSAE
& 0.16\,{\tiny $\pm$0.00} & 0.33\,{\tiny $\pm$0.00} & 0.17\,{\tiny $\pm$0.00} & 0.36\,{\tiny $\pm$0.00} & 0.21\,{\tiny $\pm$0.01} & 0.39\,{\tiny $\pm$0.00} & 0.30\,{\tiny $\pm$0.06} & 0.63\,{\tiny $\pm$0.01} & 0.37\,{\tiny $\pm$0.02} & 0.73\,{\tiny $\pm$0.00} & 0.25\,{\tiny $\pm$0.09} & 0.38\,{\tiny $\pm$0.01} & 0.24\,{\tiny $\pm$0.02} & 0.47\,{\tiny $\pm$0.00} \\
&  & ConSDL
& 0.33\,{\tiny $\pm$0.00} & 0.38\,{\tiny $\pm$0.00} & 0.33\,{\tiny $\pm$0.00} & 0.39\,{\tiny $\pm$0.00} & \textbf{0.39}\,{\tiny $\pm$0.00} & 0.42\,{\tiny $\pm$0.00} & \textbf{0.53}\,{\tiny $\pm$0.01} & 0.66\,{\tiny $\pm$0.02} & 0.71\,{\tiny $\pm$0.01} & 0.77\,{\tiny $\pm$0.00} & 0.35\,{\tiny $\pm$0.00} & 0.40\,{\tiny $\pm$0.02} & 0.44\,{\tiny $\pm$0.00} & 0.50\,{\tiny $\pm$0.01} \\
\bottomrule
\end{tabular}%
}
\end{table*}

\begin{table*}[t]
\centering
\caption{
\textbf{Sparse probing performance of language-model dictionaries.}
We report AUROC using the $k$ most discriminative dictionary atoms per probe for
$k \in \{1,5\}$ across three matched expansion and sparsity settings: $4\times$ with $\ell_0 \approx 32$, $8\times$ with $\ell_0 \approx 64$, and $16\times$ with $\ell_0 \approx 128$.
Scores are averaged over the 113 binary probes in SAE-Probes.
$\pm$ is the standard deviation across training seeds of the reported score. 
}
\label{tab:lm_probing_main}

\setlength{\tabcolsep}{4pt}

\small
\begin{tabular}{ll cc cc cc}
\toprule
\multirow{2}{*}{\textbf{Backbone}} &
\multirow{2}{*}{\textbf{Method}} &
\multicolumn{2}{c}{$\mathbf{4\times}$, $\ell_0 \approx 32$} &
\multicolumn{2}{c}{$\mathbf{8\times}$, $\ell_0 \approx 64$} &
\multicolumn{2}{c}{$\mathbf{16\times}$, $\ell_0 \approx 128$} \\
\cmidrule(lr){3-4}
\cmidrule(lr){5-6}
\cmidrule(lr){7-8}
&
& @1 & @5 & @1 & @5 & @1 & @5 \\
\midrule

\multirow{7}{*}{Pythia-160M}
& ReLU
& 0.69\,{\scriptsize $\pm$0.00} & 0.75\,{\scriptsize $\pm$0.00}
& 0.70\,{\scriptsize $\pm$0.00} & 0.77\,{\scriptsize $\pm$0.00}
& 0.70\,{\scriptsize $\pm$0.00} & 0.78\,{\scriptsize $\pm$0.00} \\

& TopK
& 0.70\,{\scriptsize $\pm$0.00} & 0.77\,{\scriptsize $\pm$0.00}
& 0.71\,{\scriptsize $\pm$0.00} & 0.78\,{\scriptsize $\pm$0.00}
& 0.72\,{\scriptsize $\pm$0.00} & 0.80\,{\scriptsize $\pm$0.00} \\

& BatchTopK
& 0.70\,{\scriptsize $\pm$0.00} & 0.77\,{\scriptsize $\pm$0.00}
& \textbf{0.72}\,{\scriptsize $\pm$0.01} & 0.79\,{\scriptsize $\pm$0.00}
& 0.73\,{\scriptsize $\pm$0.00} & \textbf{0.80}\,{\scriptsize $\pm$0.00} \\

& Matryoshka
& \textbf{0.71}\,{\scriptsize $\pm$0.00} & 0.77\,{\scriptsize $\pm$0.00}
& 0.72\,{\scriptsize $\pm$0.01} & 0.79\,{\scriptsize $\pm$0.00}
& 0.72\,{\scriptsize $\pm$0.00} & 0.79\,{\scriptsize $\pm$0.00} \\

& MP-SAE
& 0.70\,{\scriptsize $\pm$0.00} & \textbf{0.78}\,{\scriptsize $\pm$0.00}
& 0.68\,{\scriptsize $\pm$0.00} & 0.76\,{\scriptsize $\pm$0.00}
& 0.67\,{\scriptsize $\pm$0.00} & 0.74\,{\scriptsize $\pm$0.00} \\

& iSAE
& 0.70\,{\scriptsize $\pm$0.00} & 0.77\,{\scriptsize $\pm$0.00}
& 0.71\,{\scriptsize $\pm$0.00} & 0.78\,{\scriptsize $\pm$0.00}
& \textbf{0.73}\,{\scriptsize $\pm$0.00} & 0.79\,{\scriptsize $\pm$0.00} \\

& ConSDL
& 0.70\,{\scriptsize $\pm$0.00} & 0.76\,{\scriptsize $\pm$0.00}
& 0.71\,{\scriptsize $\pm$0.00} & 0.77\,{\scriptsize $\pm$0.00}
& 0.73\,{\scriptsize $\pm$0.00} & 0.78\,{\scriptsize $\pm$0.00} \\

\midrule

\multirow{7}{*}{GPT-2 Small}
& ReLU
& 0.71\,{\scriptsize $\pm$0.00} & 0.77\,{\scriptsize $\pm$0.00}
& 0.73\,{\scriptsize $\pm$0.00} & 0.79\,{\scriptsize $\pm$0.00}
& 0.74\,{\scriptsize $\pm$0.00} & 0.79\,{\scriptsize $\pm$0.00} \\

& TopK
& \textbf{0.72}\,{\scriptsize $\pm$0.00} & \textbf{0.78}\,{\scriptsize $\pm$0.00}
& \textbf{0.73}\,{\scriptsize $\pm$0.00} & 0.79\,{\scriptsize $\pm$0.00}
& 0.75\,{\scriptsize $\pm$0.00} & 0.80\,{\scriptsize $\pm$0.00} \\

& BatchTopK
& 0.72\,{\scriptsize $\pm$0.00} & 0.78\,{\scriptsize $\pm$0.00}
& 0.73\,{\scriptsize $\pm$0.00} & 0.79\,{\scriptsize $\pm$0.00}
& 0.75\,{\scriptsize $\pm$0.00} & 0.80\,{\scriptsize $\pm$0.00} \\

& Matryoshka
& 0.71\,{\scriptsize $\pm$0.00} & 0.78\,{\scriptsize $\pm$0.00}
& 0.73\,{\scriptsize $\pm$0.00} & 0.79\,{\scriptsize $\pm$0.00}
& 0.74\,{\scriptsize $\pm$0.00} & 0.79\,{\scriptsize $\pm$0.00} \\

& MP-SAE
& 0.70\,{\scriptsize $\pm$0.00} & 0.77\,{\scriptsize $\pm$0.00}
& 0.69\,{\scriptsize $\pm$0.00} & 0.76\,{\scriptsize $\pm$0.00}
& 0.67\,{\scriptsize $\pm$0.00} & 0.75\,{\scriptsize $\pm$0.00} \\

& iSAE
& 0.72\,{\scriptsize $\pm$0.00} & 0.77\,{\scriptsize $\pm$0.00}
& 0.73\,{\scriptsize $\pm$0.00} & 0.79\,{\scriptsize $\pm$0.00}
& 0.74\,{\scriptsize $\pm$0.00} & 0.80\,{\scriptsize $\pm$0.00} \\

& ConSDL
& 0.71\,{\scriptsize $\pm$0.00} & 0.77\,{\scriptsize $\pm$0.00}
& 0.73\,{\scriptsize $\pm$0.00} & 0.78\,{\scriptsize $\pm$0.00}
& 0.74\,{\scriptsize $\pm$0.00} & 0.79\,{\scriptsize $\pm$0.00} \\

\bottomrule
\end{tabular}
\end{table*}

\subsection{Sparse Probing Evaluation}
\label{sec:app:sparse_probing}

We provide the full sparse probing protocol and results for vision and language models. Following prior work~\citep{gurnee2023finding,karvonen2025saebench}, we identify dictionary atoms that best distinguish positive from negative examples and evaluate probes restricted to the Top-1 or Top-5 selected atoms.

\textbf{Atom selection.}
We follow the class-conditional activation criterion used in sparse probing~\citep{gurnee2023finding}. We first normalize each atom by its mean nonzero activation magnitude and then rank atoms by the absolute difference between their mean normalized activations on positive and negative training examples. Atom selection is performed using only the training split.

\textbf{Probe construction.}
We evaluate the Top-1 and Top-5 selected atoms. For Top-1, we use the absolute activation of the selected atom directly as the prediction score, without fitting an additional classifier. For Top-5, we fit a logistic regression classifier using the raw activations of the five selected atoms on the training split and evaluate its decision scores on the evaluation split.

\textbf{Language benchmarks.}
For language models, we use the SAE-Probes benchmark \citep{kantamneni2025sparse}, which consists of 113 binary classification tasks spanning a diverse range of linguistic and semantic concepts. For each task, examples are labeled as positive or negative according to the target concept, and the learned SAE activations are used as probe inputs. We follow the benchmark protocol for constructing the training and evaluation sets and report the area under the receiver operating characteristic curve (AUROC).

\textbf{Vision benchmarks.} For vision models, following the evaluation strategy of ConceptScope \citep{choi2026conceptscope}, we evaluate sparse probing across datasets covering diverse types of visual concepts. We use ImageNet \citep{russakovsky2015imagenet} with 1,000 object classes, SUN397 \citep{xiao2010sun} with 397 scene categories, Food-101 \citep{bossard2014food}with 101 food categories, Caltech-256 \citep{griffin2007caltech} with 256 object categories, DTD \citep{cimpoi2014describing} with 47 texture categories, and RAF-DB \citep{li2017reliable} with 7 facial emotion categories. Each category is treated as a one-vs-rest binary prediction task. We use 50 training images per category for atom selection, and evaluate on the full validation or test split. All available negative examples are used for evaluation. We report the area under the precision--recall curve (AUPRC) because the resulting one-vs-rest tasks are highly imbalanced.

\subsection{Analysis: Region carve and cross-seed latent matching}
\label{sec:app:analysis}

This section examines whether independently trained SAEs decompose the same input in the same way, and whether individual latents recover the same selectivity across seeds. 

\begin{figure}[p]
\centering
\includegraphics[width=1.0\textwidth]{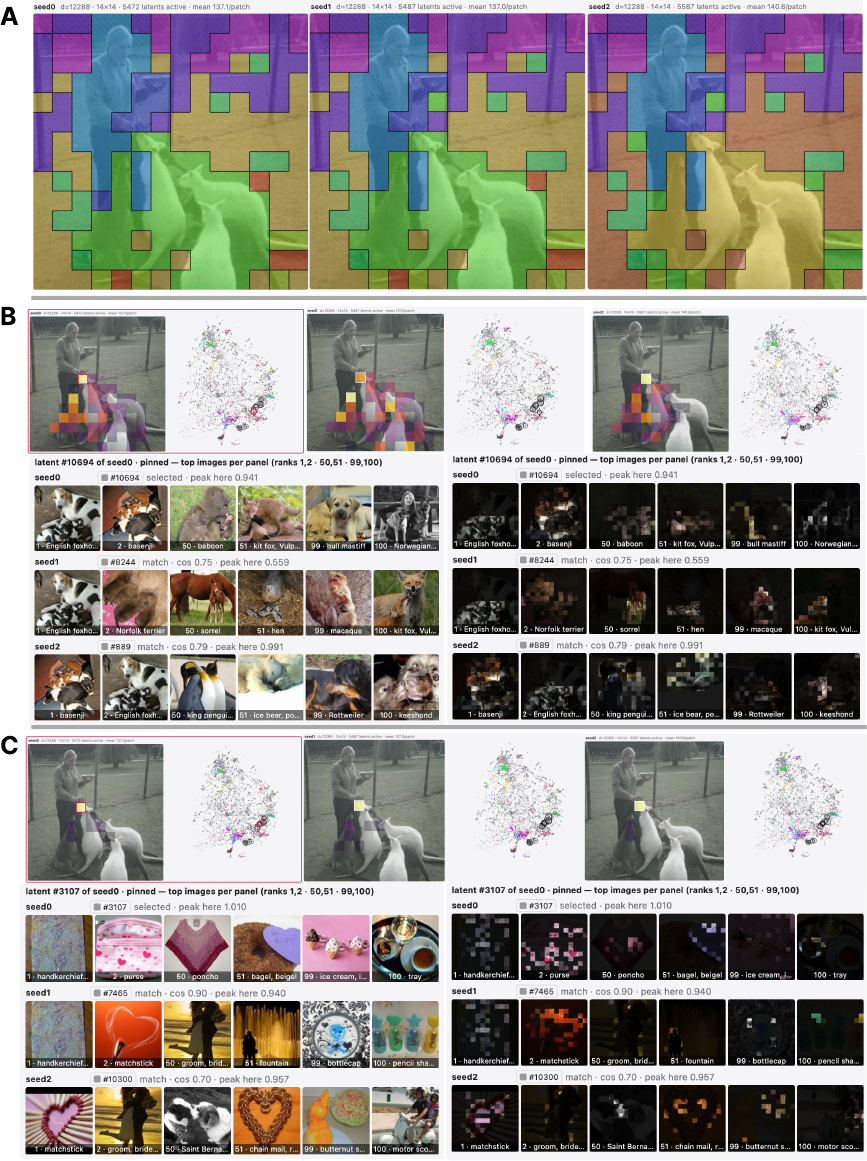}
\vspace{-0.6em}
\caption{
\textbf{Region decomposition and cross-seed latent analysis in CLIP ViT-B/16.}
$16\times$ dictionary ($m=12288$), $\ell_0\approx128$, three seeds.
\textbf{(A)} Each $14\times14$ patch grid is independently clustered into $k=10$ regions from its coefficient vectors. Hue encodes region position, and headers report active latents and their mean per patch.
\textbf{(B)} One seed-0 patch is selected (red outline). Top: activation map and shared UMAP, fitted on seed~0 with seeds~1--2 projected into the same space. Black rings mark latents active at the patch, red the selected latent, and gray its nearest matches. Latent \#10694 matches \#8244 ($\cos=0.75$) and \#889 ($\cos=0.79$). Bottom: top-activating images at ranks $1,2$, $50,51$, and $99,100$, shown unmasked and activation-masked, reveal animals in close contact.
\textbf{(C)} Same analysis for a second latent at the patch. \#3107 matches \#7465 ($\cos=0.90$) and \#10300 ($\cos=0.70$), showing a heart/romance motif.
}

\label{fig:clip_qual}
\end{figure}

\begin{figure}[p]
\centering
\includegraphics[width=1.0\textwidth]{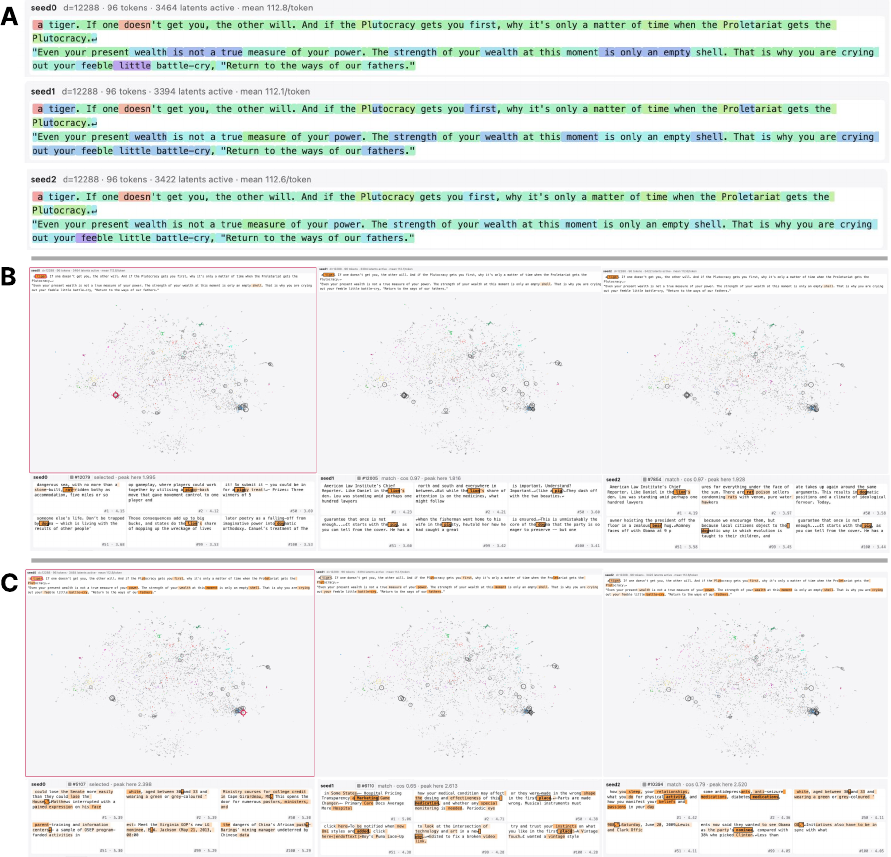}
\caption{
\textbf{Region decomposition and cross-seed latent analysis in Pythia-160M.}
$16\times$ dictionary ($m=12288$), three seeds, using a 96-token window with approximately 112 active latents per token.
\textbf{(A)} Tokens are independently clustered into $k=10$ regions for each seed. Hue encodes position in the text rather than content.
\textbf{(B)} One seed-0 token is selected (red outline). For each seed, we show the passage with latent activations highlighted, a shared UMAP fitted on seed~0 with seeds~1--2 projected into the same space, and top-activating contexts at ranks $1,2$, $50,51$, and $99,100$. Black rings mark latents active at the token, red the selected latent, and gray its nearest matches. Latent \#12079 matches \#12005 ($\cos=0.97$) and \#7854 ($\cos=0.97$), capturing animal nouns.
\textbf{(C)} Same analysis for a second latent. \#5107 matches \#6110 ($\cos=0.65$) and \#10394 ($\cos=0.79$), capturing sentence-final punctuation and the preceding token.
}
\label{fig:pythia_qual}
\end{figure}

\textbf{Region-level decomposition.} Figure~\ref{fig:clip_qual}A and Fig.~\ref{fig:pythia_qual}A test whether different seeds decompose the same input similarly. We cluster patches or tokens based on their coefficient vectors. At each position, coefficients below \(1\%\) of the maximum are discarded and the remainder renormalized. We compare positions using half the \(L_1\) distance between these vectors, ranging from \(0\) for identical coefficients to \(1\) for no overlap. Average-linkage hierarchical clustering is then applied separately for each seed to form \(k=10\) regions. Regions are colored by their mean position in the image or text, and agreement across seeds is measured by the adjusted Rand index (ARI).

\textbf{Organization of matched latents across seeds.} Figures~\ref{fig:clip_qual}B--C and~\ref{fig:pythia_qual}B--C examine whether latents matched across seeds occupy similar regions of their learned dictionaries and exhibit similar activation behavior. We match latents by decoder cosine similarity and visualize the three dictionaries side by side in a shared UMAP space, fitted on one seed and used to project the other two. We also compare the regions where matched latents activate on the same input and their top-activating samples.

\textbf{Vision}. We analyze CLIP ViT-B/16 at \(16\times\) expansion and \(\ell_0\approx128\) using an ImageNet validation image. In Fig.~\ref{fig:clip_qual}A, the three seeds produce nearly identical region decompositions, with pairwise ARI values of \(\{0.98,0.98,0.97\}\), consistently separating the handler, fence and background, and the animals. Panels B and C examine two matched latents active on a patch on the animal's head. The first responds to animals in close physical contact, such as nuzzling, grooming, and huddling, with activation concentrated around the contact region rather than a particular species. Its top-activating images show the same relation across unrelated classes. The second captures a heart or romance-related pattern and is similarly reproduced across seeds. These examples show that a single patch can be decomposed into distinct latents that recur across independent runs.

\textbf{Language.} We analyze Pythia-160M at \(16\times\) expansion and \(\ell_0\approx128\) using an example from the 20M-token FineWeb evaluation set. In Fig.~\ref{fig:pythia_qual}A, the three seeds produce broadly similar region decompositions, with pairwise ARI values of \(\{0.83,0.65,0.74\}\). We exclude position 0 because its unusually large activation would otherwise form a separate region. Panels B and C examine two matched latents. The first responds to animal nouns such as \emph{tiger}, \emph{lion}, \emph{pig}, \emph{rat}, and \emph{bear} across unrelated contexts and has high decoder similarity across seeds (\(\cos=0.97\) to both). The second responds to sentence-final punctuation and the preceding token. Despite lower decoder similarity (\(\cos=0.65\) and \(0.79\)), its top-activating contexts show the same pattern across all three seeds. This suggests that similar activation behavior can persist even when decoder similarity is moderate.

\begin{table*}[p]
\centering
\caption{
\textbf{Full synthetic results with a 3K-atom ground-truth dictionary across tail exponents $\beta$.}
We evaluate all methods across $\beta\in\{0,0.5,0.7,1.0\}$.
We report sparsity $\ell_0$, reconstruction ($R^2$), ground-truth recovery
(\texttt{DCS}, \texttt{MCC}), cross-seed consistency, and dictionary utilization
measured by the usable atom fraction. The numbers of dead and duplicated atoms
are also reported. Metrics are averaged across training seeds, while cross-seed
\texttt{DCS} and \texttt{MCC} are averaged across seed pairs.
$\pm$ denotes the corresponding standard deviation.
}
\label{tab:synthetic_3k}
\resizebox{\textwidth}{!}{%
\begin{tabular}{ll c c cc cc ccc}
\toprule
\multirow{2}{*}{\textbf{Setting}} &
\multirow{2}{*}{\textbf{Architecture}} &
\multicolumn{1}{c}{\textbf{Sparsity}} &
\multicolumn{1}{c}{\textbf{Reconstruction}} &
\multicolumn{2}{c}{\textbf{Ground-truth Recovery}} &
\multicolumn{2}{c}{\textbf{Cross-seed Consistency}} &
\multicolumn{3}{c}{\textbf{Dictionary Utilization}} \\
\cmidrule(lr){3-3}
\cmidrule(lr){4-4}
\cmidrule(lr){5-6}
\cmidrule(lr){7-8}
\cmidrule(lr){9-11}
&
&
$\ell_0$ &
$R^2$ &
DCS &
MCC &
DCS &
MCC &
\# Dup &
\# Dead &
Usable atom frac. \\
\midrule
\multirow{7}{*}{$\beta = 0.0$} & ReLU & 67.0\,{\scriptsize $\pm$0.1} & 0.67\,{\scriptsize $\pm$0.00} & 0.98\,{\scriptsize $\pm$0.00} & 0.83\,{\scriptsize $\pm$0.00} & 0.97\,{\scriptsize $\pm$0.00} & 0.97\,{\scriptsize $\pm$0.00} & 10\,{\scriptsize $\pm$2} & 0\,{\scriptsize $\pm$0} & 1.00\,{\scriptsize $\pm$0.00} \\
 & TopK & 60.0\,{\scriptsize $\pm$0.0} & 0.82\,{\scriptsize $\pm$0.00} & 0.98\,{\scriptsize $\pm$0.00} & 0.86\,{\scriptsize $\pm$0.00} & 0.97\,{\scriptsize $\pm$0.00} & 0.93\,{\scriptsize $\pm$0.00} & 1\,{\scriptsize $\pm$1} & 0\,{\scriptsize $\pm$0} & 1.00\,{\scriptsize $\pm$0.00} \\
 & BatchTopK & 61.0\,{\scriptsize $\pm$0.0} & 0.81\,{\scriptsize $\pm$0.00} & 0.98\,{\scriptsize $\pm$0.00} & 0.86\,{\scriptsize $\pm$0.00} & 0.96\,{\scriptsize $\pm$0.00} & 0.93\,{\scriptsize $\pm$0.00} & 1\,{\scriptsize $\pm$1} & 0\,{\scriptsize $\pm$0} & 1.00\,{\scriptsize $\pm$0.00} \\
 & Matryoshka & 64.0\,{\scriptsize $\pm$0.0} & 0.69\,{\scriptsize $\pm$0.00} & 0.82\,{\scriptsize $\pm$0.00} & 0.67\,{\scriptsize $\pm$0.00} & 0.69\,{\scriptsize $\pm$0.00} & 0.58\,{\scriptsize $\pm$0.00} & 1\,{\scriptsize $\pm$1} & 0\,{\scriptsize $\pm$0} & 1.00\,{\scriptsize $\pm$0.00} \\
 & MP-SAE & 57.7\,{\scriptsize $\pm$0.0} & 0.95\,{\scriptsize $\pm$0.00} & 0.99\,{\scriptsize $\pm$0.00} & 0.89\,{\scriptsize $\pm$0.00} & 0.99\,{\scriptsize $\pm$0.00} & 0.95\,{\scriptsize $\pm$0.00} & 20\,{\scriptsize $\pm$4} & 0\,{\scriptsize $\pm$0} & 1.00\,{\scriptsize $\pm$0.00} \\
 & iSAE & 60.0\,{\scriptsize $\pm$0.0} & 0.72\,{\scriptsize $\pm$0.00} & 0.86\,{\scriptsize $\pm$0.00} & 0.36\,{\scriptsize $\pm$0.01} & 0.75\,{\scriptsize $\pm$0.00} & 0.32\,{\scriptsize $\pm$0.01} & 0\,{\scriptsize $\pm$0} & 440\,{\scriptsize $\pm$16} & 0.86\,{\scriptsize $\pm$0.01} \\
 & Ours & 61.0\,{\scriptsize $\pm$0.0} & \textbf{1.00}\,{\scriptsize $\pm$0.00} & \textbf{1.00}\,{\scriptsize $\pm$0.00} & \textbf{1.00}\,{\scriptsize $\pm$0.00} & \textbf{1.00}\,{\scriptsize $\pm$0.00} & \textbf{1.00}\,{\scriptsize $\pm$0.00} & 0\,{\scriptsize $\pm$0} & 0\,{\scriptsize $\pm$0} & 1.00\,{\scriptsize $\pm$0.00} \\
\cmidrule(lr){1-11}
\multirow{7}{*}{$\beta = 0.5$} & ReLU & 113\,{\scriptsize $\pm$1} & 0.77\,{\scriptsize $\pm$0.00} & 0.97\,{\scriptsize $\pm$0.00} & 0.83\,{\scriptsize $\pm$0.00} & 0.95\,{\scriptsize $\pm$0.00} & 0.94\,{\scriptsize $\pm$0.00} & 23\,{\scriptsize $\pm$9} & 0\,{\scriptsize $\pm$0} & 1.00\,{\scriptsize $\pm$0.00} \\
 & TopK & 90.0\,{\scriptsize $\pm$0.0} & 0.79\,{\scriptsize $\pm$0.00} & 0.94\,{\scriptsize $\pm$0.00} & 0.79\,{\scriptsize $\pm$0.00} & 0.91\,{\scriptsize $\pm$0.00} & 0.86\,{\scriptsize $\pm$0.00} & 0\,{\scriptsize $\pm$0} & 0\,{\scriptsize $\pm$0} & 1.00\,{\scriptsize $\pm$0.00} \\
 & BatchTopK & 113\,{\scriptsize $\pm$1} & 0.79\,{\scriptsize $\pm$0.00} & 0.93\,{\scriptsize $\pm$0.00} & 0.77\,{\scriptsize $\pm$0.00} & 0.89\,{\scriptsize $\pm$0.00} & 0.84\,{\scriptsize $\pm$0.00} & 0\,{\scriptsize $\pm$1} & 0\,{\scriptsize $\pm$0} & 1.00\,{\scriptsize $\pm$0.00} \\
 & Matryoshka & 66.0\,{\scriptsize $\pm$0.3} & 0.63\,{\scriptsize $\pm$0.00} & 0.78\,{\scriptsize $\pm$0.00} & 0.59\,{\scriptsize $\pm$0.00} & 0.68\,{\scriptsize $\pm$0.00} & 0.56\,{\scriptsize $\pm$0.00} & 1\,{\scriptsize $\pm$1} & 0\,{\scriptsize $\pm$0} & 1.00\,{\scriptsize $\pm$0.00} \\
 & MP-SAE & 59.0\,{\scriptsize $\pm$0.0} & 0.90\,{\scriptsize $\pm$0.00} & 0.93\,{\scriptsize $\pm$0.00} & 0.88\,{\scriptsize $\pm$0.00} & 0.91\,{\scriptsize $\pm$0.00} & 0.84\,{\scriptsize $\pm$0.00} & 294\,{\scriptsize $\pm$20} & 0\,{\scriptsize $\pm$0} & 0.95\,{\scriptsize $\pm$0.00} \\
 & iSAE & 60.0\,{\scriptsize $\pm$0.0} & 0.62\,{\scriptsize $\pm$0.00} & 0.74\,{\scriptsize $\pm$0.00} & 0.27\,{\scriptsize $\pm$0.00} & 0.64\,{\scriptsize $\pm$0.00} & 0.25\,{\scriptsize $\pm$0.00} & 0\,{\scriptsize $\pm$0} & 737\,{\scriptsize $\pm$11} & 0.76\,{\scriptsize $\pm$0.00} \\
 & Ours & 62.0\,{\scriptsize $\pm$0.0} & \textbf{1.00}\,{\scriptsize $\pm$0.00} & \textbf{1.00}\,{\scriptsize $\pm$0.00} & \textbf{1.00}\,{\scriptsize $\pm$0.00} & \textbf{1.00}\,{\scriptsize $\pm$0.00} & \textbf{1.00}\,{\scriptsize $\pm$0.00} & 0\,{\scriptsize $\pm$0} & 0\,{\scriptsize $\pm$0} & 1.00\,{\scriptsize $\pm$0.00} \\
\cmidrule(lr){1-11}
\multirow{7}{*}{$\beta = 0.7$} & ReLU & 120\,{\scriptsize $\pm$2} & 0.57\,{\scriptsize $\pm$0.01} & 0.70\,{\scriptsize $\pm$0.00} & 0.56\,{\scriptsize $\pm$0.00} & 0.75\,{\scriptsize $\pm$0.00} & 0.61\,{\scriptsize $\pm$0.00} & 15\,{\scriptsize $\pm$2} & 13\,{\scriptsize $\pm$1} & 0.99\,{\scriptsize $\pm$0.00} \\
 & TopK & 90.0\,{\scriptsize $\pm$0.0} & 0.69\,{\scriptsize $\pm$0.01} & 0.89\,{\scriptsize $\pm$0.00} & 0.72\,{\scriptsize $\pm$0.00} & 0.83\,{\scriptsize $\pm$0.00} & 0.76\,{\scriptsize $\pm$0.00} & 0\,{\scriptsize $\pm$0} & 0\,{\scriptsize $\pm$0} & 1.00\,{\scriptsize $\pm$0.00} \\
 & BatchTopK & 65.0\,{\scriptsize $\pm$0.4} & 0.59\,{\scriptsize $\pm$0.01} & 0.84\,{\scriptsize $\pm$0.00} & 0.66\,{\scriptsize $\pm$0.00} & 0.78\,{\scriptsize $\pm$0.00} & 0.68\,{\scriptsize $\pm$0.00} & 2\,{\scriptsize $\pm$1} & 0\,{\scriptsize $\pm$0} & 1.00\,{\scriptsize $\pm$0.00} \\
 & Matryoshka & 66.0\,{\scriptsize $\pm$0.9} & 0.54\,{\scriptsize $\pm$0.01} & 0.71\,{\scriptsize $\pm$0.00} & 0.51\,{\scriptsize $\pm$0.00} & 0.63\,{\scriptsize $\pm$0.00} & 0.51\,{\scriptsize $\pm$0.00} & 1\,{\scriptsize $\pm$0} & 3\,{\scriptsize $\pm$2} & 1.00\,{\scriptsize $\pm$0.00} \\
 & MP-SAE & 59.7\,{\scriptsize $\pm$0.0} & 0.09\,{\scriptsize $\pm$0.01} & 0.25\,{\scriptsize $\pm$0.00} & 0.12\,{\scriptsize $\pm$0.00} & 0.60\,{\scriptsize $\pm$0.00} & 0.12\,{\scriptsize $\pm$0.00} & 718\,{\scriptsize $\pm$12} & 0\,{\scriptsize $\pm$0} & 0.88\,{\scriptsize $\pm$0.00} \\
 & iSAE & 60.0\,{\scriptsize $\pm$0.0} & 0.54\,{\scriptsize $\pm$0.01} & 0.69\,{\scriptsize $\pm$0.00} & 0.23\,{\scriptsize $\pm$0.01} & 0.61\,{\scriptsize $\pm$0.00} & 0.24\,{\scriptsize $\pm$0.00} & 0\,{\scriptsize $\pm$0} & 955\,{\scriptsize $\pm$5} & 0.69\,{\scriptsize $\pm$0.00} \\
 & Ours & 71.0\,{\scriptsize $\pm$1.4} & \textbf{0.96}\,{\scriptsize $\pm$0.01} & \textbf{0.97}\,{\scriptsize $\pm$0.00} & \textbf{0.97}\,{\scriptsize $\pm$0.00} & \textbf{0.96}\,{\scriptsize $\pm$0.00} & \textbf{0.95}\,{\scriptsize $\pm$0.00} & 0\,{\scriptsize $\pm$0} & 0\,{\scriptsize $\pm$0} & 1.00\,{\scriptsize $\pm$0.00} \\
\cmidrule(lr){1-11}
\multirow{7}{*}{$\beta = 1.0$} & ReLU & 102\,{\scriptsize $\pm$4} & -0.07\,{\scriptsize $\pm$0.00} & 0.33\,{\scriptsize $\pm$0.01} & 0.18\,{\scriptsize $\pm$0.01} & 0.39\,{\scriptsize $\pm$0.00} & 0.21\,{\scriptsize $\pm$0.00} & 56\,{\scriptsize $\pm$8} & 275\,{\scriptsize $\pm$34} & 0.90\,{\scriptsize $\pm$0.01} \\
 & TopK & 60.0\,{\scriptsize $\pm$0.0} & 0.52\,{\scriptsize $\pm$0.04} & 0.75\,{\scriptsize $\pm$0.00} & 0.58\,{\scriptsize $\pm$0.00} & 0.69\,{\scriptsize $\pm$0.00} & 0.58\,{\scriptsize $\pm$0.00} & 2\,{\scriptsize $\pm$1} & 22\,{\scriptsize $\pm$3} & 0.99\,{\scriptsize $\pm$0.00} \\
 & BatchTopK & 69.0\,{\scriptsize $\pm$1.2} & 0.13\,{\scriptsize $\pm$0.13} & 0.74\,{\scriptsize $\pm$0.00} & 0.57\,{\scriptsize $\pm$0.00} & 0.68\,{\scriptsize $\pm$0.00} & 0.57\,{\scriptsize $\pm$0.01} & 4\,{\scriptsize $\pm$0} & 5\,{\scriptsize $\pm$3} & \textbf{1.00}\,{\scriptsize $\pm$0.00} \\
 & Matryoshka & 67.0\,{\scriptsize $\pm$1.5} & 0.44\,{\scriptsize $\pm$0.01} & 0.70\,{\scriptsize $\pm$0.00} & 0.51\,{\scriptsize $\pm$0.00} & 0.64\,{\scriptsize $\pm$0.00} & 0.52\,{\scriptsize $\pm$0.00} & 6\,{\scriptsize $\pm$0} & 139\,{\scriptsize $\pm$8} & 0.95\,{\scriptsize $\pm$0.00} \\
 & MP-SAE & 59.6\,{\scriptsize $\pm$0.0} & -0.31\,{\scriptsize $\pm$0.01} & 0.20\,{\scriptsize $\pm$0.00} & 0.06\,{\scriptsize $\pm$0.00} & 0.56\,{\scriptsize $\pm$0.00} & 0.06\,{\scriptsize $\pm$0.00} & 780\,{\scriptsize $\pm$4} & \textbf{4}\,{\scriptsize $\pm$1} & 0.86\,{\scriptsize $\pm$0.00} \\
 & iSAE & 60.0\,{\scriptsize $\pm$0.0} & 0.47\,{\scriptsize $\pm$0.01} & 0.71\,{\scriptsize $\pm$0.00} & 0.24\,{\scriptsize $\pm$0.00} & 0.66\,{\scriptsize $\pm$0.00} & 0.26\,{\scriptsize $\pm$0.00} & 0\,{\scriptsize $\pm$0} & 1035\,{\scriptsize $\pm$10} & 0.66\,{\scriptsize $\pm$0.00} \\
 & Ours & 103\,{\scriptsize $\pm$1} & \textbf{0.98}\,{\scriptsize $\pm$0.00} & \textbf{0.94}\,{\scriptsize $\pm$0.00} & \textbf{0.93}\,{\scriptsize $\pm$0.00} & \textbf{0.93}\,{\scriptsize $\pm$0.00} & \textbf{0.90}\,{\scriptsize $\pm$0.00} & 0\,{\scriptsize $\pm$0} & 13\,{\scriptsize $\pm$1} & 1.00\,{\scriptsize $\pm$0.00} \\
\bottomrule
\end{tabular}%
}
\end{table*}

\begin{table*}[p]
\centering
\caption{
\textbf{Full synthetic results with a 6K-atom ground-truth dictionary across tail exponents $\beta$.}
We evaluate all methods across $\beta\in\{0,0.5,0.7,1.0\}$.
We report sparsity $\ell_0$, reconstruction ($R^2$), ground-truth recovery
(\texttt{DCS}, \texttt{MCC}), cross-seed consistency, and dictionary utilization
measured by the usable atom fraction. The numbers of dead and duplicated atoms
are also reported. Metrics are averaged across training seeds, while cross-seed
\texttt{DCS} and \texttt{MCC} are averaged across seed pairs.
$\pm$ denotes the corresponding standard deviation.
}
\label{tab:synthetic_6k}
\resizebox{\textwidth}{!}{%
\begin{tabular}{ll c c cc cc ccc}
\toprule
\multirow{2}{*}{\textbf{Setting}} &
\multirow{2}{*}{\textbf{Architecture}} &
\multicolumn{1}{c}{\textbf{Sparsity}} &
\multicolumn{1}{c}{\textbf{Reconstruction}} &
\multicolumn{2}{c}{\textbf{Ground-truth Recovery}} &
\multicolumn{2}{c}{\textbf{Cross-seed Consistency}} &
\multicolumn{3}{c}{\textbf{Dictionary Utilization}} \\
\cmidrule(lr){3-3}
\cmidrule(lr){4-4}
\cmidrule(lr){5-6}
\cmidrule(lr){7-8}
\cmidrule(lr){9-11}
&
&
$\ell_0$ &
$R^2$ &
DCS &
MCC &
DCS &
MCC &
\# Dup &
\# Dead &
Usable atom frac. \\
\midrule
\multirow{7}{*}{$\beta = 0.0$} & ReLU & 76.0\,{\scriptsize $\pm$0.3} & 0.64\,{\scriptsize $\pm$0.00} & 0.97\,{\scriptsize $\pm$0.00} & 0.78\,{\scriptsize $\pm$0.00} & 0.94\,{\scriptsize $\pm$0.00} & 0.94\,{\scriptsize $\pm$0.00} & 44\,{\scriptsize $\pm$5} & 0\,{\scriptsize $\pm$0} & 1.00\,{\scriptsize $\pm$0.00} \\
 & TopK & 60.0\,{\scriptsize $\pm$0.0} & 0.65\,{\scriptsize $\pm$0.00} & 0.81\,{\scriptsize $\pm$0.00} & 0.61\,{\scriptsize $\pm$0.00} & 0.67\,{\scriptsize $\pm$0.00} & 0.53\,{\scriptsize $\pm$0.00} & 2\,{\scriptsize $\pm$1} & 0\,{\scriptsize $\pm$0} & 1.00\,{\scriptsize $\pm$0.00} \\
 & BatchTopK & 71.0\,{\scriptsize $\pm$0.0} & 0.63\,{\scriptsize $\pm$0.00} & 0.76\,{\scriptsize $\pm$0.00} & 0.55\,{\scriptsize $\pm$0.00} & 0.59\,{\scriptsize $\pm$0.00} & 0.44\,{\scriptsize $\pm$0.00} & 1\,{\scriptsize $\pm$1} & 0\,{\scriptsize $\pm$0} & 1.00\,{\scriptsize $\pm$0.00} \\
 & Matryoshka & 75.0\,{\scriptsize $\pm$0.0} & 0.48\,{\scriptsize $\pm$0.00} & 0.35\,{\scriptsize $\pm$0.00} & 0.17\,{\scriptsize $\pm$0.00} & 0.20\,{\scriptsize $\pm$0.00} & 0.05\,{\scriptsize $\pm$0.00} & 0\,{\scriptsize $\pm$0} & 0\,{\scriptsize $\pm$0} & 1.00\,{\scriptsize $\pm$0.00} \\
 & MP-SAE & 58.0\,{\scriptsize $\pm$0.0} & \textbf{0.95}\,{\scriptsize $\pm$0.00} & 0.99\,{\scriptsize $\pm$0.00} & \textbf{0.98}\,{\scriptsize $\pm$0.00} & 0.98\,{\scriptsize $\pm$0.00} & 0.98\,{\scriptsize $\pm$0.00} & 51\,{\scriptsize $\pm$12} & 0\,{\scriptsize $\pm$0} & 1.00\,{\scriptsize $\pm$0.00} \\
 & iSAE & 60.0\,{\scriptsize $\pm$0.0} & 0.41\,{\scriptsize $\pm$0.00} & 0.19\,{\scriptsize $\pm$0.00} & 0.03\,{\scriptsize $\pm$0.00} & 0.15\,{\scriptsize $\pm$0.00} & 0.01\,{\scriptsize $\pm$0.00} & 0\,{\scriptsize $\pm$0} & 0\,{\scriptsize $\pm$0} & 1.00\,{\scriptsize $\pm$0.00} \\
 & Ours & 60.0\,{\scriptsize $\pm$0.0} & 0.92\,{\scriptsize $\pm$0.00} & \textbf{0.99}\,{\scriptsize $\pm$0.00} & 0.97\,{\scriptsize $\pm$0.00} & \textbf{0.99}\,{\scriptsize $\pm$0.00} & \textbf{1.00}\,{\scriptsize $\pm$0.00} & 0\,{\scriptsize $\pm$0} & 0\,{\scriptsize $\pm$0} & 1.00\,{\scriptsize $\pm$0.00} \\
\cmidrule(lr){1-11}
\multirow{7}{*}{$\beta = 0.5$} & ReLU & 116\,{\scriptsize $\pm$0} & 0.67\,{\scriptsize $\pm$0.00} & 0.90\,{\scriptsize $\pm$0.00} & 0.71\,{\scriptsize $\pm$0.00} & 0.85\,{\scriptsize $\pm$0.00} & 0.81\,{\scriptsize $\pm$0.00} & 30\,{\scriptsize $\pm$2} & 0\,{\scriptsize $\pm$0} & 1.00\,{\scriptsize $\pm$0.00} \\
 & TopK & 90.0\,{\scriptsize $\pm$0.0} & 0.67\,{\scriptsize $\pm$0.00} & 0.80\,{\scriptsize $\pm$0.00} & 0.58\,{\scriptsize $\pm$0.00} & 0.72\,{\scriptsize $\pm$0.00} & 0.61\,{\scriptsize $\pm$0.00} & 1\,{\scriptsize $\pm$0} & 0\,{\scriptsize $\pm$0} & 1.00\,{\scriptsize $\pm$0.00} \\
 & BatchTopK & 119\,{\scriptsize $\pm$0} & 0.66\,{\scriptsize $\pm$0.00} & 0.76\,{\scriptsize $\pm$0.00} & 0.53\,{\scriptsize $\pm$0.00} & 0.67\,{\scriptsize $\pm$0.00} & 0.56\,{\scriptsize $\pm$0.00} & 1\,{\scriptsize $\pm$0} & 0\,{\scriptsize $\pm$0} & 1.00\,{\scriptsize $\pm$0.00} \\
 & Matryoshka & 75.0\,{\scriptsize $\pm$0.5} & 0.51\,{\scriptsize $\pm$0.00} & 0.57\,{\scriptsize $\pm$0.00} & 0.34\,{\scriptsize $\pm$0.00} & 0.47\,{\scriptsize $\pm$0.00} & 0.33\,{\scriptsize $\pm$0.00} & 5\,{\scriptsize $\pm$1} & 0\,{\scriptsize $\pm$0} & 1.00\,{\scriptsize $\pm$0.00} \\
 & MP-SAE & 59.2\,{\scriptsize $\pm$0.1} & 0.91\,{\scriptsize $\pm$0.00} & 0.94\,{\scriptsize $\pm$0.00} & 0.89\,{\scriptsize $\pm$0.00} & 0.92\,{\scriptsize $\pm$0.00} & 0.83\,{\scriptsize $\pm$0.00} & 532\,{\scriptsize $\pm$17} & 0\,{\scriptsize $\pm$0} & 0.95\,{\scriptsize $\pm$0.00} \\
 & iSAE & 60.0\,{\scriptsize $\pm$0.0} & 0.45\,{\scriptsize $\pm$0.00} & 0.39\,{\scriptsize $\pm$0.00} & 0.09\,{\scriptsize $\pm$0.00} & 0.35\,{\scriptsize $\pm$0.00} & 0.09\,{\scriptsize $\pm$0.00} & 0\,{\scriptsize $\pm$0} & 0\,{\scriptsize $\pm$0} & 1.00\,{\scriptsize $\pm$0.00} \\
 & Ours & 61.0\,{\scriptsize $\pm$0.3} & \textbf{0.95}\,{\scriptsize $\pm$0.00} & \textbf{0.99}\,{\scriptsize $\pm$0.00} & \textbf{0.99}\,{\scriptsize $\pm$0.00} & \textbf{0.99}\,{\scriptsize $\pm$0.00} & \textbf{1.00}\,{\scriptsize $\pm$0.00} & 0\,{\scriptsize $\pm$0} & 0\,{\scriptsize $\pm$0} & 1.00\,{\scriptsize $\pm$0.00} \\
\cmidrule(lr){1-11}
\multirow{7}{*}{$\beta = 0.7$} & ReLU & 73.0\,{\scriptsize $\pm$3.6} & 0.17\,{\scriptsize $\pm$0.00} & 0.37\,{\scriptsize $\pm$0.00} & 0.20\,{\scriptsize $\pm$0.00} & 0.43\,{\scriptsize $\pm$0.00} & 0.22\,{\scriptsize $\pm$0.00} & 50\,{\scriptsize $\pm$5} & 20\,{\scriptsize $\pm$1} & 0.99\,{\scriptsize $\pm$0.00} \\
 & TopK & 90.0\,{\scriptsize $\pm$0.0} & 0.55\,{\scriptsize $\pm$0.00} & 0.70\,{\scriptsize $\pm$0.00} & 0.46\,{\scriptsize $\pm$0.00} & 0.62\,{\scriptsize $\pm$0.00} & 0.50\,{\scriptsize $\pm$0.00} & 0\,{\scriptsize $\pm$0} & 0\,{\scriptsize $\pm$0} & 1.00\,{\scriptsize $\pm$0.00} \\
 & BatchTopK & 77.0\,{\scriptsize $\pm$0.5} & 0.47\,{\scriptsize $\pm$0.00} & 0.59\,{\scriptsize $\pm$0.00} & 0.35\,{\scriptsize $\pm$0.00} & 0.52\,{\scriptsize $\pm$0.00} & 0.39\,{\scriptsize $\pm$0.00} & 1\,{\scriptsize $\pm$0} & 0\,{\scriptsize $\pm$0} & 1.00\,{\scriptsize $\pm$0.00} \\
 & Matryoshka & 77.0\,{\scriptsize $\pm$0.3} & 0.43\,{\scriptsize $\pm$0.00} & 0.49\,{\scriptsize $\pm$0.00} & 0.26\,{\scriptsize $\pm$0.00} & 0.43\,{\scriptsize $\pm$0.00} & 0.29\,{\scriptsize $\pm$0.00} & 1\,{\scriptsize $\pm$1} & 0\,{\scriptsize $\pm$0} & 1.00\,{\scriptsize $\pm$0.00} \\
 & MP-SAE & 59.9\,{\scriptsize $\pm$0.0} & 0.19\,{\scriptsize $\pm$0.01} & 0.25\,{\scriptsize $\pm$0.00} & 0.11\,{\scriptsize $\pm$0.00} & 0.64\,{\scriptsize $\pm$0.00} & 0.10\,{\scriptsize $\pm$0.00} & 1479\,{\scriptsize $\pm$31} & 0\,{\scriptsize $\pm$0} & 0.87\,{\scriptsize $\pm$0.00} \\
 & iSAE & 60.0\,{\scriptsize $\pm$0.0} & 0.40\,{\scriptsize $\pm$0.00} & 0.39\,{\scriptsize $\pm$0.00} & 0.08\,{\scriptsize $\pm$0.00} & 0.35\,{\scriptsize $\pm$0.00} & 0.09\,{\scriptsize $\pm$0.00} & 0\,{\scriptsize $\pm$0} & 3746\,{\scriptsize $\pm$21} & 0.39\,{\scriptsize $\pm$0.00} \\
 & Ours & 67.0\,{\scriptsize $\pm$1.5} & \textbf{0.99}\,{\scriptsize $\pm$0.00} & \textbf{0.99}\,{\scriptsize $\pm$0.00} & \textbf{1.00}\,{\scriptsize $\pm$0.00} & \textbf{0.99}\,{\scriptsize $\pm$0.00} & \textbf{0.99}\,{\scriptsize $\pm$0.00} & 0\,{\scriptsize $\pm$0} & 0\,{\scriptsize $\pm$0} & 1.00\,{\scriptsize $\pm$0.00} \\
\cmidrule(lr){1-11}
\multirow{7}{*}{$\beta = 1.0$} & ReLU & 56.0\,{\scriptsize $\pm$12.1} & -0.38\,{\scriptsize $\pm$0.01} & 0.16\,{\scriptsize $\pm$0.00} & 0.02\,{\scriptsize $\pm$0.00} & 0.17\,{\scriptsize $\pm$0.00} & 0.03\,{\scriptsize $\pm$0.00} & 8\,{\scriptsize $\pm$2} & 135\,{\scriptsize $\pm$101} & 0.98\,{\scriptsize $\pm$0.02} \\
 & TopK & 60.0\,{\scriptsize $\pm$0.0} & 0.35\,{\scriptsize $\pm$0.11} & 0.55\,{\scriptsize $\pm$0.00} & 0.33\,{\scriptsize $\pm$0.00} & 0.51\,{\scriptsize $\pm$0.00} & 0.36\,{\scriptsize $\pm$0.00} & 1\,{\scriptsize $\pm$1} & 200\,{\scriptsize $\pm$10} & 0.97\,{\scriptsize $\pm$0.00} \\
 & BatchTopK & 72.0\,{\scriptsize $\pm$1.0} & 0.14\,{\scriptsize $\pm$0.02} & 0.53\,{\scriptsize $\pm$0.00} & 0.26\,{\scriptsize $\pm$0.00} & 0.48\,{\scriptsize $\pm$0.00} & 0.37\,{\scriptsize $\pm$0.00} & 2\,{\scriptsize $\pm$0} & 62\,{\scriptsize $\pm$4} & 0.99\,{\scriptsize $\pm$0.00} \\
 & Matryoshka & 75.0\,{\scriptsize $\pm$0.8} & 0.24\,{\scriptsize $\pm$0.01} & 0.43\,{\scriptsize $\pm$0.00} & 0.19\,{\scriptsize $\pm$0.00} & 0.40\,{\scriptsize $\pm$0.00} & 0.27\,{\scriptsize $\pm$0.00} & 5\,{\scriptsize $\pm$1} & 202\,{\scriptsize $\pm$18} & 0.97\,{\scriptsize $\pm$0.00} \\
 & MP-SAE & 59.7\,{\scriptsize $\pm$0.0} & -0.18\,{\scriptsize $\pm$0.01} & 0.19\,{\scriptsize $\pm$0.00} & 0.04\,{\scriptsize $\pm$0.00} & 0.61\,{\scriptsize $\pm$0.01} & 0.05\,{\scriptsize $\pm$0.00} & 1490\,{\scriptsize $\pm$32} & \textbf{11}\,{\scriptsize $\pm$3} & 0.87\,{\scriptsize $\pm$0.00} \\
 & iSAE & 60.0\,{\scriptsize $\pm$0.0} & 0.25\,{\scriptsize $\pm$0.01} & 0.41\,{\scriptsize $\pm$0.00} & 0.08\,{\scriptsize $\pm$0.00} & 0.39\,{\scriptsize $\pm$0.00} & 0.11\,{\scriptsize $\pm$0.00} & 0\,{\scriptsize $\pm$0} & 3994\,{\scriptsize $\pm$10} & 0.35\,{\scriptsize $\pm$0.00} \\
 & Ours & 106\,{\scriptsize $\pm$4} & \textbf{0.98}\,{\scriptsize $\pm$0.00} & \textbf{0.95}\,{\scriptsize $\pm$0.00} & \textbf{0.95}\,{\scriptsize $\pm$0.00} & \textbf{0.95}\,{\scriptsize $\pm$0.00} & \textbf{0.93}\,{\scriptsize $\pm$0.00} & 0\,{\scriptsize $\pm$0} & 23\,{\scriptsize $\pm$7} & \textbf{1.00}\,{\scriptsize $\pm$0.00} \\
\bottomrule
\end{tabular}%
}
\end{table*}

\begin{table*}[p]
\centering
\caption{
\textbf{Full synthetic results with a 9K-atom ground-truth dictionary across tail exponents $\beta$.}
We evaluate all methods across $\beta\in\{0,0.5,0.7,1.0\}$.
We report sparsity $\ell_0$, reconstruction ($R^2$), ground-truth recovery
(\texttt{DCS}, \texttt{MCC}), cross-seed consistency, and dictionary utilization
measured by the usable atom fraction. The numbers of dead and duplicated atoms
are also reported. Metrics are averaged across training seeds, while cross-seed
\texttt{DCS} and \texttt{MCC} are averaged across seed pairs.
$\pm$ denotes the corresponding standard deviation.
}
\label{tab:synthetic_9k}
\resizebox{\textwidth}{!}{%
\begin{tabular}{ll c c cc cc ccc}
\toprule
\multirow{2}{*}{\textbf{Setting}} &
\multirow{2}{*}{\textbf{Architecture}} &
\multicolumn{1}{c}{\textbf{Sparsity}} &
\multicolumn{1}{c}{\textbf{Reconstruction}} &
\multicolumn{2}{c}{\textbf{Ground-truth Recovery}} &
\multicolumn{2}{c}{\textbf{Cross-seed Consistency}} &
\multicolumn{3}{c}{\textbf{Dictionary Utilization}} \\
\cmidrule(lr){3-3}
\cmidrule(lr){4-4}
\cmidrule(lr){5-6}
\cmidrule(lr){7-8}
\cmidrule(lr){9-11}
&
&
$\ell_0$ &
$R^2$ &
DCS &
MCC &
DCS &
MCC &
\# Dup &
\# Dead &
Usable atom frac. \\
\midrule
\multirow{7}{*}{$\beta = 0.0$} & ReLU & 63.0\,{\scriptsize $\pm$0.0} & 0.44\,{\scriptsize $\pm$0.00} & 0.78\,{\scriptsize $\pm$0.00} & 0.55\,{\scriptsize $\pm$0.00} & 0.62\,{\scriptsize $\pm$0.00} & 0.53\,{\scriptsize $\pm$0.00} & 66\,{\scriptsize $\pm$5} & 0\,{\scriptsize $\pm$0} & 1.00\,{\scriptsize $\pm$0.00} \\
 & TopK & 60.0\,{\scriptsize $\pm$0.0} & 0.41\,{\scriptsize $\pm$0.00} & 0.23\,{\scriptsize $\pm$0.00} & 0.05\,{\scriptsize $\pm$0.00} & 0.15\,{\scriptsize $\pm$0.00} & 0.01\,{\scriptsize $\pm$0.00} & 0\,{\scriptsize $\pm$0} & 0\,{\scriptsize $\pm$0} & 1.00\,{\scriptsize $\pm$0.00} \\
 & BatchTopK & 89.0\,{\scriptsize $\pm$0.0} & 0.45\,{\scriptsize $\pm$0.00} & 0.22\,{\scriptsize $\pm$0.00} & 0.05\,{\scriptsize $\pm$0.00} & 0.15\,{\scriptsize $\pm$0.00} & 0.02\,{\scriptsize $\pm$0.00} & 0\,{\scriptsize $\pm$0} & 0\,{\scriptsize $\pm$0} & 1.00\,{\scriptsize $\pm$0.00} \\
 & Matryoshka & 80.0\,{\scriptsize $\pm$0.4} & 0.44\,{\scriptsize $\pm$0.00} & 0.19\,{\scriptsize $\pm$0.00} & 0.03\,{\scriptsize $\pm$0.00} & 0.14\,{\scriptsize $\pm$0.00} & 0.01\,{\scriptsize $\pm$0.00} & 0\,{\scriptsize $\pm$0} & 0\,{\scriptsize $\pm$0} & 1.00\,{\scriptsize $\pm$0.00} \\
 & MP-SAE & 60.0\,{\scriptsize $\pm$0.0} & 0.47\,{\scriptsize $\pm$0.00} & 0.17\,{\scriptsize $\pm$0.00} & 0.01\,{\scriptsize $\pm$0.00} & 0.18\,{\scriptsize $\pm$0.00} & 0.00\,{\scriptsize $\pm$0.00} & 0\,{\scriptsize $\pm$0} & 0\,{\scriptsize $\pm$0} & 1.00\,{\scriptsize $\pm$0.00} \\
 & iSAE & 60.0\,{\scriptsize $\pm$0.0} & 0.40\,{\scriptsize $\pm$0.00} & 0.16\,{\scriptsize $\pm$0.00} & 0.01\,{\scriptsize $\pm$0.00} & 0.15\,{\scriptsize $\pm$0.00} & 0.01\,{\scriptsize $\pm$0.00} & 0\,{\scriptsize $\pm$0} & 0\,{\scriptsize $\pm$0} & 1.00\,{\scriptsize $\pm$0.00} \\
 & Ours & 63.0\,{\scriptsize $\pm$0.0} & \textbf{0.93}\,{\scriptsize $\pm$0.00} & \textbf{0.99}\,{\scriptsize $\pm$0.00} & \textbf{0.98}\,{\scriptsize $\pm$0.00} & \textbf{0.99}\,{\scriptsize $\pm$0.00} & \textbf{1.00}\,{\scriptsize $\pm$0.00} & 0\,{\scriptsize $\pm$0} & 0\,{\scriptsize $\pm$0} & 1.00\,{\scriptsize $\pm$0.00} \\
\cmidrule(lr){1-11}
\multirow{7}{*}{$\beta = 0.5$} & ReLU & 76.0\,{\scriptsize $\pm$0.8} & 0.33\,{\scriptsize $\pm$0.00} & 0.51\,{\scriptsize $\pm$0.00} & 0.31\,{\scriptsize $\pm$0.00} & 0.48\,{\scriptsize $\pm$0.00} & 0.33\,{\scriptsize $\pm$0.00} & 30\,{\scriptsize $\pm$4} & 0\,{\scriptsize $\pm$0} & 1.00\,{\scriptsize $\pm$0.00} \\
 & TopK & 60.0\,{\scriptsize $\pm$0.0} & 0.46\,{\scriptsize $\pm$0.00} & 0.46\,{\scriptsize $\pm$0.00} & 0.23\,{\scriptsize $\pm$0.00} & 0.39\,{\scriptsize $\pm$0.00} & 0.23\,{\scriptsize $\pm$0.00} & 0\,{\scriptsize $\pm$0} & 0\,{\scriptsize $\pm$0} & 1.00\,{\scriptsize $\pm$0.00} \\
 & BatchTopK & 81.0\,{\scriptsize $\pm$0.0} & 0.48\,{\scriptsize $\pm$0.00} & 0.43\,{\scriptsize $\pm$0.00} & 0.21\,{\scriptsize $\pm$0.00} & 0.36\,{\scriptsize $\pm$0.00} & 0.21\,{\scriptsize $\pm$0.00} & 0\,{\scriptsize $\pm$0} & 0\,{\scriptsize $\pm$0} & 1.00\,{\scriptsize $\pm$0.00} \\
 & Matryoshka & 79.0\,{\scriptsize $\pm$0.0} & 0.45\,{\scriptsize $\pm$0.00} & 0.37\,{\scriptsize $\pm$0.00} & 0.15\,{\scriptsize $\pm$0.00} & 0.31\,{\scriptsize $\pm$0.00} & 0.16\,{\scriptsize $\pm$0.00} & 1\,{\scriptsize $\pm$0} & 0\,{\scriptsize $\pm$0} & 1.00\,{\scriptsize $\pm$0.00} \\
 & MP-SAE & 60.0\,{\scriptsize $\pm$0.0} & 0.42\,{\scriptsize $\pm$0.00} & 0.18\,{\scriptsize $\pm$0.00} & 0.03\,{\scriptsize $\pm$0.00} & 0.50\,{\scriptsize $\pm$0.00} & 0.02\,{\scriptsize $\pm$0.00} & 146\,{\scriptsize $\pm$4} & 0\,{\scriptsize $\pm$0} & 0.99\,{\scriptsize $\pm$0.00} \\
 & iSAE & 60.0\,{\scriptsize $\pm$0.0} & 0.41\,{\scriptsize $\pm$0.00} & 0.24\,{\scriptsize $\pm$0.00} & 0.03\,{\scriptsize $\pm$0.00} & 0.22\,{\scriptsize $\pm$0.00} & 0.03\,{\scriptsize $\pm$0.00} & 0\,{\scriptsize $\pm$0} & 0\,{\scriptsize $\pm$0} & 1.00\,{\scriptsize $\pm$0.00} \\
 & Ours & 64.0\,{\scriptsize $\pm$0.0} & \textbf{0.96}\,{\scriptsize $\pm$0.00} & \textbf{0.99}\,{\scriptsize $\pm$0.00} & \textbf{0.99}\,{\scriptsize $\pm$0.00} & \textbf{0.99}\,{\scriptsize $\pm$0.00} & \textbf{1.00}\,{\scriptsize $\pm$0.00} & 0\,{\scriptsize $\pm$0} & 0\,{\scriptsize $\pm$0} & 1.00\,{\scriptsize $\pm$0.00} \\
\cmidrule(lr){1-11}
\multirow{7}{*}{$\beta = 0.7$} & ReLU & 109\,{\scriptsize $\pm$8} & 0.26\,{\scriptsize $\pm$0.01} & 0.36\,{\scriptsize $\pm$0.00} & 0.18\,{\scriptsize $\pm$0.00} & 0.42\,{\scriptsize $\pm$0.00} & 0.21\,{\scriptsize $\pm$0.00} & 21\,{\scriptsize $\pm$1} & 9\,{\scriptsize $\pm$1} & 1.00\,{\scriptsize $\pm$0.00} \\
 & TopK & 90.0\,{\scriptsize $\pm$0.0} & 0.48\,{\scriptsize $\pm$0.00} & 0.51\,{\scriptsize $\pm$0.00} & 0.27\,{\scriptsize $\pm$0.00} & 0.45\,{\scriptsize $\pm$0.00} & 0.30\,{\scriptsize $\pm$0.00} & 0\,{\scriptsize $\pm$0} & 0\,{\scriptsize $\pm$0} & 1.00\,{\scriptsize $\pm$0.00} \\
 & BatchTopK & 78.0\,{\scriptsize $\pm$0.8} & 0.41\,{\scriptsize $\pm$0.01} & 0.44\,{\scriptsize $\pm$0.00} & 0.21\,{\scriptsize $\pm$0.00} & 0.39\,{\scriptsize $\pm$0.00} & 0.24\,{\scriptsize $\pm$0.00} & 0\,{\scriptsize $\pm$0} & 0\,{\scriptsize $\pm$0} & 1.00\,{\scriptsize $\pm$0.00} \\
 & Matryoshka & 78.0\,{\scriptsize $\pm$0.4} & 0.39\,{\scriptsize $\pm$0.00} & 0.37\,{\scriptsize $\pm$0.00} & 0.16\,{\scriptsize $\pm$0.00} & 0.33\,{\scriptsize $\pm$0.00} & 0.18\,{\scriptsize $\pm$0.00} & 1\,{\scriptsize $\pm$1} & 0\,{\scriptsize $\pm$0} & 1.00\,{\scriptsize $\pm$0.00} \\
 & MP-SAE & 60.0\,{\scriptsize $\pm$0.0} & 0.28\,{\scriptsize $\pm$0.00} & 0.20\,{\scriptsize $\pm$0.00} & 0.04\,{\scriptsize $\pm$0.00} & 0.61\,{\scriptsize $\pm$0.00} & 0.04\,{\scriptsize $\pm$0.00} & 1504\,{\scriptsize $\pm$13} & 0\,{\scriptsize $\pm$0} & 0.92\,{\scriptsize $\pm$0.00} \\
 & iSAE & 60.0\,{\scriptsize $\pm$0.0} & 0.37\,{\scriptsize $\pm$0.00} & 0.30\,{\scriptsize $\pm$0.00} & 0.04\,{\scriptsize $\pm$0.00} & 0.28\,{\scriptsize $\pm$0.00} & 0.05\,{\scriptsize $\pm$0.00} & 0\,{\scriptsize $\pm$0} & 6826\,{\scriptsize $\pm$50} & 0.26\,{\scriptsize $\pm$0.01} \\
 & Ours & 70.0\,{\scriptsize $\pm$0.4} & \textbf{0.98}\,{\scriptsize $\pm$0.00} & \textbf{0.99}\,{\scriptsize $\pm$0.00} & \textbf{0.99}\,{\scriptsize $\pm$0.00} & \textbf{0.99}\,{\scriptsize $\pm$0.00} & \textbf{0.99}\,{\scriptsize $\pm$0.00} & 0\,{\scriptsize $\pm$0} & 1\,{\scriptsize $\pm$1} & 1.00\,{\scriptsize $\pm$0.00} \\
\cmidrule(lr){1-11}
\multirow{7}{*}{$\beta = 1.0$} & ReLU & 126\,{\scriptsize $\pm$20} & -0.30\,{\scriptsize $\pm$0.01} & 0.18\,{\scriptsize $\pm$0.00} & 0.03\,{\scriptsize $\pm$0.00} & 0.20\,{\scriptsize $\pm$0.00} & 0.05\,{\scriptsize $\pm$0.00} & 37\,{\scriptsize $\pm$2} & 68\,{\scriptsize $\pm$29} & 0.99\,{\scriptsize $\pm$0.00} \\
 & TopK & 60.0\,{\scriptsize $\pm$0.0} & 0.11\,{\scriptsize $\pm$0.02} & 0.45\,{\scriptsize $\pm$0.00} & 0.23\,{\scriptsize $\pm$0.00} & 0.43\,{\scriptsize $\pm$0.00} & 0.26\,{\scriptsize $\pm$0.00} & 2\,{\scriptsize $\pm$2} & 573\,{\scriptsize $\pm$71} & 0.94\,{\scriptsize $\pm$0.01} \\
 & BatchTopK & 75.0\,{\scriptsize $\pm$4.3} & 0.16\,{\scriptsize $\pm$0.05} & 0.41\,{\scriptsize $\pm$0.00} & 0.18\,{\scriptsize $\pm$0.00} & 0.38\,{\scriptsize $\pm$0.00} & 0.22\,{\scriptsize $\pm$0.00} & 2\,{\scriptsize $\pm$1} & \textbf{1}\,{\scriptsize $\pm$2} & \textbf{1.00}\,{\scriptsize $\pm$0.00} \\
 & Matryoshka & 77.0\,{\scriptsize $\pm$1.1} & 0.20\,{\scriptsize $\pm$0.00} & 0.32\,{\scriptsize $\pm$0.00} & 0.11\,{\scriptsize $\pm$0.00} & 0.31\,{\scriptsize $\pm$0.00} & 0.15\,{\scriptsize $\pm$0.00} & 2\,{\scriptsize $\pm$1} & 7\,{\scriptsize $\pm$4} & 1.00\,{\scriptsize $\pm$0.00} \\
 & MP-SAE & 59.8\,{\scriptsize $\pm$0.0} & -0.05\,{\scriptsize $\pm$0.01} & 0.19\,{\scriptsize $\pm$0.00} & 0.03\,{\scriptsize $\pm$0.00} & 0.61\,{\scriptsize $\pm$0.00} & 0.05\,{\scriptsize $\pm$0.00} & 1906\,{\scriptsize $\pm$4} & 10\,{\scriptsize $\pm$2} & 0.89\,{\scriptsize $\pm$0.00} \\
 & iSAE & 60.0\,{\scriptsize $\pm$0.0} & 0.23\,{\scriptsize $\pm$0.01} & 0.31\,{\scriptsize $\pm$0.00} & 0.04\,{\scriptsize $\pm$0.00} & 0.29\,{\scriptsize $\pm$0.00} & 0.07\,{\scriptsize $\pm$0.00} & 0\,{\scriptsize $\pm$0} & 7082\,{\scriptsize $\pm$5} & 0.23\,{\scriptsize $\pm$0.00} \\
 & Ours & 128\,{\scriptsize $\pm$6} & \textbf{0.93}\,{\scriptsize $\pm$0.01} & \textbf{0.94}\,{\scriptsize $\pm$0.00} & \textbf{0.92}\,{\scriptsize $\pm$0.00} & \textbf{0.92}\,{\scriptsize $\pm$0.00} & \textbf{0.90}\,{\scriptsize $\pm$0.00} & 0\,{\scriptsize $\pm$0} & 83\,{\scriptsize $\pm$6} & 0.99\,{\scriptsize $\pm$0.00} \\
\bottomrule
\end{tabular}%
}
\end{table*}

\begin{table*}[p]
\centering
\caption{
\textbf{Full synthetic results with a 12K-atom ground-truth dictionary across tail exponents $\beta$.}
We evaluate all methods across $\beta\in\{0,0.5,0.7,1.0\}$.
We report sparsity $\ell_0$, reconstruction ($R^2$), ground-truth recovery
(\texttt{DCS}, \texttt{MCC}), cross-seed consistency, and dictionary utilization
measured by the usable atom fraction. The numbers of dead and duplicated atoms
are also reported. Metrics are averaged across training seeds, while cross-seed
\texttt{DCS} and \texttt{MCC} are averaged across seed pairs.
$\pm$ denotes the corresponding standard deviation.
}
\label{tab:synthetic_12k}
\resizebox{\textwidth}{!}{%
\begin{tabular}{ll c c cc cc ccc}
\toprule
\multirow{2}{*}{\textbf{Setting}} &
\multirow{2}{*}{\textbf{Architecture}} &
\multicolumn{1}{c}{\textbf{Sparsity}} &
\multicolumn{1}{c}{\textbf{Reconstruction}} &
\multicolumn{2}{c}{\textbf{Ground-truth Recovery}} &
\multicolumn{2}{c}{\textbf{Cross-seed Consistency}} &
\multicolumn{3}{c}{\textbf{Dictionary Utilization}} \\
\cmidrule(lr){3-3}
\cmidrule(lr){4-4}
\cmidrule(lr){5-6}
\cmidrule(lr){7-8}
\cmidrule(lr){9-11}
&
&
$\ell_0$ &
$R^2$ &
DCS &
MCC &
DCS &
MCC &
\# Dup &
\# Dead &
Usable atom frac. \\
\midrule
\multirow{7}{*}{$\beta = 0.0$} & ReLU & 87.0\,{\scriptsize $\pm$0.0} & 0.27\,{\scriptsize $\pm$0.00} & 0.37\,{\scriptsize $\pm$0.00} & 0.16\,{\scriptsize $\pm$0.00} & 0.22\,{\scriptsize $\pm$0.00} & 0.05\,{\scriptsize $\pm$0.00} & 1\,{\scriptsize $\pm$1} & 0\,{\scriptsize $\pm$0} & 1.00\,{\scriptsize $\pm$0.00} \\
 & TopK & 60.0\,{\scriptsize $\pm$0.0} & 0.40\,{\scriptsize $\pm$0.00} & 0.18\,{\scriptsize $\pm$0.00} & 0.02\,{\scriptsize $\pm$0.00} & 0.15\,{\scriptsize $\pm$0.00} & 0.01\,{\scriptsize $\pm$0.00} & 0\,{\scriptsize $\pm$0} & 0\,{\scriptsize $\pm$0} & 1.00\,{\scriptsize $\pm$0.00} \\
 & BatchTopK & 88.0\,{\scriptsize $\pm$0.6} & 0.45\,{\scriptsize $\pm$0.00} & 0.17\,{\scriptsize $\pm$0.00} & 0.02\,{\scriptsize $\pm$0.00} & 0.15\,{\scriptsize $\pm$0.00} & 0.02\,{\scriptsize $\pm$0.00} & 0\,{\scriptsize $\pm$0} & 0\,{\scriptsize $\pm$0} & 1.00\,{\scriptsize $\pm$0.00} \\
 & Matryoshka & 85.0\,{\scriptsize $\pm$0.0} & 0.44\,{\scriptsize $\pm$0.00} & 0.17\,{\scriptsize $\pm$0.00} & 0.02\,{\scriptsize $\pm$0.00} & 0.14\,{\scriptsize $\pm$0.00} & 0.01\,{\scriptsize $\pm$0.00} & 0\,{\scriptsize $\pm$0} & 0\,{\scriptsize $\pm$0} & 1.00\,{\scriptsize $\pm$0.00} \\
 & MP-SAE & 60.0\,{\scriptsize $\pm$0.0} & 0.57\,{\scriptsize $\pm$0.00} & 0.18\,{\scriptsize $\pm$0.00} & 0.02\,{\scriptsize $\pm$0.00} & 0.15\,{\scriptsize $\pm$0.00} & 0.01\,{\scriptsize $\pm$0.00} & 0\,{\scriptsize $\pm$0} & 0\,{\scriptsize $\pm$0} & 1.00\,{\scriptsize $\pm$0.00} \\
 & iSAE & 60.0\,{\scriptsize $\pm$0.0} & 0.40\,{\scriptsize $\pm$0.00} & 0.16\,{\scriptsize $\pm$0.00} & 0.01\,{\scriptsize $\pm$0.00} & 0.15\,{\scriptsize $\pm$0.00} & 0.01\,{\scriptsize $\pm$0.00} & 0\,{\scriptsize $\pm$0} & 0\,{\scriptsize $\pm$0} & 1.00\,{\scriptsize $\pm$0.00} \\
 & Ours & 63.0\,{\scriptsize $\pm$0.0} & \textbf{0.82}\,{\scriptsize $\pm$0.00} & \textbf{0.98}\,{\scriptsize $\pm$0.00} & \textbf{0.94}\,{\scriptsize $\pm$0.00} & \textbf{0.96}\,{\scriptsize $\pm$0.00} & \textbf{0.98}\,{\scriptsize $\pm$0.00} & 0\,{\scriptsize $\pm$0} & 0\,{\scriptsize $\pm$0} & 1.00\,{\scriptsize $\pm$0.00} \\
\cmidrule(lr){1-11}
\multirow{7}{*}{$\beta = 0.5$} & ReLU & 102\,{\scriptsize $\pm$0} & 0.36\,{\scriptsize $\pm$0.00} & 0.44\,{\scriptsize $\pm$0.00} & 0.24\,{\scriptsize $\pm$0.00} & 0.39\,{\scriptsize $\pm$0.00} & 0.25\,{\scriptsize $\pm$0.00} & 29\,{\scriptsize $\pm$3} & 0\,{\scriptsize $\pm$0} & 1.00\,{\scriptsize $\pm$0.00} \\
 & TopK & 60.0\,{\scriptsize $\pm$0.0} & 0.42\,{\scriptsize $\pm$0.00} & 0.30\,{\scriptsize $\pm$0.00} & 0.10\,{\scriptsize $\pm$0.00} & 0.25\,{\scriptsize $\pm$0.00} & 0.09\,{\scriptsize $\pm$0.00} & 0\,{\scriptsize $\pm$0} & 0\,{\scriptsize $\pm$0} & 1.00\,{\scriptsize $\pm$0.00} \\
 & BatchTopK & 86.0\,{\scriptsize $\pm$0.0} & 0.45\,{\scriptsize $\pm$0.00} & 0.29\,{\scriptsize $\pm$0.00} & 0.09\,{\scriptsize $\pm$0.00} & 0.24\,{\scriptsize $\pm$0.00} & 0.09\,{\scriptsize $\pm$0.00} & 0\,{\scriptsize $\pm$0} & 0\,{\scriptsize $\pm$0} & 1.00\,{\scriptsize $\pm$0.00} \\
 & Matryoshka & 82.0\,{\scriptsize $\pm$0.0} & 0.44\,{\scriptsize $\pm$0.00} & 0.27\,{\scriptsize $\pm$0.00} & 0.08\,{\scriptsize $\pm$0.00} & 0.23\,{\scriptsize $\pm$0.00} & 0.08\,{\scriptsize $\pm$0.00} & 0\,{\scriptsize $\pm$0} & 0\,{\scriptsize $\pm$0} & 1.00\,{\scriptsize $\pm$0.00} \\
 & MP-SAE & 60.0\,{\scriptsize $\pm$0.0} & 0.59\,{\scriptsize $\pm$0.00} & 0.32\,{\scriptsize $\pm$0.00} & 0.11\,{\scriptsize $\pm$0.00} & 0.28\,{\scriptsize $\pm$0.00} & 0.08\,{\scriptsize $\pm$0.00} & 352\,{\scriptsize $\pm$8} & 0\,{\scriptsize $\pm$0} & 0.98\,{\scriptsize $\pm$0.00} \\
 & iSAE & 60.0\,{\scriptsize $\pm$0.0} & 0.40\,{\scriptsize $\pm$0.00} & 0.19\,{\scriptsize $\pm$0.00} & 0.02\,{\scriptsize $\pm$0.00} & 0.18\,{\scriptsize $\pm$0.00} & 0.02\,{\scriptsize $\pm$0.00} & 0\,{\scriptsize $\pm$0} & 0\,{\scriptsize $\pm$0} & 1.00\,{\scriptsize $\pm$0.00} \\
 & Ours & 67.0\,{\scriptsize $\pm$1.2} & \textbf{0.90}\,{\scriptsize $\pm$0.01} & \textbf{0.98}\,{\scriptsize $\pm$0.00} & \textbf{0.97}\,{\scriptsize $\pm$0.00} & \textbf{0.97}\,{\scriptsize $\pm$0.00} & \textbf{0.98}\,{\scriptsize $\pm$0.00} & 0\,{\scriptsize $\pm$0} & 0\,{\scriptsize $\pm$0} & 1.00\,{\scriptsize $\pm$0.00} \\
\cmidrule(lr){1-11}
\multirow{7}{*}{$\beta = 0.7$} & ReLU & 100\,{\scriptsize $\pm$9.7} & 0.22\,{\scriptsize $\pm$0.01} & 0.27\,{\scriptsize $\pm$0.00} & 0.09\,{\scriptsize $\pm$0.00} & 0.30\,{\scriptsize $\pm$0.00} & 0.11\,{\scriptsize $\pm$0.00} & 28\,{\scriptsize $\pm$7} & 8\,{\scriptsize $\pm$0} & 1.00\,{\scriptsize $\pm$0.00} \\
 & TopK & 60.0\,{\scriptsize $\pm$0.0} & 0.38\,{\scriptsize $\pm$0.02} & 0.35\,{\scriptsize $\pm$0.00} & 0.14\,{\scriptsize $\pm$0.00} & 0.32\,{\scriptsize $\pm$0.00} & 0.16\,{\scriptsize $\pm$0.00} & 0\,{\scriptsize $\pm$0} & 0\,{\scriptsize $\pm$0} & 1.00\,{\scriptsize $\pm$0.00} \\
 & BatchTopK & 87.0\,{\scriptsize $\pm$1.0} & 0.34\,{\scriptsize $\pm$0.00} & 0.35\,{\scriptsize $\pm$0.00} & 0.13\,{\scriptsize $\pm$0.00} & 0.31\,{\scriptsize $\pm$0.00} & 0.16\,{\scriptsize $\pm$0.00} & 0\,{\scriptsize $\pm$0} & 0\,{\scriptsize $\pm$0} & 1.00\,{\scriptsize $\pm$0.00} \\
 & Matryoshka & 85.0\,{\scriptsize $\pm$0.0} & 0.37\,{\scriptsize $\pm$0.00} & 0.30\,{\scriptsize $\pm$0.00} & 0.10\,{\scriptsize $\pm$0.00} & 0.28\,{\scriptsize $\pm$0.00} & 0.13\,{\scriptsize $\pm$0.00} & 1\,{\scriptsize $\pm$0} & 0\,{\scriptsize $\pm$0} & 1.00\,{\scriptsize $\pm$0.00} \\
 & MP-SAE & 60.0\,{\scriptsize $\pm$0.0} & 0.38\,{\scriptsize $\pm$0.00} & 0.18\,{\scriptsize $\pm$0.00} & 0.02\,{\scriptsize $\pm$0.00} & 0.52\,{\scriptsize $\pm$0.00} & 0.02\,{\scriptsize $\pm$0.00} & 261\,{\scriptsize $\pm$4} & 0\,{\scriptsize $\pm$0} & 0.99\,{\scriptsize $\pm$0.00} \\
 & iSAE & 60.0\,{\scriptsize $\pm$0.0} & 0.36\,{\scriptsize $\pm$0.00} & 0.24\,{\scriptsize $\pm$0.00} & 0.02\,{\scriptsize $\pm$0.00} & 0.24\,{\scriptsize $\pm$0.00} & 0.03\,{\scriptsize $\pm$0.00} & 0\,{\scriptsize $\pm$0} & 0\,{\scriptsize $\pm$0} & 1.00\,{\scriptsize $\pm$0.00} \\
 & Ours & 90.0\,{\scriptsize $\pm$1.0} & \textbf{0.94}\,{\scriptsize $\pm$0.00} & \textbf{0.97}\,{\scriptsize $\pm$0.00} & \textbf{0.97}\,{\scriptsize $\pm$0.00} & \textbf{0.96}\,{\scriptsize $\pm$0.00} & \textbf{0.96}\,{\scriptsize $\pm$0.00} & 0\,{\scriptsize $\pm$0} & 8\,{\scriptsize $\pm$2} & 1.00\,{\scriptsize $\pm$0.00} \\
\cmidrule(lr){1-11}
\multirow{7}{*}{$\beta = 1.0$} & ReLU & 109\,{\scriptsize $\pm$25} & -0.29\,{\scriptsize $\pm$0.03} & 0.17\,{\scriptsize $\pm$0.00} & 0.02\,{\scriptsize $\pm$0.00} & 0.18\,{\scriptsize $\pm$0.00} & 0.03\,{\scriptsize $\pm$0.00} & 31\,{\scriptsize $\pm$3} & 198\,{\scriptsize $\pm$93} & 0.98\,{\scriptsize $\pm$0.01} \\
 & TopK & 60.0\,{\scriptsize $\pm$0.0} & 0.27\,{\scriptsize $\pm$0.08} & 0.36\,{\scriptsize $\pm$0.00} & 0.14\,{\scriptsize $\pm$0.00} & 0.36\,{\scriptsize $\pm$0.00} & 0.17\,{\scriptsize $\pm$0.00} & 1\,{\scriptsize $\pm$0} & 1723\,{\scriptsize $\pm$91} & 0.86\,{\scriptsize $\pm$0.01} \\
 & BatchTopK & 80.0\,{\scriptsize $\pm$1.2} & 0.11\,{\scriptsize $\pm$0.01} & 0.32\,{\scriptsize $\pm$0.00} & 0.09\,{\scriptsize $\pm$0.00} & 0.31\,{\scriptsize $\pm$0.00} & 0.15\,{\scriptsize $\pm$0.00} & 0\,{\scriptsize $\pm$0} & 1002\,{\scriptsize $\pm$46} & 0.92\,{\scriptsize $\pm$0.00} \\
 & Matryoshka & 78.0\,{\scriptsize $\pm$2.5} & 0.09\,{\scriptsize $\pm$0.00} & 0.27\,{\scriptsize $\pm$0.00} & 0.07\,{\scriptsize $\pm$0.00} & 0.26\,{\scriptsize $\pm$0.00} & 0.11\,{\scriptsize $\pm$0.00} & 1\,{\scriptsize $\pm$0} & 541\,{\scriptsize $\pm$38} & 0.96\,{\scriptsize $\pm$0.00} \\
 & MP-SAE & 59.9\,{\scriptsize $\pm$0.0} & -0.01\,{\scriptsize $\pm$0.01} & 0.19\,{\scriptsize $\pm$0.00} & 0.03\,{\scriptsize $\pm$0.00} & 0.60\,{\scriptsize $\pm$0.00} & 0.05\,{\scriptsize $\pm$0.00} & 2217\,{\scriptsize $\pm$33} & \textbf{4}\,{\scriptsize $\pm$1} & 0.90\,{\scriptsize $\pm$0.00} \\
 & iSAE & 60.0\,{\scriptsize $\pm$0.0} & 0.17\,{\scriptsize $\pm$0.01} & 0.26\,{\scriptsize $\pm$0.00} & 0.02\,{\scriptsize $\pm$0.00} & 0.25\,{\scriptsize $\pm$0.00} & 0.05\,{\scriptsize $\pm$0.00} & 0\,{\scriptsize $\pm$0} & 10167\,{\scriptsize $\pm$17} & 0.17\,{\scriptsize $\pm$0.00} \\
 & Ours & 117\,{\scriptsize $\pm$5} & \textbf{0.93}\,{\scriptsize $\pm$0.00} & \textbf{0.92}\,{\scriptsize $\pm$0.00} & \textbf{0.90}\,{\scriptsize $\pm$0.01} & \textbf{0.91}\,{\scriptsize $\pm$0.00} & \textbf{0.86}\,{\scriptsize $\pm$0.00} & 0\,{\scriptsize $\pm$0} & 154\,{\scriptsize $\pm$17} & \textbf{0.99}\,{\scriptsize $\pm$0.00} \\
\bottomrule
\end{tabular}%
}
\end{table*}

\begin{table*}[t]
\centering
\caption{
\textbf{Full results on CLIP ViT-B/16 across dictionary widths and sparsity levels.}
We evaluate three expansion factors ($4\times$, $8\times$, $16\times$) and three target sparsity levels ($\ell_0\approx32$, $64$, $128$).
We report achieved $\ell_0$, reconstruction ($R^2$), cross-seed consistency (\texttt{DCS}, \texttt{MCC}), and dictionary utilization, including usable atom fraction and numbers of dead and duplicated atoms.
Metrics are averaged across training seeds, while \texttt{DCS} and \texttt{MCC} are averaged across seed pairs.
$\pm$ denotes the corresponding standard deviation.
}
\label{tab:appendix_clip_full}

\setlength{\tabcolsep}{3.5pt}
\renewcommand{\arraystretch}{0.90}

\resizebox{\textwidth}{!}{%
\begin{tabular}{ll l c c cc ccc}
\toprule
\multirow{2}{*}{\textbf{Expansion}} &
\multirow{2}{*}{\textbf{$\ell_0$ band}} &
\multirow{2}{*}{\textbf{Method}} &
\multicolumn{1}{c}{\textbf{Sparsity}} &
\multicolumn{1}{c}{\textbf{Reconstruction}} &
\multicolumn{2}{c}{\textbf{Consistency}} &
\multicolumn{3}{c}{\textbf{Dictionary Utilization}} \\
\cmidrule(lr){4-4}
\cmidrule(lr){5-5}
\cmidrule(lr){6-7}
\cmidrule(lr){8-10}
&
&
&
$\ell_0$ &
$R^2$ &
DCS &
MCC &
Usable atom frac. &
\# Dup. &
\# Dead \\
\midrule
\multirow{21}{*}{$4\times$}
& \multirow{7}{*}{$\ell_0 \approx 32$}
& ReLU       & 34\,{\scriptsize $\pm$0.4} & 0.89\,{\scriptsize $\pm$0.00} & 0.74\,{\scriptsize $\pm$0.00} & 0.68\,{\scriptsize $\pm$0.00} & 0.99\,{\scriptsize $\pm$0.00} & 36\,{\scriptsize $\pm$4.1} & 0\,{\scriptsize $\pm$0.00} \\
& & TopK       & 32\,{\scriptsize $\pm$0.0} & \textbf{0.93}\,{\scriptsize $\pm$0.00} & 0.80\,{\scriptsize $\pm$0.00} & 0.77\,{\scriptsize $\pm$0.00} & 1.00\,{\scriptsize $\pm$0.00} & 1\,{\scriptsize $\pm$0.58} & 0\,{\scriptsize $\pm$0.00} \\
& & BatchTopK  & 32\,{\scriptsize $\pm$0.0} & 0.93\,{\scriptsize $\pm$0.00} & 0.72\,{\scriptsize $\pm$0.00} & 0.70\,{\scriptsize $\pm$0.00} & 1.00\,{\scriptsize $\pm$0.00} & 1\,{\scriptsize $\pm$0.29} & 0\,{\scriptsize $\pm$0.00} \\
& & Matryoshka & 32\,{\scriptsize $\pm$0.0} & 0.92\,{\scriptsize $\pm$0.00} & 0.51\,{\scriptsize $\pm$0.00} & 0.47\,{\scriptsize $\pm$0.01} & 1.00\,{\scriptsize $\pm$0.00} & 6\,{\scriptsize $\pm$0.87} & 0\,{\scriptsize $\pm$0.00} \\
& & MP-SAE     & 32\,{\scriptsize $\pm$0.0} & 0.93\,{\scriptsize $\pm$0.00} & 0.79\,{\scriptsize $\pm$0.01} & 0.63\,{\scriptsize $\pm$0.00} & 0.91\,{\scriptsize $\pm$0.00} & 275\,{\scriptsize $\pm$9} & 0\,{\scriptsize $\pm$0.00} \\
& & iSAE       & 32\,{\scriptsize $\pm$0.0} & 0.92\,{\scriptsize $\pm$0.00} & 0.82\,{\scriptsize $\pm$0.00} & 0.46\,{\scriptsize $\pm$0.01} & 1.00\,{\scriptsize $\pm$0.00} & 0\,{\scriptsize $\pm$0.00} & 0\,{\scriptsize $\pm$0.00} \\
& & ConSDL     & 35\,{\scriptsize $\pm$0.5} & 0.91\,{\scriptsize $\pm$0.00} & \textbf{0.85}\,{\scriptsize $\pm$0.00} & \textbf{0.81}\,{\scriptsize $\pm$0.00} & 1.00\,{\scriptsize $\pm$0.00} & 0\,{\scriptsize $\pm$0.00} & 0\,{\scriptsize $\pm$0.00} \\
\cmidrule(lr){2-10}
& \multirow{7}{*}{$\ell_0 \approx 64$}
& ReLU       & 65\,{\scriptsize $\pm$0.8} & 0.91\,{\scriptsize $\pm$0.00} & 0.80\,{\scriptsize $\pm$0.00} & 0.75\,{\scriptsize $\pm$0.00} & 0.99\,{\scriptsize $\pm$0.00} & 23\,{\scriptsize $\pm$2.6} & 0\,{\scriptsize $\pm$0.00} \\
& & TopK       & 64\,{\scriptsize $\pm$0.0} & 0.94\,{\scriptsize $\pm$0.00} & 0.78\,{\scriptsize $\pm$0.00} & 0.78\,{\scriptsize $\pm$0.00} & 1.00\,{\scriptsize $\pm$0.00} & 0\,{\scriptsize $\pm$0.29} & 0\,{\scriptsize $\pm$0.00} \\
& & BatchTopK  & 64\,{\scriptsize $\pm$0.0} & 0.94\,{\scriptsize $\pm$0.00} & 0.68\,{\scriptsize $\pm$0.01} & 0.66\,{\scriptsize $\pm$0.01} & 1.00\,{\scriptsize $\pm$0.00} & 0\,{\scriptsize $\pm$0.29} & 0\,{\scriptsize $\pm$0.00} \\
& & Matryoshka & 64\,{\scriptsize $\pm$0.2} & 0.93\,{\scriptsize $\pm$0.00} & 0.50\,{\scriptsize $\pm$0.00} & 0.48\,{\scriptsize $\pm$0.00} & 1.00\,{\scriptsize $\pm$0.00} & 1\,{\scriptsize $\pm$0.29} & 0\,{\scriptsize $\pm$0.00} \\
& & MP-SAE     & 64\,{\scriptsize $\pm$0.0} & \textbf{0.94}\,{\scriptsize $\pm$0.00} & 0.79\,{\scriptsize $\pm$0.01} & 0.63\,{\scriptsize $\pm$0.00} & 0.91\,{\scriptsize $\pm$0.00} & 275\,{\scriptsize $\pm$9} & 0\,{\scriptsize $\pm$0.00} \\
& & iSAE       & 64\,{\scriptsize $\pm$0.0} & 0.94\,{\scriptsize $\pm$0.00} & 0.81\,{\scriptsize $\pm$0.00} & 0.46\,{\scriptsize $\pm$0.01} & 1.00\,{\scriptsize $\pm$0.00} & 0\,{\scriptsize $\pm$0.00} & 0\,{\scriptsize $\pm$0.00} \\
& & ConSDL     & 57\,{\scriptsize $\pm$0.8} & 0.92\,{\scriptsize $\pm$0.00} & \textbf{0.86}\,{\scriptsize $\pm$0.00} & \textbf{0.82}\,{\scriptsize $\pm$0.00} & 1.00\,{\scriptsize $\pm$0.00} & 0\,{\scriptsize $\pm$0.00} & 0\,{\scriptsize $\pm$0.00} \\
\cmidrule(lr){2-10}
& \multirow{7}{*}{$\ell_0 \approx 128$}
& ReLU       & 113\,{\scriptsize $\pm$1} & 0.92\,{\scriptsize $\pm$0.00} & 0.77\,{\scriptsize $\pm$0.00} & 0.73\,{\scriptsize $\pm$0.00} & 1.00\,{\scriptsize $\pm$0.00} & 12\,{\scriptsize $\pm$2.1} & 0\,{\scriptsize $\pm$0.00} \\
& & TopK       & 128\,{\scriptsize $\pm$0} & 0.96\,{\scriptsize $\pm$0.00} & 0.65\,{\scriptsize $\pm$0.00} & 0.67\,{\scriptsize $\pm$0.00} & 1.00\,{\scriptsize $\pm$0.00} & 1\,{\scriptsize $\pm$0.76} & 0\,{\scriptsize $\pm$0.00} \\
& & BatchTopK  & 129\,{\scriptsize $\pm$0} & 0.96\,{\scriptsize $\pm$0.00} & 0.51\,{\scriptsize $\pm$0.00} & 0.49\,{\scriptsize $\pm$0.00} & 1.00\,{\scriptsize $\pm$0.00} & 0\,{\scriptsize $\pm$0.00} & 0\,{\scriptsize $\pm$0.00} \\
& & Matryoshka & 129\,{\scriptsize $\pm$0} & 0.95\,{\scriptsize $\pm$0.00} & 0.37\,{\scriptsize $\pm$0.00} & 0.35\,{\scriptsize $\pm$0.00} & 1.00\,{\scriptsize $\pm$0.00} & 0\,{\scriptsize $\pm$0.29} & 0\,{\scriptsize $\pm$0.00} \\
& & MP-SAE     & 128\,{\scriptsize $\pm$0} & \textbf{0.96}\,{\scriptsize $\pm$0.00} & 0.79\,{\scriptsize $\pm$0.01} & 0.63\,{\scriptsize $\pm$0.00} & 0.91\,{\scriptsize $\pm$0.00} & 275\,{\scriptsize $\pm$9} & 0\,{\scriptsize $\pm$0.00} \\
& & iSAE       & 128\,{\scriptsize $\pm$0} & 0.95\,{\scriptsize $\pm$0.00} & 0.75\,{\scriptsize $\pm$0.00} & 0.42\,{\scriptsize $\pm$0.01} & 1.00\,{\scriptsize $\pm$0.00} & 0\,{\scriptsize $\pm$0.00} & 0\,{\scriptsize $\pm$0.00} \\
& & ConSDL     & 121\,{\scriptsize $\pm$0} & 0.96\,{\scriptsize $\pm$0.00} & \textbf{0.83}\,{\scriptsize $\pm$0.00} & \textbf{0.78}\,{\scriptsize $\pm$0.00} & 1.00\,{\scriptsize $\pm$0.00} & 0\,{\scriptsize $\pm$0.00} & 0\,{\scriptsize $\pm$0.00} \\
\midrule
\multirow{21}{*}{$8\times$}
& \multirow{7}{*}{$\ell_0 \approx 32$}
& ReLU       & 34\,{\scriptsize $\pm$1.8} & 0.89\,{\scriptsize $\pm$0.00} & 0.52\,{\scriptsize $\pm$0.01} & 0.44\,{\scriptsize $\pm$0.01} & 0.99\,{\scriptsize $\pm$0.00} & 73\,{\scriptsize $\pm$4.5} & 0\,{\scriptsize $\pm$0.00} \\
& & TopK       & 32\,{\scriptsize $\pm$0.0} & 0.94\,{\scriptsize $\pm$0.00} & 0.73\,{\scriptsize $\pm$0.00} & 0.67\,{\scriptsize $\pm$0.00} & 1.00\,{\scriptsize $\pm$0.00} & 4\,{\scriptsize $\pm$2.00} & 0\,{\scriptsize $\pm$0.00} \\
& & BatchTopK  & 32\,{\scriptsize $\pm$0.4} & 0.94\,{\scriptsize $\pm$0.00} & 0.61\,{\scriptsize $\pm$0.00} & 0.57\,{\scriptsize $\pm$0.00} & 1.00\,{\scriptsize $\pm$0.00} & 1\,{\scriptsize $\pm$0.76} & 0\,{\scriptsize $\pm$0.00} \\
& & Matryoshka & 32\,{\scriptsize $\pm$0.0} & 0.92\,{\scriptsize $\pm$0.00} & 0.39\,{\scriptsize $\pm$0.00} & 0.32\,{\scriptsize $\pm$0.00} & 1.00\,{\scriptsize $\pm$0.00} & 27\,{\scriptsize $\pm$4.1} & 0\,{\scriptsize $\pm$0.00} \\
& & MP-SAE     & 32\,{\scriptsize $\pm$0.0} & 0.93\,{\scriptsize $\pm$0.00} & 0.75\,{\scriptsize $\pm$0.01} & 0.44\,{\scriptsize $\pm$0.00} & 0.84\,{\scriptsize $\pm$0.01} & 983\,{\scriptsize $\pm$80} & 0\,{\scriptsize $\pm$0.00} \\
& & iSAE       & 32\,{\scriptsize $\pm$0.0} & 0.93\,{\scriptsize $\pm$0.00} & 0.75\,{\scriptsize $\pm$0.00} & 0.40\,{\scriptsize $\pm$0.00} & 1.00\,{\scriptsize $\pm$0.00} & 0\,{\scriptsize $\pm$0.00} & 0\,{\scriptsize $\pm$0.00} \\
& & ConSDL     & 35\,{\scriptsize $\pm$0.4} & \textbf{0.94}\,{\scriptsize $\pm$0.00} & \textbf{0.76}\,{\scriptsize $\pm$0.00} & \textbf{0.69}\,{\scriptsize $\pm$0.00} & 1.00\,{\scriptsize $\pm$0.00} & 0\,{\scriptsize $\pm$0.00} & 0\,{\scriptsize $\pm$0.00} \\
\cmidrule(lr){2-10}
& \multirow{7}{*}{$\ell_0 \approx 64$}
& ReLU       & 65\,{\scriptsize $\pm$3.2} & 0.91\,{\scriptsize $\pm$0.00} & 0.66\,{\scriptsize $\pm$0.00} & 0.58\,{\scriptsize $\pm$0.00} & 0.99\,{\scriptsize $\pm$0.00} & 76\,{\scriptsize $\pm$7.3} & 0\,{\scriptsize $\pm$0.00} \\
& & TopK       & 64\,{\scriptsize $\pm$0.0} & 0.95\,{\scriptsize $\pm$0.00} & 0.74\,{\scriptsize $\pm$0.00} & 0.71\,{\scriptsize $\pm$0.00} & 1.00\,{\scriptsize $\pm$0.00} & 4\,{\scriptsize $\pm$1.04} & 0\,{\scriptsize $\pm$0.00} \\
& & BatchTopK  & 64\,{\scriptsize $\pm$0.4} & 0.95\,{\scriptsize $\pm$0.00} & 0.62\,{\scriptsize $\pm$0.00} & 0.59\,{\scriptsize $\pm$0.00} & 1.00\,{\scriptsize $\pm$0.00} & 2\,{\scriptsize $\pm$1.04} & 0\,{\scriptsize $\pm$0.00} \\
& & Matryoshka & 65\,{\scriptsize $\pm$0.0} & 0.94\,{\scriptsize $\pm$0.00} & 0.40\,{\scriptsize $\pm$0.00} & 0.35\,{\scriptsize $\pm$0.00} & 1.00\,{\scriptsize $\pm$0.00} & 9\,{\scriptsize $\pm$2.00} & 0\,{\scriptsize $\pm$0.00} \\
& & MP-SAE     & 64\,{\scriptsize $\pm$0.0} & \textbf{0.95}\,{\scriptsize $\pm$0.00} & 0.75\,{\scriptsize $\pm$0.01} & 0.44\,{\scriptsize $\pm$0.00} & 0.84\,{\scriptsize $\pm$0.01} & 983\,{\scriptsize $\pm$80} & 0\,{\scriptsize $\pm$0.00} \\
& & iSAE       & 64\,{\scriptsize $\pm$0.0} & 0.94\,{\scriptsize $\pm$0.00} & 0.80\,{\scriptsize $\pm$0.00} & 0.44\,{\scriptsize $\pm$0.01} & 1.00\,{\scriptsize $\pm$0.00} & \textbf{0}\,{\scriptsize $\pm$0.00} & 0\,{\scriptsize $\pm$0.00} \\
& & ConSDL     & 64\,{\scriptsize $\pm$0.7} & 0.91\,{\scriptsize $\pm$0.00} & \textbf{0.82}\,{\scriptsize $\pm$0.00} & \textbf{0.79}\,{\scriptsize $\pm$0.00} & 1.00\,{\scriptsize $\pm$0.00} & 2\,{\scriptsize $\pm$0.58} & 0\,{\scriptsize $\pm$0.00} \\
\cmidrule(lr){2-10}
& \multirow{7}{*}{$\ell_0 \approx 128$}
& ReLU       & 121\,{\scriptsize $\pm$0} & 0.93\,{\scriptsize $\pm$0.00} & 0.67\,{\scriptsize $\pm$0.00} & 0.60\,{\scriptsize $\pm$0.00} & 0.99\,{\scriptsize $\pm$0.00} & 47\,{\scriptsize $\pm$11.0} & 0\,{\scriptsize $\pm$0.00} \\
& & TopK       & 128\,{\scriptsize $\pm$0} & 0.96\,{\scriptsize $\pm$0.00} & 0.68\,{\scriptsize $\pm$0.00} & 0.66\,{\scriptsize $\pm$0.00} & 1.00\,{\scriptsize $\pm$0.00} & 3\,{\scriptsize $\pm$1.76} & 0\,{\scriptsize $\pm$0.00} \\
& & BatchTopK  & 129\,{\scriptsize $\pm$0} & 0.96\,{\scriptsize $\pm$0.00} & 0.53\,{\scriptsize $\pm$0.00} & 0.50\,{\scriptsize $\pm$0.00} & 1.00\,{\scriptsize $\pm$0.00} & 0\,{\scriptsize $\pm$0.00} & 0\,{\scriptsize $\pm$0.00} \\
& & Matryoshka & 129\,{\scriptsize $\pm$0} & 0.95\,{\scriptsize $\pm$0.00} & 0.30\,{\scriptsize $\pm$0.00} & 0.25\,{\scriptsize $\pm$0.00} & 1.00\,{\scriptsize $\pm$0.00} & 3\,{\scriptsize $\pm$2.31} & 0\,{\scriptsize $\pm$0.00} \\
& & MP-SAE     & 128\,{\scriptsize $\pm$0} & \textbf{0.97}\,{\scriptsize $\pm$0.00} & 0.75\,{\scriptsize $\pm$0.01} & 0.44\,{\scriptsize $\pm$0.00} & 0.84\,{\scriptsize $\pm$0.01} & 983\,{\scriptsize $\pm$80} & 0\,{\scriptsize $\pm$0.00} \\
& & iSAE       & 128\,{\scriptsize $\pm$0} & 0.96\,{\scriptsize $\pm$0.00} & 0.78\,{\scriptsize $\pm$0.00} & 0.43\,{\scriptsize $\pm$0.00} & 1.00\,{\scriptsize $\pm$0.00} & 0\,{\scriptsize $\pm$0.00} & 0\,{\scriptsize $\pm$0.00} \\
& & ConSDL     & 134\,{\scriptsize $\pm$1} & 0.94\,{\scriptsize $\pm$0.00} & \textbf{0.82}\,{\scriptsize $\pm$0.00} & \textbf{0.79}\,{\scriptsize $\pm$0.00} & 1.00\,{\scriptsize $\pm$0.00} & 0\,{\scriptsize $\pm$0.00} & 0\,{\scriptsize $\pm$0.00} \\
\midrule
\multirow{21}{*}{$16\times$}
& \multirow{7}{*}{$\ell_0 \approx 32$}
& ReLU       & 29\,{\scriptsize $\pm$2.1} & 0.89\,{\scriptsize $\pm$0.00} & 0.33\,{\scriptsize $\pm$0.00} & 0.23\,{\scriptsize $\pm$0.01} & 0.99\,{\scriptsize $\pm$0.00} & 109\,{\scriptsize $\pm$8} & 0\,{\scriptsize $\pm$0.00} \\
& & TopK       & 32\,{\scriptsize $\pm$0.0} & \textbf{0.94}\,{\scriptsize $\pm$0.00} & 0.48\,{\scriptsize $\pm$0.00} & 0.38\,{\scriptsize $\pm$0.00} & 1.00\,{\scriptsize $\pm$0.00} & 4\,{\scriptsize $\pm$2.50} & 0\,{\scriptsize $\pm$0.00} \\
& & BatchTopK  & 32\,{\scriptsize $\pm$0.0} & 0.93\,{\scriptsize $\pm$0.00} & 0.42\,{\scriptsize $\pm$0.00} & 0.34\,{\scriptsize $\pm$0.00} & 1.00\,{\scriptsize $\pm$0.00} & 1\,{\scriptsize $\pm$0.29} & 0\,{\scriptsize $\pm$0.00} \\
& & Matryoshka & 32\,{\scriptsize $\pm$0.0} & 0.93\,{\scriptsize $\pm$0.00} & 0.29\,{\scriptsize $\pm$0.00} & 0.20\,{\scriptsize $\pm$0.00} & 1.00\,{\scriptsize $\pm$0.00} & 9\,{\scriptsize $\pm$4.77} & 0\,{\scriptsize $\pm$0.00} \\
& & MP-SAE     & 32\,{\scriptsize $\pm$0.0} & 0.94\,{\scriptsize $\pm$0.00} & \textbf{0.69}\,{\scriptsize $\pm$0.00} & 0.27\,{\scriptsize $\pm$0.00} & 0.79\,{\scriptsize $\pm$0.00} & 2623\,{\scriptsize $\pm$67} & 0\,{\scriptsize $\pm$0.00} \\
& & iSAE       & 32\,{\scriptsize $\pm$0.0} & 0.93\,{\scriptsize $\pm$0.00} & 0.65\,{\scriptsize $\pm$0.00} & 0.27\,{\scriptsize $\pm$0.00} & 0.67\,{\scriptsize $\pm$0.00} & 1589\,{\scriptsize $\pm$6} & 2517\,{\scriptsize $\pm$51} \\
& & ConSDL     & 35\,{\scriptsize $\pm$0.7} & 0.92\,{\scriptsize $\pm$0.00} & 0.64\,{\scriptsize $\pm$0.00} & \textbf{0.56}\,{\scriptsize $\pm$0.00} & 1.00\,{\scriptsize $\pm$0.00} & \textbf{0}\,{\scriptsize $\pm$0.00} & 0\,{\scriptsize $\pm$0.00} \\
\cmidrule(lr){2-10}
& \multirow{7}{*}{$\ell_0 \approx 64$}
& ReLU       & 64\,{\scriptsize $\pm$1.4} & 0.91\,{\scriptsize $\pm$0.00} & 0.47\,{\scriptsize $\pm$0.00} & 0.36\,{\scriptsize $\pm$0.00} & 0.99\,{\scriptsize $\pm$0.00} & 181\,{\scriptsize $\pm$7} & 0\,{\scriptsize $\pm$0.00} \\
& & TopK       & 64\,{\scriptsize $\pm$0.0} & \textbf{0.95}\,{\scriptsize $\pm$0.00} & 0.60\,{\scriptsize $\pm$0.00} & 0.52\,{\scriptsize $\pm$0.00} & 1.00\,{\scriptsize $\pm$0.00} & 16\,{\scriptsize $\pm$0.3} & 0\,{\scriptsize $\pm$0.00} \\
& & BatchTopK  & 64\,{\scriptsize $\pm$0.0} & 0.95\,{\scriptsize $\pm$0.00} & 0.48\,{\scriptsize $\pm$0.00} & 0.42\,{\scriptsize $\pm$0.00} & 1.00\,{\scriptsize $\pm$0.00} & 3\,{\scriptsize $\pm$1.26} & 0\,{\scriptsize $\pm$0.00} \\
& & Matryoshka & 64\,{\scriptsize $\pm$0.0} & 0.94\,{\scriptsize $\pm$0.00} & 0.30\,{\scriptsize $\pm$0.00} & 0.22\,{\scriptsize $\pm$0.00} & 1.00\,{\scriptsize $\pm$0.00} & 13\,{\scriptsize $\pm$2.8} & 0\,{\scriptsize $\pm$0.00} \\
& & MP-SAE     & 64\,{\scriptsize $\pm$0.0} & 0.95\,{\scriptsize $\pm$0.00} & \textbf{0.69}\,{\scriptsize $\pm$0.00} & 0.27\,{\scriptsize $\pm$0.00} & 0.79\,{\scriptsize $\pm$0.00} & 2623\,{\scriptsize $\pm$67} & 0\,{\scriptsize $\pm$0.00} \\
& & iSAE       & 64\,{\scriptsize $\pm$0.0} & 0.95\,{\scriptsize $\pm$0.00} & 0.68\,{\scriptsize $\pm$0.00} & 0.36\,{\scriptsize $\pm$0.00} & 1.00\,{\scriptsize $\pm$0.00} & 11\,{\scriptsize $\pm$3.3} & 0\,{\scriptsize $\pm$0.58} \\
& & ConSDL     & 63\,{\scriptsize $\pm$0.7} & 0.90\,{\scriptsize $\pm$0.01} & 0.67\,{\scriptsize $\pm$0.00} & \textbf{0.63}\,{\scriptsize $\pm$0.00} & 1.00\,{\scriptsize $\pm$0.00} & \textbf{2}\,{\scriptsize $\pm$0.72} & 0\,{\scriptsize $\pm$0.00} \\
\cmidrule(lr){2-10}
& \multirow{7}{*}{$\ell_0 \approx 128$}
& ReLU       & 116\,{\scriptsize $\pm$1} & 0.93\,{\scriptsize $\pm$0.00} & 0.52\,{\scriptsize $\pm$0.00} & 0.42\,{\scriptsize $\pm$0.00} & 0.99\,{\scriptsize $\pm$0.00} & 168\,{\scriptsize $\pm$9} & 0\,{\scriptsize $\pm$0.00} \\
& & TopK       & 128\,{\scriptsize $\pm$0} & 0.96\,{\scriptsize $\pm$0.00} & 0.56\,{\scriptsize $\pm$0.00} & 0.50\,{\scriptsize $\pm$0.00} & 1.00\,{\scriptsize $\pm$0.00} & 12\,{\scriptsize $\pm$0.6} & 0\,{\scriptsize $\pm$0.00} \\
& & BatchTopK  & 129\,{\scriptsize $\pm$0} & 0.96\,{\scriptsize $\pm$0.00} & 0.48\,{\scriptsize $\pm$0.00} & 0.44\,{\scriptsize $\pm$0.00} & 1.00\,{\scriptsize $\pm$0.00} & 4\,{\scriptsize $\pm$1.00} & 0\,{\scriptsize $\pm$0.00} \\
& & Matryoshka & 129\,{\scriptsize $\pm$0} & 0.95\,{\scriptsize $\pm$0.00} & 0.27\,{\scriptsize $\pm$0.00} & 0.19\,{\scriptsize $\pm$0.00} & 1.00\,{\scriptsize $\pm$0.00} & 6\,{\scriptsize $\pm$3.01} & 0\,{\scriptsize $\pm$0.00} \\
& & MP-SAE     & 128\,{\scriptsize $\pm$0} & \textbf{0.97}\,{\scriptsize $\pm$0.00} & 0.69\,{\scriptsize $\pm$0.00} & 0.27\,{\scriptsize $\pm$0.00} & 0.79\,{\scriptsize $\pm$0.00} & 2623\,{\scriptsize $\pm$67} & 0\,{\scriptsize $\pm$0.00} \\
& & iSAE       & 128\,{\scriptsize $\pm$0} & 0.96\,{\scriptsize $\pm$0.00} & 0.68\,{\scriptsize $\pm$0.00} & 0.37\,{\scriptsize $\pm$0.00} & 1.00\,{\scriptsize $\pm$0.00} & 6\,{\scriptsize $\pm$0.76} & 0\,{\scriptsize $\pm$0.00} \\
& & ConSDL     & 130\,{\scriptsize $\pm$2} & 0.92\,{\scriptsize $\pm$0.00} & \textbf{0.78}\,{\scriptsize $\pm$0.00} & \textbf{0.74}\,{\scriptsize $\pm$0.00} & 1.00\,{\scriptsize $\pm$0.00} & \textbf{1}\,{\scriptsize $\pm$0.90} & 0\,{\scriptsize $\pm$0.00} \\
\bottomrule
\end{tabular}%
}
\end{table*}

\begin{table*}[t]
\centering
\caption{
\textbf{Full results on DINOv2-B/14 across dictionary widths and sparsity levels.}
We evaluate three expansion factors ($4\times$, $8\times$, $16\times$)
and three target sparsity levels ($\ell_0\approx32$, $64$, $128$).
We report achieved $\ell_0$, reconstruction ($R^2$), cross-seed
consistency (\texttt{DCS}, \texttt{MCC}), and dictionary utilization,
including usable atom fraction and numbers of dead and duplicated atoms.
Metrics are averaged across training seeds, while \texttt{DCS} and
\texttt{MCC} are averaged across seed pairs.
$\pm$ denotes the corresponding standard deviation.
}
\label{tab:appendix_dinov2_full}

\setlength{\tabcolsep}{3.5pt}
\renewcommand{\arraystretch}{0.90}

\resizebox{\textwidth}{!}{%
\begin{tabular}{ll l c c cc ccc}
\toprule
\multirow{2}{*}{\textbf{Expansion}} &
\multirow{2}{*}{\textbf{$\ell_0$ band}} &
\multirow{2}{*}{\textbf{Method}} &
\multicolumn{1}{c}{\textbf{Sparsity}} &
\multicolumn{1}{c}{\textbf{Reconstruction}} &
\multicolumn{2}{c}{\textbf{Consistency}} &
\multicolumn{3}{c}{\textbf{Dictionary Utilization}} \\
\cmidrule(lr){4-4}
\cmidrule(lr){5-5}
\cmidrule(lr){6-7}
\cmidrule(lr){8-10}
&
&
&
$\ell_0$ &
$R^2$ &
DCS &
MCC &
Usable atom frac. &
\# Dup. &
\# Dead \\
\midrule
\multirow{21}{*}{$4\times$}
& \multirow{7}{*}{$\ell_0 \approx 32$}
& ReLU       & 29\,{\scriptsize $\pm$0.0} & 0.93\,{\scriptsize $\pm$0.00} & 0.79\,{\scriptsize $\pm$0.00} & 0.75\,{\scriptsize $\pm$0.00} & 0.99\,{\scriptsize $\pm$0.00} & 26\,{\scriptsize $\pm$2.6} & 0\,{\scriptsize $\pm$0.00} \\
& & TopK       & 32\,{\scriptsize $\pm$0.0} & 0.96\,{\scriptsize $\pm$0.00} & 0.90\,{\scriptsize $\pm$0.00} & \textbf{0.88}\,{\scriptsize $\pm$0.00} & 1.00\,{\scriptsize $\pm$0.00} & 3\,{\scriptsize $\pm$2.02} & 0\,{\scriptsize $\pm$0.00} \\
& & BatchTopK  & 32\,{\scriptsize $\pm$0.0} & 0.96\,{\scriptsize $\pm$0.00} & 0.83\,{\scriptsize $\pm$0.00} & 0.81\,{\scriptsize $\pm$0.00} & 1.00\,{\scriptsize $\pm$0.00} & 0\,{\scriptsize $\pm$0.29} & 0\,{\scriptsize $\pm$0.00} \\
& & Matryoshka & 32\,{\scriptsize $\pm$0.0} & 0.95\,{\scriptsize $\pm$0.00} & 0.67\,{\scriptsize $\pm$0.00} & 0.67\,{\scriptsize $\pm$0.00} & 1.00\,{\scriptsize $\pm$0.00} & 9\,{\scriptsize $\pm$2.84} & 0\,{\scriptsize $\pm$0.00} \\
& & MP-SAE     & 32\,{\scriptsize $\pm$0.0} & 0.96\,{\scriptsize $\pm$0.00} & 0.87\,{\scriptsize $\pm$0.02} & 0.69\,{\scriptsize $\pm$0.01} & 0.89\,{\scriptsize $\pm$0.01} & 352\,{\scriptsize $\pm$28} & 0\,{\scriptsize $\pm$0.00} \\
& & iSAE       & 32\,{\scriptsize $\pm$0.0} & 0.95\,{\scriptsize $\pm$0.00} & 0.90\,{\scriptsize $\pm$0.00} & 0.49\,{\scriptsize $\pm$0.01} & 1.00\,{\scriptsize $\pm$0.00} & 0\,{\scriptsize $\pm$0.00} & 0\,{\scriptsize $\pm$0.58} \\
& & ConSDL     & 32\,{\scriptsize $\pm$0.0} & 0.95\,{\scriptsize $\pm$0.00} & \textbf{0.90}\,{\scriptsize $\pm$0.00} & 0.87\,{\scriptsize $\pm$0.00} & 1.00\,{\scriptsize $\pm$0.00} & 0\,{\scriptsize $\pm$0.00} & 0\,{\scriptsize $\pm$0.00} \\
\cmidrule(lr){2-10}
& \multirow{7}{*}{$\ell_0 \approx 64$}
& ReLU       & 66\,{\scriptsize $\pm$0.5} & 0.95\,{\scriptsize $\pm$0.00} & 0.88\,{\scriptsize $\pm$0.00} & 0.85\,{\scriptsize $\pm$0.00} & 1.00\,{\scriptsize $\pm$0.00} & 9\,{\scriptsize $\pm$1.42} & 0\,{\scriptsize $\pm$0.00} \\
& & TopK       & 64\,{\scriptsize $\pm$0.0} & 0.97\,{\scriptsize $\pm$0.00} & 0.89\,{\scriptsize $\pm$0.00} & \textbf{0.89}\,{\scriptsize $\pm$0.00} & 1.00\,{\scriptsize $\pm$0.00} & 1\,{\scriptsize $\pm$0.76} & 0\,{\scriptsize $\pm$0.00} \\
& & BatchTopK  & 64\,{\scriptsize $\pm$0.2} & 0.97\,{\scriptsize $\pm$0.00} & 0.88\,{\scriptsize $\pm$0.00} & 0.87\,{\scriptsize $\pm$0.00} & 1.00\,{\scriptsize $\pm$0.00} & 0\,{\scriptsize $\pm$0.58} & 0\,{\scriptsize $\pm$0.00} \\
& & Matryoshka & 64\,{\scriptsize $\pm$0.0} & 0.96\,{\scriptsize $\pm$0.00} & 0.67\,{\scriptsize $\pm$0.00} & 0.70\,{\scriptsize $\pm$0.00} & 1.00\,{\scriptsize $\pm$0.00} & 0\,{\scriptsize $\pm$0.50} & 0\,{\scriptsize $\pm$0.00} \\
& & MP-SAE     & 64\,{\scriptsize $\pm$0.0} & 0.97\,{\scriptsize $\pm$0.00} & 0.87\,{\scriptsize $\pm$0.02} & 0.69\,{\scriptsize $\pm$0.01} & 0.89\,{\scriptsize $\pm$0.01} & 352\,{\scriptsize $\pm$28} & 0\,{\scriptsize $\pm$0.00} \\
& & iSAE       & 64\,{\scriptsize $\pm$0.0} & 0.96\,{\scriptsize $\pm$0.00} & 0.89\,{\scriptsize $\pm$0.00} & 0.50\,{\scriptsize $\pm$0.01} & 1.00\,{\scriptsize $\pm$0.00} & 0\,{\scriptsize $\pm$0.00} & 2\,{\scriptsize $\pm$1.00} \\
& & ConSDL     & 66\,{\scriptsize $\pm$1.4} & \textbf{0.97}\,{\scriptsize $\pm$0.00} & \textbf{0.90}\,{\scriptsize $\pm$0.00} & 0.86\,{\scriptsize $\pm$0.01} & 1.00\,{\scriptsize $\pm$0.00} & 0\,{\scriptsize $\pm$0.00} & 0\,{\scriptsize $\pm$0.00} \\
\cmidrule(lr){2-10}
& \multirow{7}{*}{$\ell_0 \approx 128$}
& ReLU       & 123\,{\scriptsize $\pm$0} & 0.96\,{\scriptsize $\pm$0.00} & 0.86\,{\scriptsize $\pm$0.00} & 0.84\,{\scriptsize $\pm$0.00} & 1.00\,{\scriptsize $\pm$0.00} & 7\,{\scriptsize $\pm$0.63} & 0\,{\scriptsize $\pm$0.00} \\
& & TopK       & 128\,{\scriptsize $\pm$0} & 0.97\,{\scriptsize $\pm$0.00} & 0.82\,{\scriptsize $\pm$0.00} & 0.84\,{\scriptsize $\pm$0.00} & 1.00\,{\scriptsize $\pm$0.00} & 0\,{\scriptsize $\pm$0.29} & 0\,{\scriptsize $\pm$0.00} \\
& & BatchTopK  & 128\,{\scriptsize $\pm$0} & 0.97\,{\scriptsize $\pm$0.00} & 0.80\,{\scriptsize $\pm$0.00} & 0.83\,{\scriptsize $\pm$0.00} & 1.00\,{\scriptsize $\pm$0.00} & 1\,{\scriptsize $\pm$0.76} & 0\,{\scriptsize $\pm$0.00} \\
& & Matryoshka & 128\,{\scriptsize $\pm$0} & 0.96\,{\scriptsize $\pm$0.00} & 0.57\,{\scriptsize $\pm$0.00} & 0.63\,{\scriptsize $\pm$0.00} & 1.00\,{\scriptsize $\pm$0.00} & 1\,{\scriptsize $\pm$0.29} & 0\,{\scriptsize $\pm$0.00} \\
& & MP-SAE     & 127\,{\scriptsize $\pm$0} & 0.98\,{\scriptsize $\pm$0.00} & 0.87\,{\scriptsize $\pm$0.02} & 0.69\,{\scriptsize $\pm$0.01} & 0.89\,{\scriptsize $\pm$0.01} & 352\,{\scriptsize $\pm$28} & 0\,{\scriptsize $\pm$0.00} \\
& & iSAE       & 128\,{\scriptsize $\pm$0} & 0.97\,{\scriptsize $\pm$0.00} & 0.86\,{\scriptsize $\pm$0.00} & 0.51\,{\scriptsize $\pm$0.00} & 1.00\,{\scriptsize $\pm$0.00} & 0\,{\scriptsize $\pm$0.00} & 0\,{\scriptsize $\pm$0.00} \\
& & ConSDL     & 138\,{\scriptsize $\pm$0} & \textbf{0.98}\,{\scriptsize $\pm$0.00} & \textbf{0.88}\,{\scriptsize $\pm$0.00} & 0.84\,{\scriptsize $\pm$0.00} & 1.00\,{\scriptsize $\pm$0.00} & 0\,{\scriptsize $\pm$0.00} & 0\,{\scriptsize $\pm$0.00} \\
\midrule
\multirow{21}{*}{$8\times$}
& \multirow{7}{*}{$\ell_0 \approx 32$}
& ReLU       & 32\,{\scriptsize $\pm$0.4} & 0.94\,{\scriptsize $\pm$0.00} & 0.57\,{\scriptsize $\pm$0.00} & 0.49\,{\scriptsize $\pm$0.00} & 0.99\,{\scriptsize $\pm$0.00} & 89\,{\scriptsize $\pm$11.5} & 0\,{\scriptsize $\pm$0.00} \\
& & TopK       & 32\,{\scriptsize $\pm$0.0} & 0.96\,{\scriptsize $\pm$0.00} & 0.75\,{\scriptsize $\pm$0.00} & 0.67\,{\scriptsize $\pm$0.01} & 0.99\,{\scriptsize $\pm$0.00} & 35\,{\scriptsize $\pm$9.6} & 0\,{\scriptsize $\pm$0.00} \\
& & BatchTopK  & 32\,{\scriptsize $\pm$0.0} & 0.96\,{\scriptsize $\pm$0.00} & 0.60\,{\scriptsize $\pm$0.00} & 0.53\,{\scriptsize $\pm$0.00} & 1.00\,{\scriptsize $\pm$0.00} & 4\,{\scriptsize $\pm$1.80} & 0\,{\scriptsize $\pm$0.00} \\
& & Matryoshka & 32\,{\scriptsize $\pm$0.0} & 0.95\,{\scriptsize $\pm$0.00} & 0.50\,{\scriptsize $\pm$0.00} & 0.43\,{\scriptsize $\pm$0.00} & 0.98\,{\scriptsize $\pm$0.00} & 121\,{\scriptsize $\pm$12} & 0\,{\scriptsize $\pm$0.00} \\
& & MP-SAE     & 32\,{\scriptsize $\pm$0.0} & \textbf{0.96}\,{\scriptsize $\pm$0.00} & 0.83\,{\scriptsize $\pm$0.00} & 0.49\,{\scriptsize $\pm$0.00} & 0.74\,{\scriptsize $\pm$0.01} & 1577\,{\scriptsize $\pm$64} & 0\,{\scriptsize $\pm$0.00} \\
& & iSAE       & 32\,{\scriptsize $\pm$0.0} & 0.96\,{\scriptsize $\pm$0.00} & 0.77\,{\scriptsize $\pm$0.00} & 0.41\,{\scriptsize $\pm$0.00} & 1.00\,{\scriptsize $\pm$0.00} & 4\,{\scriptsize $\pm$1.50} & 14\,{\scriptsize $\pm$5.1} \\
& & ConSDL     & 34\,{\scriptsize $\pm$1.3} & 0.96\,{\scriptsize $\pm$0.00} & \textbf{0.84}\,{\scriptsize $\pm$0.00} & \textbf{0.76}\,{\scriptsize $\pm$0.00} & \textbf{1.00}\,{\scriptsize $\pm$0.00} & \textbf{0}\,{\scriptsize $\pm$0.00} & 0\,{\scriptsize $\pm$0.00} \\
\cmidrule(lr){2-10}
& \multirow{7}{*}{$\ell_0 \approx 64$}
& ReLU       & 54\,{\scriptsize $\pm$0.6} & 0.95\,{\scriptsize $\pm$0.00} & 0.74\,{\scriptsize $\pm$0.00} & 0.66\,{\scriptsize $\pm$0.00} & 0.98\,{\scriptsize $\pm$0.00} & 148\,{\scriptsize $\pm$5} & 0\,{\scriptsize $\pm$0.00} \\
& & TopK       & 64\,{\scriptsize $\pm$0.0} & 0.97\,{\scriptsize $\pm$0.00} & 0.78\,{\scriptsize $\pm$0.00} & 0.75\,{\scriptsize $\pm$0.00} & 1.00\,{\scriptsize $\pm$0.00} & 22\,{\scriptsize $\pm$3.1} & 0\,{\scriptsize $\pm$0.00} \\
& & BatchTopK  & 64\,{\scriptsize $\pm$0.0} & 0.97\,{\scriptsize $\pm$0.00} & 0.67\,{\scriptsize $\pm$0.00} & 0.63\,{\scriptsize $\pm$0.00} & 1.00\,{\scriptsize $\pm$0.00} & 5\,{\scriptsize $\pm$1.44} & 0\,{\scriptsize $\pm$0.00} \\
& & Matryoshka & 64\,{\scriptsize $\pm$0.0} & 0.96\,{\scriptsize $\pm$0.00} & 0.53\,{\scriptsize $\pm$0.00} & 0.50\,{\scriptsize $\pm$0.00} & 0.98\,{\scriptsize $\pm$0.00} & 94\,{\scriptsize $\pm$8.4} & 0\,{\scriptsize $\pm$0.00} \\
& & MP-SAE     & 64\,{\scriptsize $\pm$0.0} & \textbf{0.97}\,{\scriptsize $\pm$0.00} & 0.83\,{\scriptsize $\pm$0.00} & 0.49\,{\scriptsize $\pm$0.00} & 0.74\,{\scriptsize $\pm$0.01} & 1577\,{\scriptsize $\pm$64} & 0\,{\scriptsize $\pm$0.00} \\
& & iSAE       & 64\,{\scriptsize $\pm$0.0} & 0.97\,{\scriptsize $\pm$0.00} & 0.79\,{\scriptsize $\pm$0.00} & 0.47\,{\scriptsize $\pm$0.00} & 1.00\,{\scriptsize $\pm$0.00} & 2\,{\scriptsize $\pm$0.58} & 10\,{\scriptsize $\pm$2.6} \\
& & ConSDL     & 61\,{\scriptsize $\pm$0.4} & 0.96\,{\scriptsize $\pm$0.00} & \textbf{0.85}\,{\scriptsize $\pm$0.00} & \textbf{0.81}\,{\scriptsize $\pm$0.00} & \textbf{1.00}\,{\scriptsize $\pm$0.00} & \textbf{0}\,{\scriptsize $\pm$0.00} & 0\,{\scriptsize $\pm$0.00} \\
\cmidrule(lr){2-10}
& \multirow{7}{*}{$\ell_0 \approx 128$}
& ReLU       & 128\,{\scriptsize $\pm$0} & 0.96\,{\scriptsize $\pm$0.00} & 0.75\,{\scriptsize $\pm$0.00} & 0.68\,{\scriptsize $\pm$0.00} & 0.99\,{\scriptsize $\pm$0.00} & 57\,{\scriptsize $\pm$4.9} & 0\,{\scriptsize $\pm$0.00} \\
& & TopK       & 128\,{\scriptsize $\pm$0} & 0.98\,{\scriptsize $\pm$0.00} & 0.73\,{\scriptsize $\pm$0.00} & 0.74\,{\scriptsize $\pm$0.00} & 1.00\,{\scriptsize $\pm$0.00} & 20\,{\scriptsize $\pm$1.3} & 0\,{\scriptsize $\pm$0.00} \\
& & BatchTopK  & 128\,{\scriptsize $\pm$0} & 0.98\,{\scriptsize $\pm$0.00} & 0.65\,{\scriptsize $\pm$0.00} & 0.65\,{\scriptsize $\pm$0.00} & 1.00\,{\scriptsize $\pm$0.00} & 9\,{\scriptsize $\pm$0.58} & 0\,{\scriptsize $\pm$0.00} \\
& & Matryoshka & 128\,{\scriptsize $\pm$0} & 0.97\,{\scriptsize $\pm$0.00} & 0.48\,{\scriptsize $\pm$0.00} & 0.48\,{\scriptsize $\pm$0.00} & 0.99\,{\scriptsize $\pm$0.00} & 62\,{\scriptsize $\pm$6.9} & 0\,{\scriptsize $\pm$0.00} \\
& & MP-SAE     & 127\,{\scriptsize $\pm$0} & \textbf{0.98}\,{\scriptsize $\pm$0.00} & \textbf{0.83}\,{\scriptsize $\pm$0.00} & 0.49\,{\scriptsize $\pm$0.00} & 0.74\,{\scriptsize $\pm$0.01} & 1577\,{\scriptsize $\pm$64} & 0\,{\scriptsize $\pm$0.00} \\
& & iSAE       & 128\,{\scriptsize $\pm$0} & 0.97\,{\scriptsize $\pm$0.00} & 0.78\,{\scriptsize $\pm$0.00} & 0.51\,{\scriptsize $\pm$0.01} & 1.00\,{\scriptsize $\pm$0.00} & 0\,{\scriptsize $\pm$0.29} & 3\,{\scriptsize $\pm$1.53} \\
& & ConSDL     & 122\,{\scriptsize $\pm$2} & 0.98\,{\scriptsize $\pm$0.00} & 0.81\,{\scriptsize $\pm$0.00} & \textbf{0.74}\,{\scriptsize $\pm$0.00} & \textbf{1.00}\,{\scriptsize $\pm$0.00} & \textbf{0}\,{\scriptsize $\pm$0.00} & 0\,{\scriptsize $\pm$0.00} \\
\midrule
\multirow{21}{*}{$16\times$}
& \multirow{7}{*}{$\ell_0 \approx 32$}
& ReLU       & 33\,{\scriptsize $\pm$0.0} & 0.94\,{\scriptsize $\pm$0.00} & 0.38\,{\scriptsize $\pm$0.00} & 0.28\,{\scriptsize $\pm$0.00} & 0.98\,{\scriptsize $\pm$0.00} & 187\,{\scriptsize $\pm$10} & 96\,{\scriptsize $\pm$35.2} \\
& & TopK       & 32\,{\scriptsize $\pm$0.0} & 0.96\,{\scriptsize $\pm$0.00} & 0.50\,{\scriptsize $\pm$0.04} & 0.37\,{\scriptsize $\pm$0.02} & 0.88\,{\scriptsize $\pm$0.08} & 218\,{\scriptsize $\pm$116} & 1300\,{\scriptsize $\pm$1126} \\
& & BatchTopK  & 32\,{\scriptsize $\pm$0.0} & 0.96\,{\scriptsize $\pm$0.00} & 0.42\,{\scriptsize $\pm$0.00} & 0.29\,{\scriptsize $\pm$0.00} & \textbf{1.00}\,{\scriptsize $\pm$0.00} & 2\,{\scriptsize $\pm$1.32} & 0\,{\scriptsize $\pm$0.00} \\
& & Matryoshka & 32\,{\scriptsize $\pm$0.0} & 0.95\,{\scriptsize $\pm$0.00} & 0.34\,{\scriptsize $\pm$0.00} & 0.22\,{\scriptsize $\pm$0.00} & 0.99\,{\scriptsize $\pm$0.00} & 64\,{\scriptsize $\pm$6.9} & 0\,{\scriptsize $\pm$0.58} \\
& & MP-SAE     & 32\,{\scriptsize $\pm$0.0} & \textbf{0.96}\,{\scriptsize $\pm$0.00} & \textbf{0.78}\,{\scriptsize $\pm$0.00} & 0.31\,{\scriptsize $\pm$0.00} & 0.62\,{\scriptsize $\pm$0.00} & 4640\,{\scriptsize $\pm$39} & 0\,{\scriptsize $\pm$0.00} \\
& & iSAE       & 32\,{\scriptsize $\pm$0.0} & 0.96\,{\scriptsize $\pm$0.00} & 0.63\,{\scriptsize $\pm$0.00} & 0.29\,{\scriptsize $\pm$0.00} & 0.99\,{\scriptsize $\pm$0.00} & 52\,{\scriptsize $\pm$5.3} & 94\,{\scriptsize $\pm$29.8} \\
& & ConSDL     & 36\,{\scriptsize $\pm$0.0} & 0.95\,{\scriptsize $\pm$0.00} & 0.75\,{\scriptsize $\pm$0.00} & \textbf{0.66}\,{\scriptsize $\pm$0.00} & 0.94\,{\scriptsize $\pm$0.00} & \textbf{0}\,{\scriptsize $\pm$0.00} & 704\,{\scriptsize $\pm$36} \\
\cmidrule(lr){2-10}
& \multirow{7}{*}{$\ell_0 \approx 64$}
& ReLU       & 62\,{\scriptsize $\pm$0.7} & 0.95\,{\scriptsize $\pm$0.00} & 0.53\,{\scriptsize $\pm$0.00} & 0.42\,{\scriptsize $\pm$0.00} & 0.96\,{\scriptsize $\pm$0.00} & 414\,{\scriptsize $\pm$35} & 29\,{\scriptsize $\pm$23.3} \\
& & TopK       & 64\,{\scriptsize $\pm$0.0} & 0.97\,{\scriptsize $\pm$0.00} & 0.60\,{\scriptsize $\pm$0.00} & 0.51\,{\scriptsize $\pm$0.01} & 0.99\,{\scriptsize $\pm$0.00} & 70\,{\scriptsize $\pm$8.4} & 0\,{\scriptsize $\pm$0.00} \\
& & BatchTopK  & 64\,{\scriptsize $\pm$0.0} & 0.97\,{\scriptsize $\pm$0.00} & 0.48\,{\scriptsize $\pm$0.00} & 0.41\,{\scriptsize $\pm$0.00} & 1.00\,{\scriptsize $\pm$0.00} & 35\,{\scriptsize $\pm$7.0} & 0\,{\scriptsize $\pm$0.00} \\
& & Matryoshka & 64\,{\scriptsize $\pm$2.1} & 0.96\,{\scriptsize $\pm$0.00} & 0.38\,{\scriptsize $\pm$0.00} & 0.31\,{\scriptsize $\pm$0.02} & 0.99\,{\scriptsize $\pm$0.00} & 126\,{\scriptsize $\pm$31} & 0\,{\scriptsize $\pm$0.00} \\
& & MP-SAE     & 64\,{\scriptsize $\pm$0.0} & \textbf{0.97}\,{\scriptsize $\pm$0.00} & 0.78\,{\scriptsize $\pm$0.00} & 0.31\,{\scriptsize $\pm$0.00} & 0.62\,{\scriptsize $\pm$0.00} & 4640\,{\scriptsize $\pm$39} & 0\,{\scriptsize $\pm$0.00} \\
& & iSAE       & 64\,{\scriptsize $\pm$0.0} & 0.97\,{\scriptsize $\pm$0.00} & 0.65\,{\scriptsize $\pm$0.00} & 0.36\,{\scriptsize $\pm$0.00} & 0.98\,{\scriptsize $\pm$0.00} & 233\,{\scriptsize $\pm$6} & 52\,{\scriptsize $\pm$24.7} \\
& & ConSDL     & 71\,{\scriptsize $\pm$0.0} & 0.95\,{\scriptsize $\pm$0.00} & \textbf{0.80}\,{\scriptsize $\pm$0.00} & \textbf{0.75}\,{\scriptsize $\pm$0.00} & \textbf{1.00}\,{\scriptsize $\pm$0.00} & \textbf{14}\,{\scriptsize $\pm$1.4} & 0\,{\scriptsize $\pm$0.00} \\
\cmidrule(lr){2-10}
& \multirow{7}{*}{$\ell_0 \approx 128$}
& ReLU       & 125\,{\scriptsize $\pm$1} & 0.96\,{\scriptsize $\pm$0.00} & 0.62\,{\scriptsize $\pm$0.00} & 0.50\,{\scriptsize $\pm$0.00} & 0.96\,{\scriptsize $\pm$0.00} & 462\,{\scriptsize $\pm$11} & 0\,{\scriptsize $\pm$0.00} \\
& & TopK       & 128\,{\scriptsize $\pm$0} & 0.98\,{\scriptsize $\pm$0.00} & 0.58\,{\scriptsize $\pm$0.00} & 0.53\,{\scriptsize $\pm$0.00} & 0.99\,{\scriptsize $\pm$0.00} & 62\,{\scriptsize $\pm$4.6} & 0\,{\scriptsize $\pm$0.00} \\
& & BatchTopK  & 128\,{\scriptsize $\pm$0} & 0.98\,{\scriptsize $\pm$0.00} & 0.48\,{\scriptsize $\pm$0.00} & 0.44\,{\scriptsize $\pm$0.00} & 1.00\,{\scriptsize $\pm$0.00} & 27\,{\scriptsize $\pm$5.6} & 0\,{\scriptsize $\pm$0.00} \\
& & Matryoshka & 128\,{\scriptsize $\pm$0} & 0.97\,{\scriptsize $\pm$0.00} & 0.35\,{\scriptsize $\pm$0.00} & 0.30\,{\scriptsize $\pm$0.00} & 0.99\,{\scriptsize $\pm$0.00} & 88\,{\scriptsize $\pm$3.3} & 0\,{\scriptsize $\pm$0.00} \\
& & MP-SAE     & 128\,{\scriptsize $\pm$0} & \textbf{0.98}\,{\scriptsize $\pm$0.00} & 0.78\,{\scriptsize $\pm$0.00} & 0.31\,{\scriptsize $\pm$0.00} & 0.62\,{\scriptsize $\pm$0.00} & 4640\,{\scriptsize $\pm$39} & 0\,{\scriptsize $\pm$0.00} \\
& & iSAE       & 128\,{\scriptsize $\pm$0} & 0.98\,{\scriptsize $\pm$0.00} & 0.68\,{\scriptsize $\pm$0.00} & 0.44\,{\scriptsize $\pm$0.00} & 0.99\,{\scriptsize $\pm$0.00} & 88\,{\scriptsize $\pm$2.3} & 4\,{\scriptsize $\pm$0.58} \\
& & ConSDL     & 138\,{\scriptsize $\pm$0} & 0.95\,{\scriptsize $\pm$0.00} & \textbf{0.81}\,{\scriptsize $\pm$0.00} & \textbf{0.77}\,{\scriptsize $\pm$0.00} & 1.00\,{\scriptsize $\pm$0.00} & \textbf{24}\,{\scriptsize $\pm$3.6} & 0\,{\scriptsize $\pm$0.00} \\
\bottomrule
\end{tabular}%
}
\end{table*}

\begin{table*}[t]
\centering
\caption{
\textbf{Full results on GPT-2 Small across dictionary widths and sparsity levels.}
We evaluate three expansion factors ($4\times$, $8\times$, $16\times$)
and three target sparsity levels ($\ell_0\approx32$, $64$, $128$).
We report achieved $\ell_0$, reconstruction ($R^2$), cross-seed
consistency (\texttt{DCS}, \texttt{MCC}), and dictionary utilization,
including usable atom fraction and numbers of dead and duplicated atoms.
Metrics are averaged across training seeds, while \texttt{DCS} and
\texttt{MCC} are averaged across seed pairs.
$\pm$ denotes the corresponding standard deviation.
}
\label{tab:appendix_gpt2_full}

\setlength{\tabcolsep}{3.5pt}
\renewcommand{\arraystretch}{0.90}

\resizebox{\textwidth}{!}{%
\begin{tabular}{ll l c c cc ccc}
\toprule
\multirow{2}{*}{\textbf{Expansion}} &
\multirow{2}{*}{\textbf{$\ell_0$ band}} &
\multirow{2}{*}{\textbf{Method}} &
\multicolumn{1}{c}{\textbf{Sparsity}} &
\multicolumn{1}{c}{\textbf{Reconstruction}} &
\multicolumn{2}{c}{\textbf{Consistency}} &
\multicolumn{3}{c}{\textbf{Dictionary Utilization}} \\
\cmidrule(lr){4-4}
\cmidrule(lr){5-5}
\cmidrule(lr){6-7}
\cmidrule(lr){8-10}
&
&
&
$\ell_0$ &
$R^2$ &
DCS &
MCC &
Usable atom frac. &
\# Dup. &
\# Dead \\
\midrule
\multirow{21}{*}{$4\times$}
& \multirow{7}{*}{$\ell_0 \approx 32$}
& ReLU       & 31\,{\scriptsize $\pm$0.2} & 0.87\,{\scriptsize $\pm$0.02} & \textbf{0.91}\,{\scriptsize $\pm$0.00} & \textbf{0.86}\,{\scriptsize $\pm$0.00} & 0.95\,{\scriptsize $\pm$0.00} & 161\,{\scriptsize $\pm$8} & 6\,{\scriptsize $\pm$3.00} \\
& & TopK       & 32\,{\scriptsize $\pm$0.0} & 0.89\,{\scriptsize $\pm$0.00} & 0.83\,{\scriptsize $\pm$0.00} & 0.79\,{\scriptsize $\pm$0.00} & 0.99\,{\scriptsize $\pm$0.00} & 27\,{\scriptsize $\pm$3.4} & 0\,{\scriptsize $\pm$0.00} \\
& & BatchTopK  & 32\,{\scriptsize $\pm$0.1} & 0.90\,{\scriptsize $\pm$0.00} & 0.83\,{\scriptsize $\pm$0.00} & 0.79\,{\scriptsize $\pm$0.00} & 0.99\,{\scriptsize $\pm$0.00} & 17\,{\scriptsize $\pm$3.6} & 0\,{\scriptsize $\pm$0.00} \\
& & Matryoshka & 32\,{\scriptsize $\pm$0.1} & 0.88\,{\scriptsize $\pm$0.00} & 0.71\,{\scriptsize $\pm$0.00} & 0.69\,{\scriptsize $\pm$0.00} & 1.00\,{\scriptsize $\pm$0.00} & 0\,{\scriptsize $\pm$0.00} & 0\,{\scriptsize $\pm$0.00} \\
& & MP-SAE     & 32\,{\scriptsize $\pm$0.0} & \textbf{0.98}\,{\scriptsize $\pm$0.00} & 0.87\,{\scriptsize $\pm$0.00} & 0.64\,{\scriptsize $\pm$0.01} & 0.83\,{\scriptsize $\pm$0.00} & 517\,{\scriptsize $\pm$13} & 0\,{\scriptsize $\pm$0.00} \\
& & iSAE       & 32\,{\scriptsize $\pm$0.0} & 0.89\,{\scriptsize $\pm$0.00} & 0.85\,{\scriptsize $\pm$0.00} & 0.47\,{\scriptsize $\pm$0.01} & 1.00\,{\scriptsize $\pm$0.00} & 0\,{\scriptsize $\pm$0.00} & 0\,{\scriptsize $\pm$0.00} \\
& & ConSDL     & 34\,{\scriptsize $\pm$0.1} & 0.90\,{\scriptsize $\pm$0.00} & 0.89\,{\scriptsize $\pm$0.01} & 0.83\,{\scriptsize $\pm$0.01} & 1.00\,{\scriptsize $\pm$0.00} & 0\,{\scriptsize $\pm$0.00} & 0\,{\scriptsize $\pm$0.00} \\
\cmidrule(lr){2-10}
& \multirow{7}{*}{$\ell_0 \approx 64$}
& ReLU       & 68\,{\scriptsize $\pm$0.2} & 0.89\,{\scriptsize $\pm$0.00} & 0.89\,{\scriptsize $\pm$0.00} & 0.86\,{\scriptsize $\pm$0.00} & 0.97\,{\scriptsize $\pm$0.00} & 82\,{\scriptsize $\pm$2.0} & 0\,{\scriptsize $\pm$0.00} \\
& & TopK       & 64\,{\scriptsize $\pm$0.0} & 0.92\,{\scriptsize $\pm$0.00} & 0.80\,{\scriptsize $\pm$0.00} & 0.78\,{\scriptsize $\pm$0.00} & 1.00\,{\scriptsize $\pm$0.00} & 4\,{\scriptsize $\pm$0.50} & 0\,{\scriptsize $\pm$0.00} \\
& & BatchTopK  & 64\,{\scriptsize $\pm$0.1} & 0.92\,{\scriptsize $\pm$0.00} & 0.79\,{\scriptsize $\pm$0.00} & 0.78\,{\scriptsize $\pm$0.00} & 1.00\,{\scriptsize $\pm$0.00} & 6\,{\scriptsize $\pm$1.61} & 0\,{\scriptsize $\pm$0.00} \\
& & Matryoshka & 64\,{\scriptsize $\pm$0.1} & 0.91\,{\scriptsize $\pm$0.00} & 0.67\,{\scriptsize $\pm$0.00} & 0.67\,{\scriptsize $\pm$0.00} & 1.00\,{\scriptsize $\pm$0.00} & 0\,{\scriptsize $\pm$0.87} & 0\,{\scriptsize $\pm$0.00} \\
& & MP-SAE     & 64\,{\scriptsize $\pm$0.0} & \textbf{0.98}\,{\scriptsize $\pm$0.00} & 0.87\,{\scriptsize $\pm$0.00} & 0.64\,{\scriptsize $\pm$0.01} & 0.83\,{\scriptsize $\pm$0.00} & 517\,{\scriptsize $\pm$13} & 0\,{\scriptsize $\pm$0.00} \\
& & iSAE       & 64\,{\scriptsize $\pm$0.0} & 0.91\,{\scriptsize $\pm$0.00} & 0.83\,{\scriptsize $\pm$0.00} & 0.48\,{\scriptsize $\pm$0.00} & 1.00\,{\scriptsize $\pm$0.00} & \textbf{0}\,{\scriptsize $\pm$0.00} & 0\,{\scriptsize $\pm$0.00} \\
& & ConSDL     & 62\,{\scriptsize $\pm$1.7} & 0.90\,{\scriptsize $\pm$0.00} & \textbf{0.90}\,{\scriptsize $\pm$0.00} & \textbf{0.86}\,{\scriptsize $\pm$0.00} & 1.00\,{\scriptsize $\pm$0.00} & 0\,{\scriptsize $\pm$0.38} & 0\,{\scriptsize $\pm$0.00} \\
\cmidrule(lr){2-10}
& \multirow{7}{*}{$\ell_0 \approx 128$}
& ReLU       & 134\,{\scriptsize $\pm$1} & 0.91\,{\scriptsize $\pm$0.00} & 0.87\,{\scriptsize $\pm$0.00} & \textbf{0.84}\,{\scriptsize $\pm$0.00} & 0.98\,{\scriptsize $\pm$0.00} & 48\,{\scriptsize $\pm$0.5} & 0\,{\scriptsize $\pm$0.00} \\
& & TopK       & 128\,{\scriptsize $\pm$0} & 0.94\,{\scriptsize $\pm$0.00} & 0.71\,{\scriptsize $\pm$0.00} & 0.72\,{\scriptsize $\pm$0.00} & 1.00\,{\scriptsize $\pm$0.00} & 0\,{\scriptsize $\pm$0.87} & 0\,{\scriptsize $\pm$0.00} \\
& & BatchTopK  & 128\,{\scriptsize $\pm$0} & 0.94\,{\scriptsize $\pm$0.00} & 0.70\,{\scriptsize $\pm$0.00} & 0.73\,{\scriptsize $\pm$0.00} & 1.00\,{\scriptsize $\pm$0.00} & 4\,{\scriptsize $\pm$0.76} & 0\,{\scriptsize $\pm$0.00} \\
& & Matryoshka & 128\,{\scriptsize $\pm$0} & 0.93\,{\scriptsize $\pm$0.00} & 0.59\,{\scriptsize $\pm$0.00} & 0.62\,{\scriptsize $\pm$0.00} & 1.00\,{\scriptsize $\pm$0.00} & 0\,{\scriptsize $\pm$0.00} & 0\,{\scriptsize $\pm$0.00} \\
& & MP-SAE     & 127\,{\scriptsize $\pm$0} & \textbf{0.99}\,{\scriptsize $\pm$0.00} & 0.87\,{\scriptsize $\pm$0.00} & 0.64\,{\scriptsize $\pm$0.01} & 0.83\,{\scriptsize $\pm$0.00} & 517\,{\scriptsize $\pm$13} & 0\,{\scriptsize $\pm$0.00} \\
& & iSAE       & 128\,{\scriptsize $\pm$0} & 0.94\,{\scriptsize $\pm$0.00} & 0.78\,{\scriptsize $\pm$0.00} & 0.48\,{\scriptsize $\pm$0.00} & 1.00\,{\scriptsize $\pm$0.00} & 0\,{\scriptsize $\pm$0.00} & 0\,{\scriptsize $\pm$0.00} \\
& & ConSDL     & 126\,{\scriptsize $\pm$0} & 0.94\,{\scriptsize $\pm$0.00} & \textbf{0.89}\,{\scriptsize $\pm$0.00} & 0.84\,{\scriptsize $\pm$0.00} & 1.00\,{\scriptsize $\pm$0.00} & 0\,{\scriptsize $\pm$0.00} & 0\,{\scriptsize $\pm$0.00} \\
\midrule
\multirow{21}{*}{$8\times$}
& \multirow{7}{*}{$\ell_0 \approx 32$}
& ReLU       & 33\,{\scriptsize $\pm$0.3} & 0.87\,{\scriptsize $\pm$0.00} & 0.79\,{\scriptsize $\pm$0.01} & 0.73\,{\scriptsize $\pm$0.01} & 0.80\,{\scriptsize $\pm$0.04} & 357\,{\scriptsize $\pm$40} & 845\,{\scriptsize $\pm$365} \\
& & TopK       & 32\,{\scriptsize $\pm$0.0} & 0.91\,{\scriptsize $\pm$0.00} & 0.82\,{\scriptsize $\pm$0.00} & 0.77\,{\scriptsize $\pm$0.00} & 0.99\,{\scriptsize $\pm$0.00} & 91\,{\scriptsize $\pm$6.0} & 0\,{\scriptsize $\pm$0.00} \\
& & BatchTopK  & 32\,{\scriptsize $\pm$0.1} & 0.91\,{\scriptsize $\pm$0.00} & 0.82\,{\scriptsize $\pm$0.00} & 0.77\,{\scriptsize $\pm$0.00} & 0.99\,{\scriptsize $\pm$0.00} & 91\,{\scriptsize $\pm$10.3} & 0\,{\scriptsize $\pm$0.00} \\
& & Matryoshka & 32\,{\scriptsize $\pm$0.1} & 0.90\,{\scriptsize $\pm$0.00} & 0.69\,{\scriptsize $\pm$0.00} & 0.68\,{\scriptsize $\pm$0.00} & 1.00\,{\scriptsize $\pm$0.00} & 0\,{\scriptsize $\pm$0.50} & 0\,{\scriptsize $\pm$0.00} \\
& & MP-SAE     & 32\,{\scriptsize $\pm$0.0} & \textbf{0.98}\,{\scriptsize $\pm$0.00} & 0.86\,{\scriptsize $\pm$0.00} & 0.56\,{\scriptsize $\pm$0.00} & 0.79\,{\scriptsize $\pm$0.00} & 1292\,{\scriptsize $\pm$8} & 0\,{\scriptsize $\pm$0.00} \\
& & iSAE       & 32\,{\scriptsize $\pm$0.0} & 0.90\,{\scriptsize $\pm$0.00} & 0.82\,{\scriptsize $\pm$0.00} & 0.45\,{\scriptsize $\pm$0.00} & 1.00\,{\scriptsize $\pm$0.00} & \textbf{0}\,{\scriptsize $\pm$0.00} & 0\,{\scriptsize $\pm$0.00} \\
& & ConSDL     & 37\,{\scriptsize $\pm$0.2} & 0.90\,{\scriptsize $\pm$0.00} & \textbf{0.90}\,{\scriptsize $\pm$0.00} & \textbf{0.84}\,{\scriptsize $\pm$0.00} & 1.00\,{\scriptsize $\pm$0.00} & 2\,{\scriptsize $\pm$0.43} & 0\,{\scriptsize $\pm$0.00} \\
\cmidrule(lr){2-10}
& \multirow{7}{*}{$\ell_0 \approx 64$}
& ReLU       & 67\,{\scriptsize $\pm$0.2} & 0.91\,{\scriptsize $\pm$0.01} & 0.89\,{\scriptsize $\pm$0.00} & 0.84\,{\scriptsize $\pm$0.00} & 0.95\,{\scriptsize $\pm$0.00} & 288\,{\scriptsize $\pm$7} & 0\,{\scriptsize $\pm$0.00} \\
& & TopK       & 64\,{\scriptsize $\pm$0.0} & 0.93\,{\scriptsize $\pm$0.00} & 0.78\,{\scriptsize $\pm$0.00} & 0.74\,{\scriptsize $\pm$0.00} & 1.00\,{\scriptsize $\pm$0.00} & 25\,{\scriptsize $\pm$3.4} & 0\,{\scriptsize $\pm$0.00} \\
& & BatchTopK  & 64\,{\scriptsize $\pm$0.1} & 0.93\,{\scriptsize $\pm$0.00} & 0.78\,{\scriptsize $\pm$0.00} & 0.76\,{\scriptsize $\pm$0.00} & 1.00\,{\scriptsize $\pm$0.00} & 26\,{\scriptsize $\pm$1.1} & 0\,{\scriptsize $\pm$0.00} \\
& & Matryoshka & 64\,{\scriptsize $\pm$0.1} & 0.92\,{\scriptsize $\pm$0.00} & 0.66\,{\scriptsize $\pm$0.00} & 0.67\,{\scriptsize $\pm$0.00} & 1.00\,{\scriptsize $\pm$0.00} & 0\,{\scriptsize $\pm$0.58} & 0\,{\scriptsize $\pm$0.00} \\
& & MP-SAE     & 64\,{\scriptsize $\pm$0.0} & \textbf{0.99}\,{\scriptsize $\pm$0.00} & 0.86\,{\scriptsize $\pm$0.00} & 0.56\,{\scriptsize $\pm$0.00} & 0.79\,{\scriptsize $\pm$0.00} & 1292\,{\scriptsize $\pm$8} & 0\,{\scriptsize $\pm$0.00} \\
& & iSAE       & 64\,{\scriptsize $\pm$0.0} & 0.92\,{\scriptsize $\pm$0.00} & 0.80\,{\scriptsize $\pm$0.00} & 0.47\,{\scriptsize $\pm$0.00} & 1.00\,{\scriptsize $\pm$0.00} & \textbf{0}\,{\scriptsize $\pm$0.00} & 0\,{\scriptsize $\pm$0.00} \\
& & ConSDL     & 64\,{\scriptsize $\pm$0.3} & 0.93\,{\scriptsize $\pm$0.01} & 0.89\,{\scriptsize $\pm$0.00} & \textbf{0.84}\,{\scriptsize $\pm$0.00} & 1.00\,{\scriptsize $\pm$0.00} & 0\,{\scriptsize $\pm$0.72} & 0\,{\scriptsize $\pm$0.00} \\
\cmidrule(lr){2-10}
& \multirow{7}{*}{$\ell_0 \approx 128$}
& ReLU       & 124\,{\scriptsize $\pm$1} & 0.92\,{\scriptsize $\pm$0.00} & 0.87\,{\scriptsize $\pm$0.00} & 0.81\,{\scriptsize $\pm$0.00} & 0.97\,{\scriptsize $\pm$0.00} & 171\,{\scriptsize $\pm$7} & 0\,{\scriptsize $\pm$0.00} \\
& & TopK       & 128\,{\scriptsize $\pm$0} & 0.95\,{\scriptsize $\pm$0.00} & 0.72\,{\scriptsize $\pm$0.00} & 0.72\,{\scriptsize $\pm$0.00} & 1.00\,{\scriptsize $\pm$0.00} & 5\,{\scriptsize $\pm$2.36} & 0\,{\scriptsize $\pm$0.00} \\
& & BatchTopK  & 128\,{\scriptsize $\pm$0} & 0.95\,{\scriptsize $\pm$0.00} & 0.72\,{\scriptsize $\pm$0.00} & 0.73\,{\scriptsize $\pm$0.00} & 1.00\,{\scriptsize $\pm$0.00} & 7\,{\scriptsize $\pm$1.50} & 0\,{\scriptsize $\pm$0.00} \\
& & Matryoshka & 128\,{\scriptsize $\pm$0} & 0.94\,{\scriptsize $\pm$0.00} & 0.60\,{\scriptsize $\pm$0.00} & 0.65\,{\scriptsize $\pm$0.00} & 1.00\,{\scriptsize $\pm$0.00} & 0\,{\scriptsize $\pm$0.58} & 0\,{\scriptsize $\pm$0.00} \\
& & MP-SAE     & 127\,{\scriptsize $\pm$0} & \textbf{0.99}\,{\scriptsize $\pm$0.00} & 0.86\,{\scriptsize $\pm$0.00} & 0.56\,{\scriptsize $\pm$0.00} & 0.79\,{\scriptsize $\pm$0.00} & 1292\,{\scriptsize $\pm$8} & 0\,{\scriptsize $\pm$0.00} \\
& & iSAE       & 128\,{\scriptsize $\pm$0} & 0.94\,{\scriptsize $\pm$0.00} & 0.76\,{\scriptsize $\pm$0.00} & 0.47\,{\scriptsize $\pm$0.00} & 1.00\,{\scriptsize $\pm$0.00} & \textbf{0}\,{\scriptsize $\pm$0.00} & 0\,{\scriptsize $\pm$0.00} \\
& & ConSDL     & 134\,{\scriptsize $\pm$0} & 0.94\,{\scriptsize $\pm$0.00} & \textbf{0.89}\,{\scriptsize $\pm$0.00} & \textbf{0.84}\,{\scriptsize $\pm$0.00} & 1.00\,{\scriptsize $\pm$0.00} & 0\,{\scriptsize $\pm$0.29} & 0\,{\scriptsize $\pm$0.00} \\
\midrule
\multirow{21}{*}{$16\times$}
& \multirow{7}{*}{$\ell_0 \approx 32$}
& ReLU       & 30\,{\scriptsize $\pm$0.9} & 0.86\,{\scriptsize $\pm$0.01} & 0.42\,{\scriptsize $\pm$0.01} & 0.31\,{\scriptsize $\pm$0.01} & 0.38\,{\scriptsize $\pm$0.06} & 366\,{\scriptsize $\pm$112} & 7243\,{\scriptsize $\pm$1086} \\
& & TopK       & 32\,{\scriptsize $\pm$0.0} & 0.92\,{\scriptsize $\pm$0.00} & 0.80\,{\scriptsize $\pm$0.00} & 0.73\,{\scriptsize $\pm$0.00} & 0.98\,{\scriptsize $\pm$0.00} & 205\,{\scriptsize $\pm$6} & 0\,{\scriptsize $\pm$0.00} \\
& & BatchTopK  & 32\,{\scriptsize $\pm$0.1} & 0.92\,{\scriptsize $\pm$0.00} & 0.80\,{\scriptsize $\pm$0.00} & 0.74\,{\scriptsize $\pm$0.00} & 0.98\,{\scriptsize $\pm$0.00} & 288\,{\scriptsize $\pm$1} & 0\,{\scriptsize $\pm$0.00} \\
& & Matryoshka & 32\,{\scriptsize $\pm$0.0} & 0.91\,{\scriptsize $\pm$0.00} & 0.68\,{\scriptsize $\pm$0.00} & 0.67\,{\scriptsize $\pm$0.00} & \textbf{1.00}\,{\scriptsize $\pm$0.00} & \textbf{15}\,{\scriptsize $\pm$2.8} & 0\,{\scriptsize $\pm$0.00} \\
& & MP-SAE     & 32\,{\scriptsize $\pm$0.0} & \textbf{0.98}\,{\scriptsize $\pm$0.00} & 0.84\,{\scriptsize $\pm$0.00} & 0.48\,{\scriptsize $\pm$0.00} & 0.75\,{\scriptsize $\pm$0.00} & 3033\,{\scriptsize $\pm$5} & 0\,{\scriptsize $\pm$0.00} \\
& & iSAE       & 32\,{\scriptsize $\pm$0.0} & 0.91\,{\scriptsize $\pm$0.00} & 0.68\,{\scriptsize $\pm$0.09} & 0.36\,{\scriptsize $\pm$0.05} & 0.89\,{\scriptsize $\pm$0.16} & 222\,{\scriptsize $\pm$384} & 1177\,{\scriptsize $\pm$2038} \\
& & ConSDL     & 34\,{\scriptsize $\pm$0.1} & 0.90\,{\scriptsize $\pm$0.00} & \textbf{0.85}\,{\scriptsize $\pm$0.00} & \textbf{0.78}\,{\scriptsize $\pm$0.00} & 0.99\,{\scriptsize $\pm$0.00} & 33\,{\scriptsize $\pm$2.5} & 32\,{\scriptsize $\pm$6.5} \\
\cmidrule(lr){2-10}
& \multirow{7}{*}{$\ell_0 \approx 64$}
& ReLU       & 64\,{\scriptsize $\pm$0.9} & 0.91\,{\scriptsize $\pm$0.00} & 0.80\,{\scriptsize $\pm$0.01} & 0.73\,{\scriptsize $\pm$0.01} & 0.84\,{\scriptsize $\pm$0.04} & 779\,{\scriptsize $\pm$66} & 1195\,{\scriptsize $\pm$618} \\
& & TopK       & 64\,{\scriptsize $\pm$0.0} & 0.94\,{\scriptsize $\pm$0.00} & 0.76\,{\scriptsize $\pm$0.00} & 0.72\,{\scriptsize $\pm$0.00} & 0.99\,{\scriptsize $\pm$0.00} & 103\,{\scriptsize $\pm$9} & 0\,{\scriptsize $\pm$0.00} \\
& & BatchTopK  & 64\,{\scriptsize $\pm$0.2} & 0.94\,{\scriptsize $\pm$0.00} & 0.76\,{\scriptsize $\pm$0.00} & 0.73\,{\scriptsize $\pm$0.00} & 0.99\,{\scriptsize $\pm$0.00} & 100\,{\scriptsize $\pm$3} & 0\,{\scriptsize $\pm$0.00} \\
& & Matryoshka & 64\,{\scriptsize $\pm$0.1} & 0.93\,{\scriptsize $\pm$0.00} & 0.64\,{\scriptsize $\pm$0.00} & 0.67\,{\scriptsize $\pm$0.00} & 1.00\,{\scriptsize $\pm$0.00} & 1\,{\scriptsize $\pm$0.29} & 0\,{\scriptsize $\pm$0.00} \\
& & MP-SAE     & 64\,{\scriptsize $\pm$0.0} & \textbf{0.99}\,{\scriptsize $\pm$0.00} & \textbf{0.84}\,{\scriptsize $\pm$0.00} & 0.48\,{\scriptsize $\pm$0.00} & 0.75\,{\scriptsize $\pm$0.00} & 3033\,{\scriptsize $\pm$5} & 0\,{\scriptsize $\pm$0.00} \\
& & iSAE       & 64\,{\scriptsize $\pm$0.0} & 0.93\,{\scriptsize $\pm$0.00} & 0.76\,{\scriptsize $\pm$0.00} & 0.45\,{\scriptsize $\pm$0.00} & 1.00\,{\scriptsize $\pm$0.00} & \textbf{0}\,{\scriptsize $\pm$0.29} & 0\,{\scriptsize $\pm$0.00} \\
& & ConSDL     & 68\,{\scriptsize $\pm$0.1} & 0.93\,{\scriptsize $\pm$0.00} & 0.84\,{\scriptsize $\pm$0.00} & \textbf{0.78}\,{\scriptsize $\pm$0.00} & 1.00\,{\scriptsize $\pm$0.00} & 0\,{\scriptsize $\pm$0.00} & 0\,{\scriptsize $\pm$0.00} \\
\cmidrule(lr){2-10}
& \multirow{7}{*}{$\ell_0 \approx 128$}
& ReLU       & 133\,{\scriptsize $\pm$1} & 0.93\,{\scriptsize $\pm$0.01} & 0.84\,{\scriptsize $\pm$0.00} & 0.76\,{\scriptsize $\pm$0.00} & 0.95\,{\scriptsize $\pm$0.00} & 494\,{\scriptsize $\pm$13} & 119\,{\scriptsize $\pm$85} \\
& & TopK       & 128\,{\scriptsize $\pm$0} & 0.95\,{\scriptsize $\pm$0.00} & 0.69\,{\scriptsize $\pm$0.00} & 0.67\,{\scriptsize $\pm$0.00} & 1.00\,{\scriptsize $\pm$0.00} & 34\,{\scriptsize $\pm$8.0} & 0\,{\scriptsize $\pm$0.00} \\
& & BatchTopK  & 128\,{\scriptsize $\pm$0} & 0.95\,{\scriptsize $\pm$0.00} & 0.71\,{\scriptsize $\pm$0.00} & 0.72\,{\scriptsize $\pm$0.00} & 1.00\,{\scriptsize $\pm$0.00} & 34\,{\scriptsize $\pm$6.1} & 0\,{\scriptsize $\pm$0.00} \\
& & Matryoshka & 128\,{\scriptsize $\pm$0} & 0.95\,{\scriptsize $\pm$0.00} & 0.60\,{\scriptsize $\pm$0.00} & 0.65\,{\scriptsize $\pm$0.00} & 1.00\,{\scriptsize $\pm$0.00} & 0\,{\scriptsize $\pm$0.00} & 0\,{\scriptsize $\pm$0.00} \\
& & MP-SAE     & 128\,{\scriptsize $\pm$0} & \textbf{0.99}\,{\scriptsize $\pm$0.00} & \textbf{0.84}\,{\scriptsize $\pm$0.00} & 0.48\,{\scriptsize $\pm$0.00} & 0.75\,{\scriptsize $\pm$0.00} & 3033\,{\scriptsize $\pm$5} & 0\,{\scriptsize $\pm$0.00} \\
& & iSAE       & 128\,{\scriptsize $\pm$0} & 0.95\,{\scriptsize $\pm$0.00} & 0.70\,{\scriptsize $\pm$0.00} & 0.46\,{\scriptsize $\pm$0.00} & 1.00\,{\scriptsize $\pm$0.00} & 0\,{\scriptsize $\pm$0.00} & 0\,{\scriptsize $\pm$0.00} \\
& & ConSDL     & 126\,{\scriptsize $\pm$0} & 0.94\,{\scriptsize $\pm$0.00} & 0.84\,{\scriptsize $\pm$0.00} & \textbf{0.78}\,{\scriptsize $\pm$0.00} & 1.00\,{\scriptsize $\pm$0.00} & 5\,{\scriptsize $\pm$0.95} & 0\,{\scriptsize $\pm$0.00} \\
\bottomrule
\end{tabular}%
}
\end{table*}

\begin{table*}[t]
\centering
\caption{
\textbf{Full results on Pythia-160M across dictionary widths and sparsity levels.}
We evaluate three expansion factors ($4\times$, $8\times$, $16\times$) and three target sparsity levels ($\ell_0\approx32$, $64$, $128$).
We report achieved $\ell_0$, reconstruction ($R^2$), cross-seed consistency (\texttt{DCS}, \texttt{MCC}), and dictionary utilization, including usable atom fraction and numbers of dead and duplicated atoms.
Metrics are averaged across training seeds, while \texttt{DCS} and \texttt{MCC} are averaged across seed pairs.
$\pm$ denotes the corresponding standard deviation.
}
\label{tab:appendix_pythia_full}

\setlength{\tabcolsep}{3.5pt}
\renewcommand{\arraystretch}{0.90}

\resizebox{\textwidth}{!}{%
\begin{tabular}{ll l c c cc ccc}
\toprule
\multirow{2}{*}{\textbf{Expansion}} &
\multirow{2}{*}{\textbf{$\ell_0$ band}} &
\multirow{2}{*}{\textbf{Method}} &
\multicolumn{1}{c}{\textbf{Sparsity}} &
\multicolumn{1}{c}{\textbf{Reconstruction}} &
\multicolumn{2}{c}{\textbf{Consistency}} &
\multicolumn{3}{c}{\textbf{Dictionary Utilization}} \\
\cmidrule(lr){4-4}
\cmidrule(lr){5-5}
\cmidrule(lr){6-7}
\cmidrule(lr){8-10}
&
&
&
$\ell_0$ &
$R^2$ &
DCS &
MCC &
Usable atom frac. &
\# Dup. &
\# Dead \\
\midrule
\multirow{21}{*}{$4\times$}
& \multirow{7}{*}{$\ell_0 \approx 32$}
& ReLU       & 31\,{\scriptsize $\pm$0.3} & 0.84\,{\scriptsize $\pm$0.00} & \textbf{0.90}\,{\scriptsize $\pm$0.00} & \textbf{0.82}\,{\scriptsize $\pm$0.01} & 0.92\,{\scriptsize $\pm$0.00} & 247\,{\scriptsize $\pm$5} & 1\,{\scriptsize $\pm$1.15} \\
& & TopK       & 32\,{\scriptsize $\pm$0.0} & 0.90\,{\scriptsize $\pm$0.00} & 0.84\,{\scriptsize $\pm$0.00} & 0.78\,{\scriptsize $\pm$0.00} & 1.00\,{\scriptsize $\pm$0.00} & 15\,{\scriptsize $\pm$1.8} & 0\,{\scriptsize $\pm$0.00} \\
& & BatchTopK  & 32\,{\scriptsize $\pm$0.1} & 0.91\,{\scriptsize $\pm$0.02} & 0.83\,{\scriptsize $\pm$0.00} & 0.78\,{\scriptsize $\pm$0.00} & 1.00\,{\scriptsize $\pm$0.00} & 13\,{\scriptsize $\pm$1.0} & 0\,{\scriptsize $\pm$0.00} \\
& & Matryoshka & 32\,{\scriptsize $\pm$0.0} & 0.90\,{\scriptsize $\pm$0.02} & 0.71\,{\scriptsize $\pm$0.00} & 0.67\,{\scriptsize $\pm$0.00} & 1.00\,{\scriptsize $\pm$0.00} & 0\,{\scriptsize $\pm$0.29} & 0\,{\scriptsize $\pm$0.00} \\
& & MP-SAE     & 32\,{\scriptsize $\pm$0.0} & \textbf{0.96}\,{\scriptsize $\pm$0.00} & 0.84\,{\scriptsize $\pm$0.01} & 0.51\,{\scriptsize $\pm$0.00} & 0.72\,{\scriptsize $\pm$0.00} & 874\,{\scriptsize $\pm$7} & 0\,{\scriptsize $\pm$0.00} \\
& & iSAE       & 32\,{\scriptsize $\pm$0.0} & 0.89\,{\scriptsize $\pm$0.00} & 0.84\,{\scriptsize $\pm$0.00} & 0.45\,{\scriptsize $\pm$0.01} & 1.00\,{\scriptsize $\pm$0.00} & \textbf{0}\,{\scriptsize $\pm$0.00} & 0\,{\scriptsize $\pm$0.00} \\
& & ConSDL     & 34\,{\scriptsize $\pm$0.3} & 0.88\,{\scriptsize $\pm$0.00} & 0.88\,{\scriptsize $\pm$0.00} & 0.81\,{\scriptsize $\pm$0.00} & 1.00\,{\scriptsize $\pm$0.00} & 6\,{\scriptsize $\pm$0.00} & 0\,{\scriptsize $\pm$0.00} \\
\cmidrule(lr){2-10}
& \multirow{7}{*}{$\ell_0 \approx 64$}
& ReLU       & 55\,{\scriptsize $\pm$0.5} & 0.89\,{\scriptsize $\pm$0.02} & \textbf{0.88}\,{\scriptsize $\pm$0.00} & 0.81\,{\scriptsize $\pm$0.00} & 0.96\,{\scriptsize $\pm$0.00} & 136\,{\scriptsize $\pm$4} & 0\,{\scriptsize $\pm$0.00} \\
& & TopK       & 64\,{\scriptsize $\pm$0.0} & 0.93\,{\scriptsize $\pm$0.02} & 0.79\,{\scriptsize $\pm$0.00} & 0.75\,{\scriptsize $\pm$0.00} & 1.00\,{\scriptsize $\pm$0.00} & 3\,{\scriptsize $\pm$1.15} & 0\,{\scriptsize $\pm$0.00} \\
& & BatchTopK  & 64\,{\scriptsize $\pm$0.2} & 0.93\,{\scriptsize $\pm$0.02} & 0.78\,{\scriptsize $\pm$0.00} & 0.75\,{\scriptsize $\pm$0.00} & 1.00\,{\scriptsize $\pm$0.00} & 4\,{\scriptsize $\pm$0.52} & 0\,{\scriptsize $\pm$0.00} \\
& & Matryoshka & 64\,{\scriptsize $\pm$0.1} & 0.92\,{\scriptsize $\pm$0.02} & 0.66\,{\scriptsize $\pm$0.00} & 0.64\,{\scriptsize $\pm$0.00} & 1.00\,{\scriptsize $\pm$0.00} & 0\,{\scriptsize $\pm$0.00} & 0\,{\scriptsize $\pm$0.00} \\
& & MP-SAE     & 64\,{\scriptsize $\pm$0.1} & \textbf{0.97}\,{\scriptsize $\pm$0.00} & 0.84\,{\scriptsize $\pm$0.01} & 0.51\,{\scriptsize $\pm$0.00} & 0.72\,{\scriptsize $\pm$0.00} & 874\,{\scriptsize $\pm$7} & 0\,{\scriptsize $\pm$0.00} \\
& & iSAE       & 64\,{\scriptsize $\pm$0.0} & 0.91\,{\scriptsize $\pm$0.00} & 0.83\,{\scriptsize $\pm$0.00} & 0.46\,{\scriptsize $\pm$0.00} & 1.00\,{\scriptsize $\pm$0.00} & 0\,{\scriptsize $\pm$0.00} & 0\,{\scriptsize $\pm$0.00} \\
& & ConSDL     & 63\,{\scriptsize $\pm$0.6} & 0.88\,{\scriptsize $\pm$0.00} & 0.87\,{\scriptsize $\pm$0.00} & \textbf{0.82}\,{\scriptsize $\pm$0.00} & 1.00\,{\scriptsize $\pm$0.00} & 0\,{\scriptsize $\pm$0.00} & 0\,{\scriptsize $\pm$0.00} \\
\cmidrule(lr){2-10}
& \multirow{7}{*}{$\ell_0 \approx 128$}
& ReLU       & 126\,{\scriptsize $\pm$1} & 0.90\,{\scriptsize $\pm$0.00} & 0.84\,{\scriptsize $\pm$0.00} & 0.78\,{\scriptsize $\pm$0.00} & 0.99\,{\scriptsize $\pm$0.00} & 45\,{\scriptsize $\pm$3.2} & 0\,{\scriptsize $\pm$0.00} \\
& & TopK       & 128\,{\scriptsize $\pm$0} & 0.95\,{\scriptsize $\pm$0.01} & 0.70\,{\scriptsize $\pm$0.00} & 0.68\,{\scriptsize $\pm$0.00} & 1.00\,{\scriptsize $\pm$0.00} & 1\,{\scriptsize $\pm$0.00} & 0\,{\scriptsize $\pm$0.00} \\
& & BatchTopK  & 128\,{\scriptsize $\pm$0} & 0.95\,{\scriptsize $\pm$0.01} & 0.69\,{\scriptsize $\pm$0.00} & 0.69\,{\scriptsize $\pm$0.00} & 1.00\,{\scriptsize $\pm$0.00} & 1\,{\scriptsize $\pm$0.58} & 0\,{\scriptsize $\pm$0.00} \\
& & Matryoshka & 128\,{\scriptsize $\pm$0} & 0.94\,{\scriptsize $\pm$0.01} & 0.58\,{\scriptsize $\pm$0.00} & 0.59\,{\scriptsize $\pm$0.00} & 1.00\,{\scriptsize $\pm$0.00} & 0\,{\scriptsize $\pm$0.00} & 0\,{\scriptsize $\pm$0.00} \\
& & MP-SAE     & 127\,{\scriptsize $\pm$0} & \textbf{0.98}\,{\scriptsize $\pm$0.00} & 0.84\,{\scriptsize $\pm$0.01} & 0.51\,{\scriptsize $\pm$0.00} & 0.72\,{\scriptsize $\pm$0.00} & 874\,{\scriptsize $\pm$7} & 0\,{\scriptsize $\pm$0.00} \\
& & iSAE       & 128\,{\scriptsize $\pm$0} & 0.94\,{\scriptsize $\pm$0.00} & 0.79\,{\scriptsize $\pm$0.01} & 0.45\,{\scriptsize $\pm$0.00} & 1.00\,{\scriptsize $\pm$0.00} & 0\,{\scriptsize $\pm$0.00} & 0\,{\scriptsize $\pm$0.00} \\
& & ConSDL     & 119\,{\scriptsize $\pm$0} & 0.86\,{\scriptsize $\pm$0.00} & \textbf{0.85}\,{\scriptsize $\pm$0.00} & \textbf{0.78}\,{\scriptsize $\pm$0.00} & 1.00\,{\scriptsize $\pm$0.00} & 1\,{\scriptsize $\pm$0.29} & 0\,{\scriptsize $\pm$0.00} \\
\midrule
\multirow{21}{*}{$8\times$}
& \multirow{7}{*}{$\ell_0 \approx 32$}
& ReLU       & 34\,{\scriptsize $\pm$0.3} & 0.86\,{\scriptsize $\pm$0.00} & \textbf{0.89}\,{\scriptsize $\pm$0.00} & \textbf{0.80}\,{\scriptsize $\pm$0.00} & 0.90\,{\scriptsize $\pm$0.00} & 640\,{\scriptsize $\pm$15} & 1\,{\scriptsize $\pm$1.00} \\
& & TopK       & 32\,{\scriptsize $\pm$0.0} & 0.91\,{\scriptsize $\pm$0.00} & 0.81\,{\scriptsize $\pm$0.00} & 0.74\,{\scriptsize $\pm$0.00} & 0.99\,{\scriptsize $\pm$0.00} & 51\,{\scriptsize $\pm$3.6} & 0\,{\scriptsize $\pm$0.58} \\
& & BatchTopK  & 32\,{\scriptsize $\pm$0.1} & 0.92\,{\scriptsize $\pm$0.02} & 0.82\,{\scriptsize $\pm$0.00} & 0.75\,{\scriptsize $\pm$0.00} & 0.99\,{\scriptsize $\pm$0.00} & 44\,{\scriptsize $\pm$4.5} & 0\,{\scriptsize $\pm$0.00} \\
& & Matryoshka & 32\,{\scriptsize $\pm$0.0} & 0.91\,{\scriptsize $\pm$0.02} & 0.68\,{\scriptsize $\pm$0.00} & 0.65\,{\scriptsize $\pm$0.00} & 1.00\,{\scriptsize $\pm$0.00} & 2\,{\scriptsize $\pm$0.50} & 0\,{\scriptsize $\pm$0.00} \\
& & MP-SAE     & 32\,{\scriptsize $\pm$0.0} & \textbf{0.97}\,{\scriptsize $\pm$0.00} & 0.87\,{\scriptsize $\pm$0.01} & 0.47\,{\scriptsize $\pm$0.01} & 0.63\,{\scriptsize $\pm$0.00} & 2273\,{\scriptsize $\pm$29} & 0\,{\scriptsize $\pm$0.00} \\
& & iSAE       & 32\,{\scriptsize $\pm$0.0} & 0.90\,{\scriptsize $\pm$0.00} & 0.82\,{\scriptsize $\pm$0.00} & 0.43\,{\scriptsize $\pm$0.00} & 1.00\,{\scriptsize $\pm$0.00} & \textbf{0}\,{\scriptsize $\pm$0.00} & 0\,{\scriptsize $\pm$0.00} \\
& & ConSDL     & 35\,{\scriptsize $\pm$0.2} & 0.90\,{\scriptsize $\pm$0.00} & 0.86\,{\scriptsize $\pm$0.00} & 0.77\,{\scriptsize $\pm$0.00} & 1.00\,{\scriptsize $\pm$0.00} & 7\,{\scriptsize $\pm$1.26} & 0\,{\scriptsize $\pm$0.00} \\
\cmidrule(lr){2-10}
& \multirow{7}{*}{$\ell_0 \approx 64$}
& ReLU       & 61\,{\scriptsize $\pm$0.3} & 0.90\,{\scriptsize $\pm$0.02} & 0.87\,{\scriptsize $\pm$0.00} & \textbf{0.78}\,{\scriptsize $\pm$0.00} & 0.94\,{\scriptsize $\pm$0.00} & 375\,{\scriptsize $\pm$2} & 0\,{\scriptsize $\pm$0.00} \\
& & TopK       & 64\,{\scriptsize $\pm$0.0} & 0.93\,{\scriptsize $\pm$0.01} & 0.75\,{\scriptsize $\pm$0.00} & 0.70\,{\scriptsize $\pm$0.00} & 1.00\,{\scriptsize $\pm$0.00} & 16\,{\scriptsize $\pm$1.7} & 0\,{\scriptsize $\pm$0.00} \\
& & BatchTopK  & 64\,{\scriptsize $\pm$0.2} & 0.94\,{\scriptsize $\pm$0.01} & 0.76\,{\scriptsize $\pm$0.00} & 0.71\,{\scriptsize $\pm$0.00} & 1.00\,{\scriptsize $\pm$0.00} & 15\,{\scriptsize $\pm$2.6} & 0\,{\scriptsize $\pm$0.00} \\
& & Matryoshka & 64\,{\scriptsize $\pm$0.1} & 0.93\,{\scriptsize $\pm$0.01} & 0.64\,{\scriptsize $\pm$0.00} & 0.63\,{\scriptsize $\pm$0.00} & 1.00\,{\scriptsize $\pm$0.00} & 0\,{\scriptsize $\pm$0.00} & 0\,{\scriptsize $\pm$0.00} \\
& & MP-SAE     & 64\,{\scriptsize $\pm$0.0} & \textbf{0.97}\,{\scriptsize $\pm$0.00} & 0.87\,{\scriptsize $\pm$0.01} & 0.47\,{\scriptsize $\pm$0.01} & 0.63\,{\scriptsize $\pm$0.00} & 2273\,{\scriptsize $\pm$29} & 0\,{\scriptsize $\pm$0.00} \\
& & iSAE       & 64\,{\scriptsize $\pm$0.0} & 0.92\,{\scriptsize $\pm$0.00} & 0.80\,{\scriptsize $\pm$0.00} & 0.44\,{\scriptsize $\pm$0.00} & 1.00\,{\scriptsize $\pm$0.00} & 0\,{\scriptsize $\pm$0.00} & 0\,{\scriptsize $\pm$0.00} \\
& & ConSDL     & 69\,{\scriptsize $\pm$0.1} & 0.88\,{\scriptsize $\pm$0.00} & 0.86\,{\scriptsize $\pm$0.00} & 0.77\,{\scriptsize $\pm$0.00} & 1.00\,{\scriptsize $\pm$0.00} & 0\,{\scriptsize $\pm$0.00} & 0\,{\scriptsize $\pm$0.00} \\
\cmidrule(lr){2-10}
& \multirow{7}{*}{$\ell_0 \approx 128$}
& ReLU       & 123\,{\scriptsize $\pm$0} & 0.91\,{\scriptsize $\pm$0.00} & 0.83\,{\scriptsize $\pm$0.00} & 0.74\,{\scriptsize $\pm$0.00} & 0.97\,{\scriptsize $\pm$0.00} & 172\,{\scriptsize $\pm$3} & 0\,{\scriptsize $\pm$0.00} \\
& & TopK       & 128\,{\scriptsize $\pm$0} & 0.95\,{\scriptsize $\pm$0.01} & 0.67\,{\scriptsize $\pm$0.00} & 0.63\,{\scriptsize $\pm$0.00} & 1.00\,{\scriptsize $\pm$0.00} & 4\,{\scriptsize $\pm$1.52} & 0\,{\scriptsize $\pm$0.00} \\
& & BatchTopK  & 128\,{\scriptsize $\pm$0} & 0.96\,{\scriptsize $\pm$0.01} & 0.69\,{\scriptsize $\pm$0.00} & 0.67\,{\scriptsize $\pm$0.00} & 1.00\,{\scriptsize $\pm$0.00} & 8\,{\scriptsize $\pm$1.04} & 0\,{\scriptsize $\pm$0.00} \\
& & Matryoshka & 128\,{\scriptsize $\pm$0} & 0.95\,{\scriptsize $\pm$0.01} & 0.58\,{\scriptsize $\pm$0.00} & 0.62\,{\scriptsize $\pm$0.00} & 1.00\,{\scriptsize $\pm$0.00} & 0\,{\scriptsize $\pm$0.00} & 0\,{\scriptsize $\pm$0.00} \\
& & MP-SAE     & 127\,{\scriptsize $\pm$0} & \textbf{0.98}\,{\scriptsize $\pm$0.00} & \textbf{0.87}\,{\scriptsize $\pm$0.01} & 0.47\,{\scriptsize $\pm$0.01} & 0.63\,{\scriptsize $\pm$0.00} & 2273\,{\scriptsize $\pm$29} & 0\,{\scriptsize $\pm$0.00} \\
& & iSAE       & 128\,{\scriptsize $\pm$0} & 0.94\,{\scriptsize $\pm$0.00} & 0.76\,{\scriptsize $\pm$0.01} & 0.44\,{\scriptsize $\pm$0.01} & 1.00\,{\scriptsize $\pm$0.00} & 0\,{\scriptsize $\pm$0.00} & 0\,{\scriptsize $\pm$0.00} \\
& & ConSDL     & 135\,{\scriptsize $\pm$0} & 0.87\,{\scriptsize $\pm$0.00} & 0.82\,{\scriptsize $\pm$0.00} & \textbf{0.75}\,{\scriptsize $\pm$0.00} & 1.00\,{\scriptsize $\pm$0.00} & 2\,{\scriptsize $\pm$0.75} & 0\,{\scriptsize $\pm$0.00} \\
\midrule
\multirow{21}{*}{$16\times$}
& \multirow{7}{*}{$\ell_0 \approx 32$}
& ReLU       & 33\,{\scriptsize $\pm$0.3} & 0.86\,{\scriptsize $\pm$0.00} & 0.80\,{\scriptsize $\pm$0.00} & 0.69\,{\scriptsize $\pm$0.00} & 0.83\,{\scriptsize $\pm$0.00} & 1369\,{\scriptsize $\pm$20} & 709\,{\scriptsize $\pm$34} \\
& & TopK       & 32\,{\scriptsize $\pm$0.0} & 0.92\,{\scriptsize $\pm$0.02} & 0.78\,{\scriptsize $\pm$0.00} & 0.70\,{\scriptsize $\pm$0.00} & 0.99\,{\scriptsize $\pm$0.00} & 121\,{\scriptsize $\pm$10} & 0\,{\scriptsize $\pm$0.00} \\
& & BatchTopK  & 32\,{\scriptsize $\pm$0.2} & 0.92\,{\scriptsize $\pm$0.02} & 0.79\,{\scriptsize $\pm$0.00} & 0.72\,{\scriptsize $\pm$0.00} & 0.99\,{\scriptsize $\pm$0.00} & 108\,{\scriptsize $\pm$5} & 0\,{\scriptsize $\pm$0.00} \\
& & Matryoshka & 32\,{\scriptsize $\pm$0.1} & 0.92\,{\scriptsize $\pm$0.02} & 0.58\,{\scriptsize $\pm$0.00} & 0.52\,{\scriptsize $\pm$0.00} & \textbf{1.00}\,{\scriptsize $\pm$0.00} & \textbf{5}\,{\scriptsize $\pm$2.38} & 0\,{\scriptsize $\pm$0.00} \\
& & MP-SAE     & 32\,{\scriptsize $\pm$0.0} & \textbf{0.97}\,{\scriptsize $\pm$0.00} & \textbf{0.86}\,{\scriptsize $\pm$0.01} & 0.41\,{\scriptsize $\pm$0.01} & 0.61\,{\scriptsize $\pm$0.00} & 4809\,{\scriptsize $\pm$26} & 0\,{\scriptsize $\pm$0.00} \\
& & iSAE       & 32\,{\scriptsize $\pm$0.0} & 0.90\,{\scriptsize $\pm$0.00} & 0.70\,{\scriptsize $\pm$0.00} & 0.31\,{\scriptsize $\pm$0.00} & 0.75\,{\scriptsize $\pm$0.02} & 466\,{\scriptsize $\pm$43} & 2618\,{\scriptsize $\pm$232} \\
& & ConSDL     & 33\,{\scriptsize $\pm$0.1} & 0.91\,{\scriptsize $\pm$0.00} & 0.83\,{\scriptsize $\pm$0.00} & \textbf{0.72}\,{\scriptsize $\pm$0.00} & 1.00\,{\scriptsize $\pm$0.00} & 8\,{\scriptsize $\pm$2.29} & 52\,{\scriptsize $\pm$7.2} \\
\cmidrule(lr){2-10}
& \multirow{7}{*}{$\ell_0 \approx 64$}
& ReLU       & 64\,{\scriptsize $\pm$0.5} & 0.91\,{\scriptsize $\pm$0.02} & 0.85\,{\scriptsize $\pm$0.00} & \textbf{0.73}\,{\scriptsize $\pm$0.00} & 0.92\,{\scriptsize $\pm$0.00} & 1040\,{\scriptsize $\pm$5} & 0\,{\scriptsize $\pm$0.00} \\
& & TopK       & 64\,{\scriptsize $\pm$0.0} & 0.94\,{\scriptsize $\pm$0.01} & 0.71\,{\scriptsize $\pm$0.00} & 0.65\,{\scriptsize $\pm$0.00} & 1.00\,{\scriptsize $\pm$0.00} & 50\,{\scriptsize $\pm$2.6} & 0\,{\scriptsize $\pm$0.00} \\
& & BatchTopK  & 64\,{\scriptsize $\pm$0.3} & 0.94\,{\scriptsize $\pm$0.01} & 0.73\,{\scriptsize $\pm$0.00} & 0.68\,{\scriptsize $\pm$0.00} & 1.00\,{\scriptsize $\pm$0.00} & 42\,{\scriptsize $\pm$1.3} & 0\,{\scriptsize $\pm$0.00} \\
& & Matryoshka & 64\,{\scriptsize $\pm$0.2} & 0.94\,{\scriptsize $\pm$0.01} & 0.59\,{\scriptsize $\pm$0.00} & 0.59\,{\scriptsize $\pm$0.00} & 1.00\,{\scriptsize $\pm$0.00} & 1\,{\scriptsize $\pm$0.76} & 0\,{\scriptsize $\pm$0.00} \\
& & MP-SAE     & 64\,{\scriptsize $\pm$0.0} & \textbf{0.98}\,{\scriptsize $\pm$0.00} & \textbf{0.86}\,{\scriptsize $\pm$0.01} & 0.41\,{\scriptsize $\pm$0.01} & 0.61\,{\scriptsize $\pm$0.00} & 4809\,{\scriptsize $\pm$26} & 0\,{\scriptsize $\pm$0.00} \\
& & iSAE       & 64\,{\scriptsize $\pm$0.0} & 0.93\,{\scriptsize $\pm$0.00} & 0.77\,{\scriptsize $\pm$0.00} & 0.43\,{\scriptsize $\pm$0.00} & 1.00\,{\scriptsize $\pm$0.00} & 0\,{\scriptsize $\pm$0.00} & 0\,{\scriptsize $\pm$0.00} \\
& & ConSDL     & 75\,{\scriptsize $\pm$0.3} & 0.91\,{\scriptsize $\pm$0.00} & 0.83\,{\scriptsize $\pm$0.00} & 0.73\,{\scriptsize $\pm$0.00} & 1.00\,{\scriptsize $\pm$0.00} & 0\,{\scriptsize $\pm$0.00} & 0\,{\scriptsize $\pm$0.00} \\
\cmidrule(lr){2-10}
& \multirow{7}{*}{$\ell_0 \approx 128$}
& ReLU       & 130\,{\scriptsize $\pm$0} & 0.92\,{\scriptsize $\pm$0.00} & 0.80\,{\scriptsize $\pm$0.00} & 0.68\,{\scriptsize $\pm$0.00} & 0.96\,{\scriptsize $\pm$0.00} & 490\,{\scriptsize $\pm$2} & 0\,{\scriptsize $\pm$0.00} \\
& & TopK       & 128\,{\scriptsize $\pm$0} & 0.96\,{\scriptsize $\pm$0.01} & 0.61\,{\scriptsize $\pm$0.00} & 0.56\,{\scriptsize $\pm$0.00} & 1.00\,{\scriptsize $\pm$0.00} & 19\,{\scriptsize $\pm$1.3} & 0\,{\scriptsize $\pm$0.00} \\
& & BatchTopK  & 129\,{\scriptsize $\pm$1} & 0.96\,{\scriptsize $\pm$0.01} & 0.63\,{\scriptsize $\pm$0.00} & 0.60\,{\scriptsize $\pm$0.00} & 1.00\,{\scriptsize $\pm$0.00} & 19\,{\scriptsize $\pm$4.2} & 0\,{\scriptsize $\pm$0.00} \\
& & Matryoshka & 128\,{\scriptsize $\pm$0} & 0.95\,{\scriptsize $\pm$0.01} & 0.54\,{\scriptsize $\pm$0.00} & 0.57\,{\scriptsize $\pm$0.00} & 1.00\,{\scriptsize $\pm$0.00} & 0\,{\scriptsize $\pm$0.50} & 0\,{\scriptsize $\pm$0.00} \\
& & MP-SAE     & 127\,{\scriptsize $\pm$0} & \textbf{0.99}\,{\scriptsize $\pm$0.00} & \textbf{0.86}\,{\scriptsize $\pm$0.01} & 0.41\,{\scriptsize $\pm$0.01} & 0.61\,{\scriptsize $\pm$0.00} & 4809\,{\scriptsize $\pm$26} & 0\,{\scriptsize $\pm$0.00} \\
& & iSAE       & 128\,{\scriptsize $\pm$0} & 0.95\,{\scriptsize $\pm$0.00} & 0.70\,{\scriptsize $\pm$0.00} & 0.43\,{\scriptsize $\pm$0.00} & 1.00\,{\scriptsize $\pm$0.00} & 0\,{\scriptsize $\pm$0.00} & 0\,{\scriptsize $\pm$0.00} \\
& & ConSDL     & 135\,{\scriptsize $\pm$0} & 0.92\,{\scriptsize $\pm$0.00} & 0.83\,{\scriptsize $\pm$0.00} & \textbf{0.74}\,{\scriptsize $\pm$0.00} & 1.00\,{\scriptsize $\pm$0.00} & 0\,{\scriptsize $\pm$0.00} & 0\,{\scriptsize $\pm$0.00} \\
\bottomrule
\end{tabular}%
}
\end{table*}

\begin{table*}[t]
\centering
\caption{
\textbf{Hyperparameters for sparsity matching and ConSDL training.}
For each backbone, expansion, and target sparsity level
($\ell_0 \approx 32, 64, 128$), we report the ReLU $\ell_1$ coefficient
and ConSDL sparse-coding penalty $\lambda$ used for sparsity matching.
For ConSDL, we additionally report the number of FISTA iterations
$T_{\mathrm{FISTA}}$, coefficient-refitting steps $T_{\mathrm{refit}}$,
and the co-active similarity weight $\gamma_{\mathrm{sim}}$.
}
\label{tab:hparams}
\small
\setlength{\tabcolsep}{4pt}
\resizebox{\textwidth}{!}{%
\begin{tabular}{ll ccc ccc ccc ccc ccc}
\toprule
\multirow{3}{*}{\textbf{Backbone}} & \multirow{3}{*}{\textbf{Expansion}} & \multicolumn{3}{c}{\textbf{ReLU}} & \multicolumn{12}{c}{\textbf{ConSDL}} \\
\cmidrule(lr){3-5} \cmidrule(lr){6-17}
& & \multicolumn{3}{c}{$\ell_1$} & \multicolumn{3}{c}{$\lambda$} & \multicolumn{3}{c}{$T_{\mathrm{FISTA}}$} & \multicolumn{3}{c}{$T_{\mathrm{refit}}$} & \multicolumn{3}{c}{$\gamma_{\mathrm{sim}}$} \\
\cmidrule(lr){3-5} \cmidrule(lr){6-8} \cmidrule(lr){9-11} \cmidrule(lr){12-14} \cmidrule(lr){15-17}
& & $32$ & $64$ & $128$ & $32$ & $64$ & $128$ & $32$ & $64$ & $128$ & $32$ & $64$ & $128$ & $32$ & $64$ & $128$ \\
\midrule
\multirow{3}{*}{CLIP ViT-B/16} & $\times4$ & \texttt{1.4e-4} & \texttt{1.0e-4} & \texttt{7.5e-5} & $0.8$ & $0.6$ & $0.35$ & $30$ & $30$ & $30$ & $10$ & $5$ & $5$ & \texttt{1e-3} & \texttt{1e-3} & \texttt{1e-3} \\
 & $\times8$ & \texttt{1.5e-4} & \texttt{1.2e-4} & \texttt{8.4e-5} & $0.8$ & $0.8$ & $0.5$ & $30$ & $30$ & $30$ & $60$ & $10$ & $5$ & \texttt{5e-3} & \texttt{1e-3} & \texttt{1e-3} \\
 & $\times16$ & \texttt{1.6e-4} & \texttt{1.3e-4} & \texttt{1.0e-4} & $1$ & $1.25$ & $0.75$ & $50$ & $30$ & $30$ & $50$ & $10$ & $10$ & \texttt{5e-3} & \texttt{5e-3} & \texttt{2e-3} \\
\midrule
\multirow{3}{*}{DINOv2-B/14} & $\times4$ & \texttt{1.5e-4} & \texttt{8.5e-5} & \texttt{5.6e-5} & $1.76$ & $1.08$ & $0.74$ & $30$ & $30$ & $30$ & $10$ & $10$ & $10$ & \texttt{5e-3} & \texttt{5e-3} & \texttt{1e-3} \\
 & $\times8$ & \texttt{1.5e-4} & \texttt{1.2e-4} & \texttt{6.8e-5} & $2$ & $1.65$ & $0.88$ & $50$ & $30$ & $30$ & $30$ & $10$ & $10$ & \texttt{5e-3} & \texttt{5e-3} & \texttt{1e-3} \\
 & $\times16$ & \texttt{1.5e-4} & \texttt{1.2e-4} & \texttt{8.5e-5} & $2.8$ & $2.5$ & $1.6$ & $50$ & $30$ & $30$ & $40$ & $22$ & $5$ & \texttt{5e-3} & \texttt{5e-3} & \texttt{5e-3} \\
\midrule
\multirow{3}{*}{Pythia-160M} & $\times4$ & \texttt{1.2e-4} & \texttt{8.0e-5} & \texttt{4.5e-5} & $1.5$ & $1.2$ & $1.2$ & $30$ & $30$ & $30$ & $13$ & $8$ & $4$ & \texttt{1e-3} & \texttt{5e-3} & \texttt{5e-3} \\
 & $\times8$ & \texttt{1.2e-4} & \texttt{8.0e-5} & \texttt{5.0e-5} & $1.5$ & $1.5$ & $1.5$ & $50$ & $50$ & $30$ & $22$ & $15$ & $8$ & \texttt{1e-3} & \texttt{5e-3} & \texttt{5e-3} \\
 & $\times16$ & \texttt{1.3e-4} & \texttt{8.0e-5} & \texttt{5.0e-5} & $1.7$ & $1.5$ & $1$ & $50$ & $50$ & $50$ & $50$ & $26$ & $14$ & \texttt{1e-3} & \texttt{5e-3} & \texttt{5e-3} \\
\midrule
\multirow{3}{*}{GPT-2 Small} & $\times4$ & \texttt{1.2e-4} & \texttt{7.0e-5} & \texttt{4.5e-5} & $7.1$ & $4.6$ & $2.95$ & $30$ & $30$ & $30$ & $15$ & $5$ & $5$ & \texttt{5e-3} & \texttt{5e-3} & \texttt{5e-3} \\
 & $\times8$ & \texttt{1.2e-4} & \texttt{7.5e-5} & \texttt{5.0e-5} & $8.38$ & $5.35$ & $3.4$ & $30$ & $30$ & $30$ & $15$ & $10$ & $5$ & \texttt{5e-3} & \texttt{5e-3} & \texttt{5e-3} \\
 & $\times16$ & \texttt{1.3e-4} & \texttt{8.0e-5} & \texttt{5.0e-5} & $12.5$ & $6.5$ & $4.5$ & $30$ & $30$ & $30$ & $30$ & $18$ & $10$ & \texttt{5e-3} & \texttt{5e-3} & \texttt{2e-3} \\
\bottomrule
\end{tabular}%
}
\end{table*}

\begin{table*}[t]
\centering
\caption{
\textbf{Hyperparameters for the synthetic experiments.}
For each ground-truth dictionary size and tail exponent $\beta$, we report the ReLU sparsity coefficient $\ell_1$ and the ConSDL sparse-coding penalty $\lambda$ and number of refitting steps $T_{\mathrm{refit}}$. ConSDL uses $T_{\mathrm{FISTA}}=50$, and $\gamma_{\mathrm{sim}}=\texttt{1e-3}$ in all settings.
}
\label{tab:synthetic_hparams}

\scriptsize
\setlength{\tabcolsep}{4pt}

\resizebox{\textwidth}{!}{%
\begin{tabular}{c ccc ccc ccc ccc}
\toprule
\multirow{2}{*}{$\boldsymbol{\beta}$} &
\multicolumn{3}{c}{\textbf{3K}} &
\multicolumn{3}{c}{\textbf{6K}} &
\multicolumn{3}{c}{\textbf{9K}} &
\multicolumn{3}{c}{\textbf{12K}} \\
\cmidrule(lr){2-4}
\cmidrule(lr){5-7}
\cmidrule(lr){8-10}
\cmidrule(lr){11-13}
&
ReLU $\ell_1$ & Ours $\lambda$ & Ours $T_{\mathrm{refit}}$ &
ReLU $\ell_1$ & Ours $\lambda$ & Ours $T_{\mathrm{refit}}$ &
ReLU $\ell_1$ & Ours $\lambda$ & Ours $T_{\mathrm{refit}}$ &
ReLU $\ell_1$ & Ours $\lambda$ & Ours $T_{\mathrm{refit}}$ \\
\midrule
0.0 &
\texttt{1.5e-4} & \texttt{0.3} & \texttt{30} & \texttt{1.4e-4} & \texttt{0.5} & \texttt{10} & \texttt{1.8e-4} & \texttt{0.4} & \texttt{10} & \texttt{1.8e-4} & \texttt{0.5} & \texttt{5} \\
0.5 &
\texttt{1e-4} & \texttt{0.3} & \texttt{30} & \texttt{1.2e-4} & \texttt{0.4} & \texttt{10} & \texttt{1.8e-4} & \texttt{0.3} & \texttt{10} & \texttt{1.6e-4} & \texttt{0.4} & \texttt{10} \\
0.7 &
\texttt{1.5e-4} & \texttt{0.35} & \texttt{20} & \texttt{2.1e-4} & \texttt{0.25} & \texttt{30} & \texttt{1.8e-4} & \texttt{0.35} & \texttt{30} & \texttt{2e-4} & \texttt{0.3} & \texttt{10} \\
1.0 &
\texttt{2.1e-4} & \texttt{0.15} & \texttt{30} & \texttt{2.3e-4} & \texttt{0.15} & \texttt{20} & \texttt{2.1e-4} & \texttt{0.25} & \texttt{30} & \texttt{2.1e-4} & \texttt{0.3} & \texttt{20} \\
\bottomrule
\end{tabular}%
}
\end{table*}

\end{document}